%% file: main.tex
\documentclass[11pt]{article}
\pdfoutput=1
\usepackage{booktabs}
\usepackage{pgfplots}
\usepackage{graphicx}
\usepackage{tikz}
\usepackage{amsmath, amsthm}
\usepackage{amssymb}
\usepackage{graphicx}
\usepackage{setspace}
\usepackage{xfrac}
\usepackage{hyperref}
\usepackage{xstring}
\usepackage{xspace}
\usepackage{bm}
\usepackage{microtype}
\usepackage{graphicx}

\usepackage{booktabs} 

\usepackage{comment}
\pgfplotsset{compat=1.14}

\usepackage{gensymb}
\usepackage{hyperref}       
\usepackage{amsfonts}       
\usepackage{nicefrac}       
\usepackage{microtype}      
\usepackage{amsmath}
\usepackage{amsfonts}
\usepackage{amssymb}
\usepackage{comment}

\usepackage{subcaption}
\usepackage{hyperref}

\usepackage{amsthm}
\usepackage{amscd}
\usepackage{t1enc}
\usepackage{enumerate}
\usepackage{enumitem}
\usepackage[mathscr]{eucal}
\usepackage{indentfirst}
\usepackage{listings}
\usepackage{graphicx}
\usepackage{graphics}
\usepackage{pict2e}
\usepackage{epic}
\usepackage{epstopdf} 
\usepackage{wrapfig}
\usepackage{algorithm}
\usepackage{algorithmic}
\usepackage{amsthm}
\usepackage{tabularx}

\newcommand{\bE}{\mathbb{E}}
\newcommand{\bP}{\mathbb{P}}
\newcommand{\bI}{\mathbb{I}}

\renewcommand{\tilde}{\widetilde}
\renewcommand{\nu}{\vartheta}

\usepackage{float}

\allowdisplaybreaks
\usepackage{fullpage}
\newtheorem{theorem}{Theorem}
\newtheorem*{theorem*}{Theorem}

\newtheorem{assumption}{Assumption}
\newtheorem{proposition}{Proposition}
\newtheorem*{proposition*}{Proposition}
\theoremstyle{definition}

\newtheorem*{remark*}{Remark}
\date{}
\title{\textbf{Teaching Humans When To Defer to a Classifier \\ via Exemplars}}

\author{Hussein Mozannar \thanks{Massachusetts Institute of Technology. Email: \texttt{mozannar@mit.edu}} \and Arvind Satyanarayan \thanks{Massachusetts Institute of Technology. Email: \texttt{arvindsatya@mit.edu}} \and David Sontag \thanks{Massachusetts Institute of Technology. Email: \texttt{dsontag@csail.mit.edu}} }

\begin{document}
\maketitle
\begin{abstract}
Expert decision makers are starting to rely on data-driven automated agents to assist them with various tasks. For this collaboration to perform properly, the human decision maker must have a mental model of when and when not to rely on the agent. In this work, we aim to ensure that human decision makers learn a valid mental model of the agent's strengths and weaknesses. To accomplish this goal, we propose an exemplar-based teaching strategy where humans solve a set of selected examples and with our help generalize from them to the domain. We present a novel parameterization of the human's mental model of the AI that applies a nearest neighbor rule in local regions surrounding the teaching examples. Using this model, we derive a near-optimal strategy for selecting a representative teaching set. We validate the benefits of our teaching strategy on a multi-hop question answering task with an interpretable AI model using crowd workers. We find that when workers draw the right lessons from the teaching stage, their task performance improves. We furthermore validate our method on a set of synthetic experiments. 
\end{abstract}
\input{body}

\bibliographystyle{alpha}
\bibliography{ref}
\newpage
\input{appendix}

\end{document}

%% file: body.tex
\section{Introduction}\label{sec:intro}
Automated agents powered by machine learning are augmenting the capabilities of human decision makers in settings such as healthcare \cite{beede2020human,gaube2021ai}, content moderation \cite{link2016human}
and  more routine decisions such as asking  AI-enabled virtual assistants  for recommendations \cite{shaikh2019alexa}. This mode of interaction whereby the automated agent serves only to provide a recommendation to the human decision maker, a setting typically named \emph{AI assisted decision making}, is the focus of our study here. 
A key question is how does the human expert know when to rely on the AI for advice.  
In this work, we make the case for the need to initially onboard the human decision maker on when and when not to rely on the automated agent. We propose that before an AI agent is deployed to assist a human decision maker, the human is taught through a tailored onboarding phase how to make decisions with the help of the AI. The purpose of the onboarding is to help the human understand when to trust the AI and how  the AI can complement their abilities.
This allows the human to have an accurate mental model of the AI agent, and this mental model helps in setting expectations about the performance of the AI on different examples. 


Our onboarding phase consists of letting the human predict on a series of specially selected teaching examples in a setting that mimics the deployment use case. The examples are chosen to give an overview of the AI’s strengths and weaknesses especially when it complement's the abilities of the human. After predicting on each example, the human agent then receives feedback on their performance and that of the AI. To allow the human to generalize from each example, we display features of the region surrounding the example. Finally, to enable retention of the  example, we let the human write down a lesson indicating whether they should trust the AI in that region and what characterizes the region. Our approach is inspired by research in the education literature that highlight the importance of feedback and lesson retention for learning \cite{atkinson2000learning, hattie2007power}.

To select the teaching examples, we need to have a mathematical framework of how the human mental model evolves after we give them feedback.
We model the human thought process as first deciding whether to rely on the AI's prediction or not using an internal \emph{rejector} in section \ref{sec:problem}. 
This rejector is what we refer to as the human's mental model of the AI.
We propose to model the human's rejector as consisting of a prior rejector and a nearest neighbor rule that only applies in local regions surrounding each teaching example in section \ref{sec:human_model}. This novel parameterization is inspired by work in cognitive science  that suggests that humans make decisions by weighing similar past experiences \cite{bornstein2017reminders}.
Assuming this rejector model, we give a  near-optimal greedy strategy for selecting a set of representative teaching examples that allows us to control the examples and the region surrounding them in section \ref{sec:approach}.

We first evaluate the efficacy of our algorithmic approach on a set of synthetic experiments and its robustness to the misspecification of the human's model. For our main evaluation, we conduct experiments on Amazon Mechanical Turk  on the task of passage-based question answering from HotpotQA \cite{yang2018hotpotqa} in section \ref{sec:experiments}. Crowdworkers first performed a teaching phase and were then tested on a randomly chosen subset of examples. Our results demonstrate the importance of teaching: around half of the participants who undertook the teaching phase were able to correctly determine the AI's region of error and had a resulting improved  performance. 

\section{Related Work}\label{sec:related_work}
One of the goals of  explainable machine learning is to enable humans to better evaluate the correctness of the AI's prediction by providing supporting evidence 
\cite{lai2019human,lage2019evaluation,smith2020no,hase2020evaluating,zhang2020effect,kocielnik2019will,suresh2020misplaced,suresh2021intuitively,wortmanvaughan2021a,gonzalez2020human}. However, these explanations do not inform the decision maker how to weigh their own predictions against those of the AI or how to combine the AI's evidence to make their final decision \cite{kaur2020interpreting}.
The AI explanations cannot factor in the effect of the human's side information, and thus the human has to learn what their side information reveals about the performance of the AI or themselves.
Moreover, if the AI's explanations are unfaithful or become so due to a distribution shift in the data \cite{devries2018learning}, then the human may then over-weigh the AI's abilities. 
Another direct approach for teaching is presenting the human with a set of guidelines of when to rely on the AI \cite{amershi2019guidelines}.
However, these guidelines need to be developed by a set of domain experts and no standard approach currently exists for creating such guidelines. 
As a byproduct of our teaching approach, each human writes a set of unorganized rules that can then be more easily turned into such guidelines.

The reverse setting, of teaching a classifier when to defer to a human, is dubbed as learning to defer  (LTD) \cite{madras2018predict, raghu2019algorithmic,mozannar2020consistent,wilder2020learning}.
The main goal of LTD is to learn a rejector that determines which of the AI and the human should predict on each example.
However, there are numerous legal and accountability constraints that may prohibit a machine from making final decisions in high stakes scenarios.
Additionally, the actual test-time setting may differ from that which was used during training, but since in our setting the human makes the final decision, this allows them to adapt their decision making and detect any unexpected model errors.
As an example in a clinical use case, factors such as times of substantially increased patient load may affect the human expert's accuracy. The human may also occasionally have side-information that was unavailable to the AI that could improve their decision making.
Compared to LTD, deployment may be simplified because the same AI  is used for all experts; as new experts arrive, our onboarding phase trains them to use the AI according to their unique abilities.
Our teaching setting and LTD also use very different techniques. Although the objective that we present in Equation~\eqref{eq:second_opinion_loss} is closely related to the objective used by \cite{mozannar2020consistent}, the main task in our setting is that of teaching the human when to defer. This requires us to develop a formalization of the human mental model and algorithms for selecting a subset of examples that enables accurate learning.

Related work has explored how to best onboard a human to trust or replicate a model's prediction.
LIME, a black-box feature importance method, was used to select examples so that crowdworkers could evaluate which of two models would perform better \cite{ribeiro2016should,lai2020chicago}. Their selection strategy does not take into account the human predictor, nor does their approach do more than display the examples. 
On a task of visual question answering, \cite{chandrasekaran2018explanations} handpicked 7  examples to teach crowdworkers about the AI abilities and found that teaching improved the ability to detect the AI's failure. \cite{feng2019can} on a Quizbowl question answering task highlight the importance of modeling the skill level of the human expert when designing the explanations; this further motivates our incorporation of the human predictor into the choice of the teaching set. 
Through a study of 21 pathologists,  \cite{cai2019hello} gathered a set of guidelines of what clinicians wanted to know about an AI prior to interacting with it. 
\cite{yin2019understanding} study the effect of initial debriefing of stated AI accuracy compared to observed AI accuracy in deployment and find a
significant  effect of stated accuracy on trust, but that diminishes quickly after observing the model in practice; this reinforces our approach of building trust through examples that simulate deployment. 
 \cite{bansal2019beyond} investigate the role of the human's mental model of the AI on task accuracy, however, the mental model is formed through test time interaction rather than through an onboarding stage. \cite{bansal2021most} propose a theoretical model for AI-assisted decision making, assuming that the human has a perfect mental model of the AI and that the human has uniform error.  
 
Finally, our work was  inspired by the literature on machine teaching \cite{zhu2018overview,singla2014near,kumar2021teaching,hunziker2018teaching,dasgupta2019teaching} and curriculum learning \cite{bengio2009curriculum,graves2017automated}. Our work differentiates itself from the machine teaching literature by the use of our novel radius neighbor human model and the goal of teaching how to defer to an AI rather than teaching concepts to humans.
Studies have also explored the use of reinforcement learning as a tool for online education \cite{ruan2020supporting,doroudi2019not,lee2015learning}. We further expand the related work in Appendix \ref{apx:related_work}. 
\section{Problem Setup}\label{sec:problem}

\begin{figure}[h]

\centering
    \resizebox{0.8\textwidth}{!}{

    \includegraphics[scale=0.45,trim=0cm 9cm 8cm 0.5cm, clip]{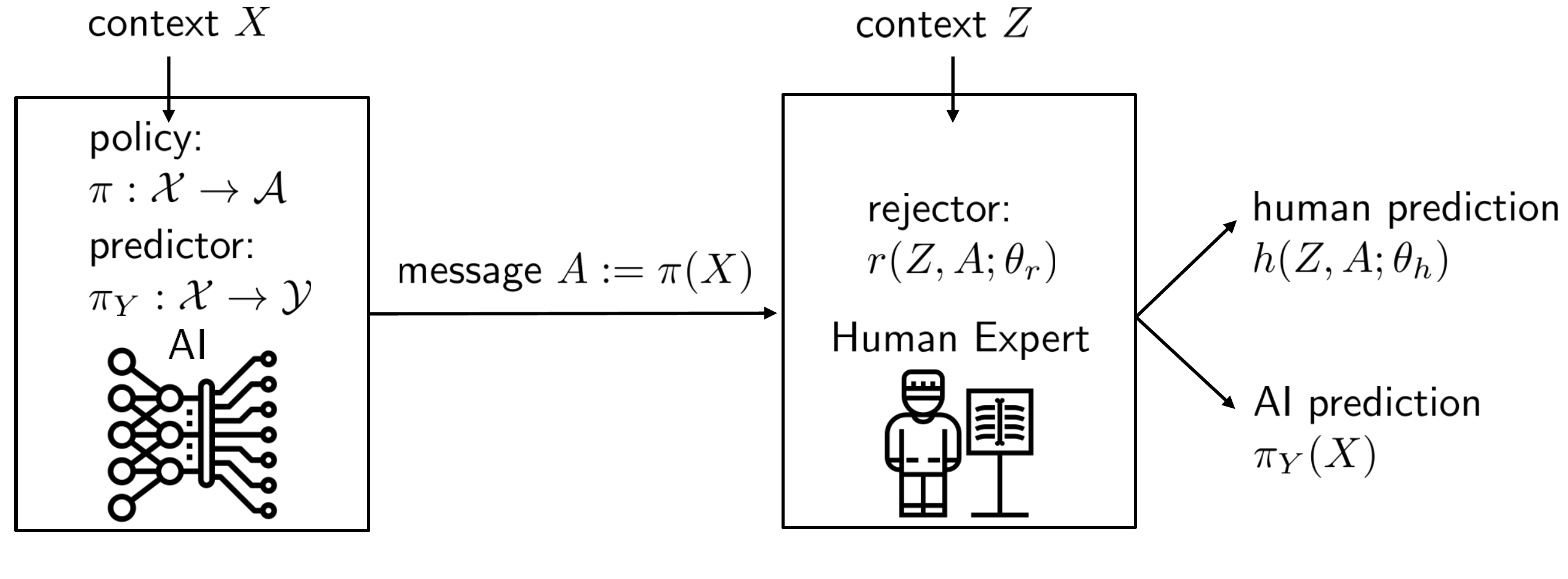}
   }
    \caption{The AI assisted decision making pipeline. The AI first sends to the human a message $A$, then the human decides with their rejector $r(Z,A)$ if they should follow the AI's advice and predict $\pi_Y(X)$ or they should predict on their own using $h(Z,A)$. }
    \label{fig:illustrate_setup}
\end{figure}

Our formalization is based on the interaction between two agents: the AI, an automated agent, and a human expert who both collaborate to predict a target  $Y \in \mathcal{Y}$ based on a given input context. The setup is as follows: the AI perceives a view of the input $X \in \mathcal{X}$, then communicates a message $A \in \mathcal{A}$ that is perceived by the human. The human expert then integrates the AI message $A$ and their own view of the input $Z \in \mathcal{Z}$ to make a final decision $M(Z,A)$ which can either be to predict on their own  or allow the AI agent to predict. The input space of the human $Z$ and that of the AI $X$  could be different since the human may have side information that the AI can't observe. This is essentially the \emph{AI-Assisted Decision Making} setup illustrated in Figure \ref{fig:illustrate_setup} which is the more common mode of interaction between humans and artificially intelligent agents in high-stakes scenarios. 

More formally, the AI consists of a predictor $\pi_Y: \mathcal{X} \to \mathcal{Y}$ that can solve the task on its own and a policy $\pi: \mathcal{X} \to \mathcal{A}$ which serves to communicate with the human. The message space $\mathcal{A}$ may consist for example of the AI's prediction $\pi_Y(X)$ alongside an explanation of their decision.  On the other hand, the human when seeing the AI's message consists of a \textbf{predictor} $h: \mathcal{Z} \times \mathcal{A} \to \mathcal{Y}$ parameterized by $\theta_h$ and the human decides to allow the AI to predict or not according to a \textbf{rejector} $r: \mathcal{Z}  \times \mathcal{A} \to \{0,1\}$ parameterized by $\theta_r$, where if $r(Z,A;\theta_r)=1$  the human uses the AI's answer for its final prediction. This implies that the final human decision $M$ is as follows: 
\begin{equation} \label{eq:human_final_decision}
M(Z, A) = \begin{cases}
\pi_Y(x) \quad, \ \text{if} \ r(Z, A; \theta_r) =1 \\
h(Z, A;\theta_h) \quad, \text{ otherwise}
\end{cases}
\end{equation}
 

\paragraph{System objective.} Given the above ingredients and a performance measure on the label space  $l(y,\hat{y}): \mathcal{Y} \times \mathcal{Y} \to \mathbb{R}^+$ (e.g. 0-1 loss), the loss that we incur is the following:
\begin{align}
    L(\pi,\pi_Y,h,r) = &\bE_{x,z,y}[ \underbrace{l(\pi_Y(x),y)}_{\text{AI cost}}  \overbrace{\mathbb{I}_{r(x, \pi(x)) = 1}}^{\text{AI predicts}} + \underbrace{l(h(z,\pi(x)),y)}_{\text{Human cost}}    \overbrace{\mathbb{I}_{r(x, \pi(x)) = 0}}^{\text{Human predicts}}   ] \label{eq:second_opinion_loss}
\end{align}

We put ourselves in the role of a system designer who has knowledge of both the human and the AI and wishes to minimize the loss of the system $L$ \eqref{eq:second_opinion_loss}. 

\paragraph{The central Human-AI interaction problem.} Given a fixed AI policy, and human parameters $(\theta_h,\theta_r)$,  the manner in which  the human expert integrates the AI's message 
depends only on the expert context $Z$ and the message itself $A$.  In particular, for two different policies $\pi_1$ and $\pi_2$ that output the same message $A$ on input $Z$, our framework tells us that the resulting behavior of the human expert would be identical in both cases. However, if it is known to the human that AI $\pi_1$ has very high error compared to AI $\pi_2$, then is more likely for them to trust the message if it is coming from $\pi_2$ rather than from $\pi_1$.
Thus it is more realistic to assume that the expert has a \emph{mental model} of the policy $\pi$ that they have arrived at from either a description of the policy or from  previously interacting with it; the rejector here formalizes the \emph{mental model}.  This insight forces us to now consider the parameters $(\theta_h, \theta_r)$ as variables that are learned by the human as a function of the underlying AI policy $\pi$. 
This makes the optimization of the loss now much more challenging as whenever  the policy $\pi$ changes, the human's mental model, $(\theta_h, \theta_r)$, needs to update. Therefore, we need to first understand how the human's mental model evolves and how we can influence it.

\paragraph{Teaching Humans about the AI.} In this work, we focus on exemplar based strategies to allow the human to update their mental models of the AI. The question is then how do we select a minimal set of examples that teaches the human an accurate mental model of the AI. To make progress, we need to first understand the form of the human's rejector and how it evolves, which we elaborate on in the following section. Crucially, we will keep the AI in this work as a fixed policy and not look to optimize for it. Once we understand this first step, future work can then look to close the loop which entails learning an updated AI with the knowledge of the human learner dynamics.

\section{Human Mental Model}\label{sec:human_model}
We now introduce our model of the human's rejector and the elements of the teaching setup.
The tasks we are interested in are where humans are \emph{domain experts}, where we define domain experts to mean that their knowledge about the task and their predictive performance are fixed. We further extend this to how they may incorporate the AI message in their prediction, but crucially not how they decide when to use the AI. This assumption translates in our formulation as follows.
\begin{assumption} \label{ass:human_indep_ai}
The human predictor does not vary as they interact with the AI, i.e. we assume $\theta_h$ to be fixed. 
\end{assumption}
While we have assumed $\theta_h$ is fixed and have so far spoken about a singular human, in reality, the AI might be deployed in conjunction with multiple human experts. These experts might have different parameters $\theta_h$ individually, however; for the rest of this paper, we focus on a singular expert that we are interacting with. 

We now move our attention to the human's rejector, which represents their mental model of the AI, and learned after observing a series of labeled examples. Research on human learning from the cognitive science literature has postulated  that for complex tasks humans make decisions by sampling similar experiences from memory  \cite{bornstein2017reminders,giguere2013limits,richler2014visual}. Moreover, \cite{bornstein2017reminders} makes the explicit comparison with nearest neighbor models found in machine learning.
However, standard nearest neighbor models don't allow for prior knowledge to be incorporated. For this reason, we postulate a nearest neighbor model for the human rejector that starts with a prior and updates in local regions of each shown example in the following assumption.

\begin{figure}

\centering
    \resizebox{0.65\textwidth}{!}{

    \includegraphics[scale=0.45,trim=0cm 7cm 18cm 0.5cm, clip]{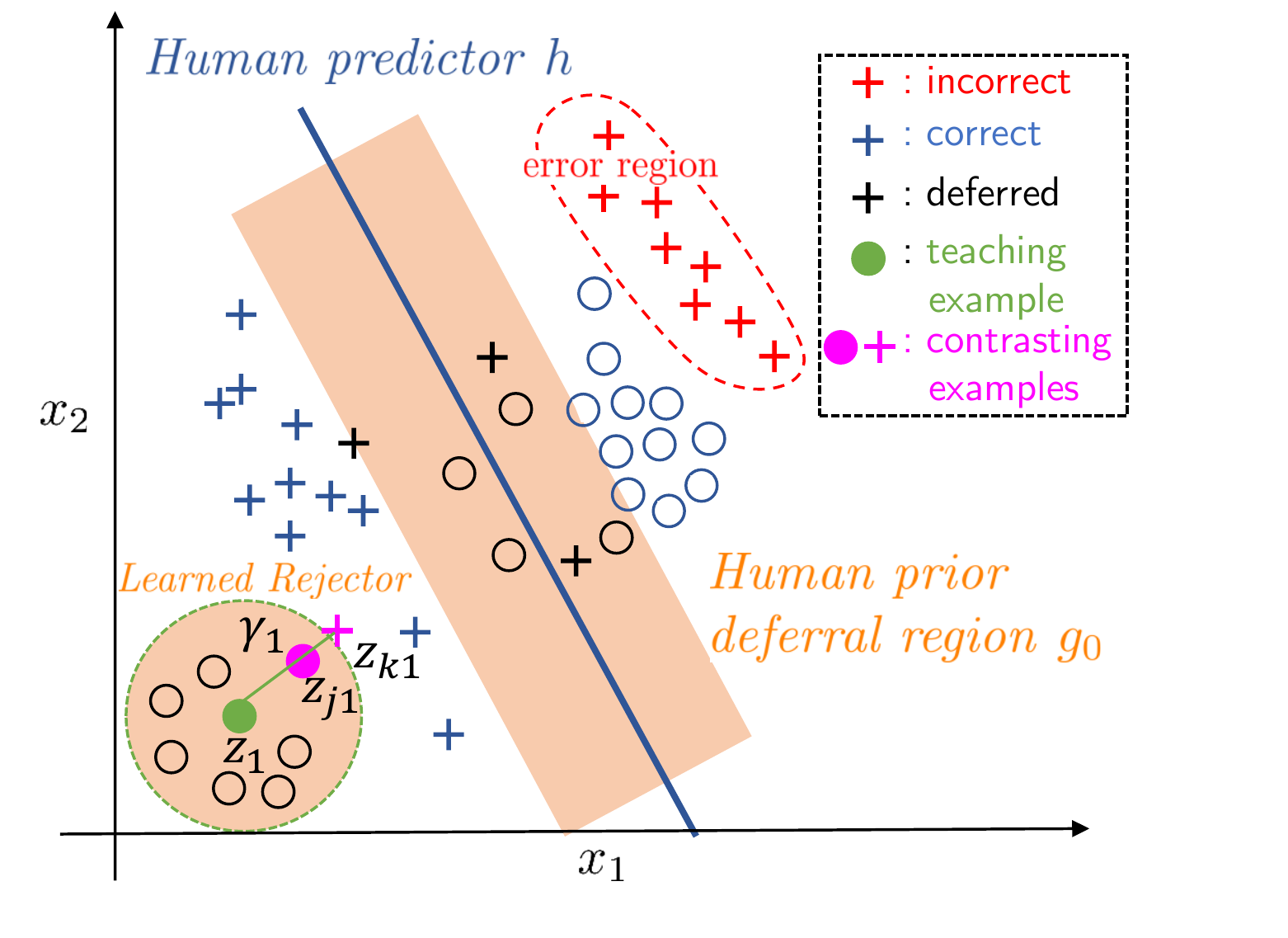}
   }
    \caption{Illustration of human rejector on toy example. The task is  classification with labels $\{o,+\}$, the human prediction $h$ is the blue line  and the prior $g_0$ is the shaded orange region surrounding the boundary. Points in red is where the human is incorrect, in blue correct and in black  point deferred to the AI. The AI is assumed to be correct on examples far from the human  boundary. The human receives a teaching example $z_1$ (in green) with radius $\gamma_1$. Also shown are the two contrasting examples $z_{j1}$ and $z_{jk}$ (in pink) that  define the region. }
    \label{fig:illustrate_rejector}
\end{figure}

\begin{assumption}[Form of Human's rejector] \label{ass:human_rejector_form}
The human's rejector consists of a prior rejector rule and a nearest neighbor rule learned after observing teaching examples $D_T=\{z_i,a_i,r_i\}_{i=1}^m$.

Formally, let $g_0(Z,A): \mathcal{Z} \times \mathcal{A} \to \{0,1\}$ be the human's prior rejector.  Figure \ref{fig:illustrate_rejector} illustrates the scenario: the prior  is the region at the boundary of the human predictor $h$. Let $K(.,.): \mathcal{Z} \times \mathcal{Z} \to \mathbb{R}^+$ be the similarity measure that the human employs to measure the degree of similarity between two instances.

The human's rejector uses a learned rule if they had  observed an example similar with respect to $K(.,.)$ during teaching, otherwise falling back on their prior:
\begin{equation}
r(Z, A; \theta_r) = \begin{cases}
\textrm{vote}(B(Z))\quad , \ if \ B(Z) \neq \emptyset \\
g_0(Z, A) \quad, \text{ otherwise}
\end{cases}
\end{equation}
where $B(Z)$ is the set of all points in $D_T$ that they observed in training sufficiently similar to $Z$:
\begin{equation}
       B(Z) = \{ i \in [m] \ | \ K(Z,z_i) > \gamma_i\}
     \end{equation}
The degree of similarity is measured by a scalar $\gamma_i$ that the human sets for each teaching example, in figure \ref{fig:illustrate_rejector} all the points in the shaded ball have $B(Z) = \{z_1\}$. The rule $\textrm{vote}(B(Z))$ defines the label for all points similar to $Z$ based on a weighted decision: \begin{equation}
\textrm{vote}(B(Z)) = \arg \max_{k \in \{0,1\}} \frac{\sum_{i \in B(Z)} \mathbb{I}\{r_i = k\} K(Z,z_i)  }{\sum_{i \in B(Z)}  K(Z,z_i) }
     \end{equation}
Where $r_i$ is the deferral rule that the human has learned on example $z_i$.

We can possibly further assume that the prior takes a rather simple form of thresholding the predictor's error: $g_0(Z,A) = \bI\{ \bP(h(Z,A) \neq Y |Z,A) \geq \epsilon \}$ for some $\epsilon >0$. One possibility for $\epsilon$ is the error rate of the AI. 

\end{assumption}

\paragraph{Discussion on the Assumptions.} In our assumptions above, we assumed knowledge of the following parameters: the human predictor $h(Z,A)$, the prior human rejector $g_0(Z,A)$ and the human similarity measure  $K(,,.)$. In fact, as we will see, we only need to know the expert error distribution $\bE[l(h(Z,A), Y)|Z,A]$ rather than the full expert predictor; it may be reasonable to estimate the expert's error distribution from previously collected data.  The prior rejector $g_0$ can also be learned by testing the human prior as evidenced by prior work on capturing human priors \cite{kim2019bayesian,bourgin2019cognitive}, otherwise, a reasonable guess is the human deferring by just thresholding their own error rate.
Finally to teach the human, we need a proxy for the similarity measure $K(.,.)$. This can be obtained in many ways: one can learn this metric with separate interactions with the human, see \cite{ilvento2019metric,qi2009efficient}, or rely on an AI based similarity measure e.g. from neural network embeddings \cite{reimers2019sentence}. This last proxy is readily available and in the framework of our study, we believe it is reasonable to use.

An important part of the rejector is the associated radius $\gamma_i$ with each teaching example $i$, the radius allows the human to generalize from each teaching example to the entire domain. The human learning process leaves the setting of $\gamma_i$ completely up to the human and is not observed. However, we  hope to directly influence the value of $\gamma_i$ that the human sets during teaching.

\section{Teaching a Student Learner}\label{sec:approach}
\paragraph{Formulation.} The previous section introduced  the model of the human learner, in this section we will set out our approach to select the teaching examples for the onboarding stage. Essentially, our approach is trying to find local regions, balls with respect to $K(.,.)$, that best teach the human about the AI. 
We assume access to a labeled dataset $S = \{x_i,z_i,y_i\}_{i=1}^n$ that is independent from the training data of the AI model.
For each point we can assign  a  deferral decision $r_i$ that the human should undertake that minimizes the system loss. Explicitly, the optimal deferral decision $r_i$ is defined to select who between the human and AI has lower loss on example $i$:
\begin{equation}
r_i = \bI \{ \bE[ l(h(z_i,a_i), y_i)] \geq \bE[l(\pi_Y(x_i),y_i]]    \}
\end{equation}
Note that to derive $r_i$ we only need to know the loss of the human on the teaching set and not their predictions. 
Define then $S^* = \{x_i,z_i,r_i\}_{i=1}^n$ as a set of examples alongside deferral decisions.
As mentioned previously, the human is also learning a radius $\gamma_i$ with each example. The radius $\gamma_i$ should be set large enough to enable generalization to the  domain, but small enough for the region to be coherent so that the human can interpret why should they follow the optimal deferral decision. 

Let $D_z \subset S^*$ and let  $D_\gamma$ be the set of radiuses associated with each point in $D_z$ and define $D=(D_z, D_\gamma)$. Define the loss of the human learner $M(.,.;D)$  now only parameterized by the teaching set $D$ as follows:
\begin{equation}
    L(D) = \sum_{i \in S} l\left(M(z_i,a_i;D),y_i\right) \label{eq:original_human_loss_D}
\end{equation}
 \paragraph{Greedy Selection.} Note that since the radiuses set by the human are learned only after observing the example, we try to jointly optimize for the teaching point and the radius to teach.
To optimize for $D$, consider the following greedy algorithm (\texttt{GREEDY-SELECT}) which starts with  an empty  set $D_0$ , and then repeats the following step for $t=1,\cdots,m$ to select the example $z$ and radius $\gamma$ that leads to the biggest reduction of loss if added to the teaching set:
\begin{align}
z, \gamma =& \arg \min_{z_i \in S \setminus D_t, \gamma }  L(D_t \cup \{z_i, \gamma\}),    \\
& \text{s.t.} \ \exists k \in [n] \ s.t. \ \gamma = K(z_i, z_k), \label{eq:gamma_in_data} \\
& \text{and} \  \frac{\sum_{j \in [n], K(z_i,z_j) > \gamma} \bI_{r_j = r_i}}{ |\{j \in [n], K(z_i,z_j) > \gamma \}|} \geq \alpha  \label{eq:alpha_constraint}
\end{align}
Constraint \eqref{eq:gamma_in_data} restricts $\gamma$ to be the similarity between $z$ and another data point and constraint \eqref{eq:alpha_constraint} ensures that $\alpha$\% of  all points inside the ball centered at $z$   share the same deferral decision as $z$. The scalar $\alpha$ is a hyperparameter that controls the consistency of the local region: when $\alpha=1$,  the region is perfectly consistent and we call this setting \texttt{CONSISTENT-RADIUS}, and when $\alpha=0$ the constraint is void and we dub the algorithm as \texttt{DOUBLE-GREEDY}.

\paragraph{Contrasting examples.} Note that the radius $\gamma$ is actually defined by two points: the point $z_k$ in equation \eqref{eq:gamma_in_data} that defines the boundary and an interior point $z_j$ that is the least similar point to $z$ with similarity at least $\gamma$; these two points are illustrated in Figure \ref{fig:illustrate_rejector} with the color pink. These two points must actually share opposing deferral actions with $r_k \neq r_j$ and thus are  contrasting points  later used as a way to describe the local region. 

\paragraph{Theoretical Guarantees.} Let $D_t$ be the solution found by the greedy algorithm and $D^*$ the optimal solution. We now try to see how we can compare $D_t$ to $D^*$. 
To do so, we  make a further assumption on the choice of radiuses that the human sets.
\begin{assumption}[Radius consistency] \label{as:gamma_and_kernnel}
We assume that if $j \in B(z_i) \cap S$ then $r_i = r_j$. This implies that if $z_j$ is at least $\gamma_j$  close to $z_i$, then the best deferral choice for $j$ is the same as that for $i$. This assumption is an assumption on the choice of $\gamma_i$'s for each example in the teaching set.
\end{assumption}

 Assumption  \ref{as:gamma_and_kernnel} in essence says that the human is always conservative enough such that the lesson drawn from example $i$ is consistent on $S$. This translates to  setting $\alpha =1$ in our algorithm; when $\alpha <1$ the guarantees may not hold. 
 Given this assumption we can deduce that our objective function is now submodular and monotone.
Furthermore, equipped with the fact that our problem is submodular we can  derive the following guarantee on the gap of performance of our algorithm versus the optimal teaching set, as the next theorem demonstrates.

\begin{theorem}\label{th:greedy_guarantee}
Let $F(X)= L(\emptyset) - L(X)$, when $\alpha =1$, $F(.)$ is submodular, monotone and positive. Moreover, the \texttt{GREEDY-SELECT} algorithm described above achieves the following performance compared to the optimal  set $D^*$:
\begin{equation}
    \underbrace{L(D_m)}_{\textrm{loss of chosen set}} \leq (1-\frac{1}{e}) \underbrace{L(D^*)}_{\textrm{loss of optimal set}} + \frac{1}{e} \underbrace{L(\emptyset)}_{\textrm{loss of prior rejector}} \nonumber
\end{equation}
\end{theorem}
All proofs can be found in Appendix \ref{apx:proofs}.

Theorem \ref{th:greedy_guarantee} gives a guarantee on the subset chosen by the greedy algorithm with an $1-\frac{1}{e}$ approximation factor, one can ask if we can do better. We prove that a generalization of our problem  is in fact NP-hard in the appendix. 
In what was previously discussed, the dataset that we measure performance on and that we teach from are the same.
We generalize to have a separate training set $S_T$ and a validation set $S_V$ and define the loss of the human with respect to $S_V$
and now define our optimization problem in terms of finding a minimal size subset $D$ that achieves a certain loss $\delta \geq 0$:
\begin{equation}
D^*_\delta = \arg \min_{D \subset S_T} |D| \quad s.t. \  \sum_{i \in S_V} l\left(M(z_i,a_i;D),y_i\right)\leq \delta \label{eq:optimal_subset_prob_delta}
\end{equation}
\begin{proposition}
Problem \eqref{eq:optimal_subset_prob_delta} is NP-hard.
\end{proposition}
The reduction is to the set cover problem and can be found in Appendix \ref{apx:proofs}.

\begin{algorithm}[tb]
\caption{Our Human Teaching Approach}
\label{alg:algorithm}
\textbf{Input}: Teaching set $D$
\begin{algorithmic}[1] 
\FOR{$i=1,\cdots,n$}
\STATE \textbf{Stage 1: Testing.} Test the human on example $z_i$ with AI message $a_i$
\STATE \textbf{Stage 2: Feedback.} Show human feedback of actual label $y_i$, AI prediction $\pi_i$, and recommended deferral action $r_i$
\STATE \textbf{Stage 3: Lesson Generalization.}  Show  the two contrasting examples $z_{j}$ and $z_{k}$ and high level features about the region to allow generalization around $z_i$.
\STATE \textbf{Stage 4: Lesson Reinforcement.} We ask the human to write a rule $R_i$ that describes the region surrounding the example $z_i$ and which action they should take.
\ENDFOR
\end{algorithmic}
\label{alg:teaching_approach}
\end{algorithm}

\paragraph{Human Teaching Approach.} After running our greedy algorithm, we obtain a teaching set $D$ that we now need to teach to the human. We rely on a four stage approach for teaching  on each  example so that they are able to learn and generalize to the neighborhood around it shown in Algorithm \ref{alg:teaching_approach}. The human first predicts on the example $z$, then they receive feedback on their prediction and the AI's prediction.
We then show them a description of the region around the example that helps them learn the radius. Specifically, we show them the two contrasting examples $z_j$ and $z_k$ defined by $\gamma_i$ and high level features about the neighborhood. 
 Finally, we ask them to formalize in writing a rule describing the region and the action to take inside that region. This rule that they write per example helps the human in creating a set of guidelines to remember for when to rely on the AI and ensures that they reflect on the teaching material.

\section{Experimental User Study}\label{sec:experiments}
We provide code to reproduce our experiments \footnote{\url{https://github.com/clinicalml/teaching-to-understand-ai}}.  Additional experimental details and results are left to Appendix \ref{apx:crowd_experiments}.

\subsection{Experimental Preliminaries}\label{sec:exp_prelim}
\paragraph{Experimental Task and Dataset.}
Our focus will be on \emph{passage-based question answering} tasks. These are akin to numerous real world applications such as customer service, virtual assistants and information retrieval. It is of interest as  relying on an AI can reduce the time one needs to answer questions by not reading the entire passage and as an experimental setup it allows a greater range in the type of \emph{sub-expertise} we can allow for compared to  experimental tasks in the literature.
We rely on the HotpotQA  dataset \cite{yang2018hotpotqa} collected by crowdsourcing based on Wikipedia articles. We slightly modify the HotpotQA examples for our experiment by removing at random a supporting sentence from the two paragraphs. The supporting sentence removed does not contain the answer, so that each question always has an answer in the passage, however, it may not always be possible to arrive at that answer. This was done to make the task harder and create incentives for expert humans to use the AI. We further remove yes/no questions from the dataset and only consider hard multi hop questions from the train set of 14631 examples and the dev set of 6947 examples.

\paragraph{Simulated AI.} One of the top performing models on HotpotQA is SAE-large: a graph neural network on top of 
RoBERTa embeddings \cite{tu2020select}. We performed a detailed error analysis in Appendix \ref{apx:ai_error_region} of the SAE-large model predictions on the dev set. However, our analysis  uncovered only few and small regions of model error. For our experimental study, we want to  evaluate the effect of teaching in two ways: 1) through systematically checking the validity of the user lessons and 2) through objective task metrics. The SAE model makes it harder for us to do both especially with a limited number of responses from crowdworkers. For this reason, we decided to create a simulated AI whose error regions are more interpretable. We first  cluster the dataset using K-means with $k_p$ clusters based on only the paragraph embeddings obtained from a pre-trained SentenceBERT  model \cite{reimers2019sentence}.  
The simulated AI model is parameterized by a vector  $err_p \in [0,1]^{k_p}$ where
the probability of error of the AI on cluster $i$ by $err_p[i]$. The answer of the AI when it is incorrect is manually constructed  to be reasonably incorrect: for example if the answer asks for a date, we provide an incorrect date rather than a random sentence. 
To summarize, the AI for each cluster in the data has a specified probability of error that is constant on the cluster. 
To show that each cluster computed has a distinct meaningful theme, we retrieve the top 10 most common Wikipedia categories in each cluster. The full categories are shown in Appendix \ref{apx:crowd_experiments}; example cluster categories include singers/musicians, movies and soccer (but not football).

\paragraph{Metrics.} Our aim will be to measure objective task performance and effort through the proxy of time spent on average per example. Our task  performance metric is the F1 score on the token level \cite{rajpurkar2016squad}; we will measure this when considering the final predictions (Overall F1), on only when  the human defers (Defer F1) and when the human does not defer (Non-Defer F1). We will also measure \emph{AI-reliance}:
this is calculated as how often they rely on the "Let AI answer for you" button in Figure \ref{fig:test_interface}.
\subsection{Simulated Users}

\begin{figure}[h] 
    \centering

    \subfloat[Setting B and \texttt{CONSISTENT-RADIUS} ]{%
        \includegraphics[trim={0 0 0 2cm},scale=0.55]{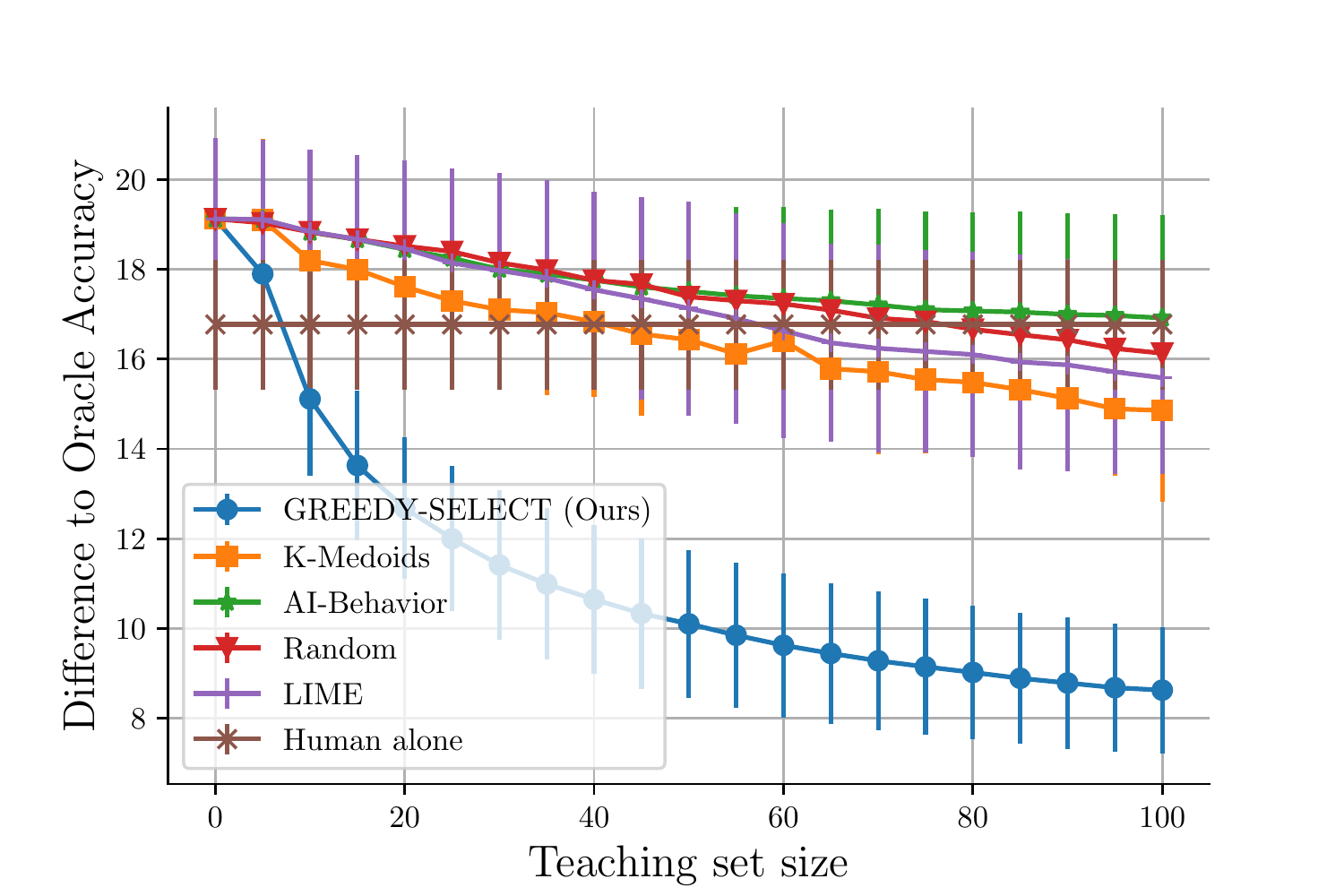}%
        \label{fig:hotpot_c_b}%
        }%
    \subfloat[Setting A and \texttt{GREEDY-RADIUS} ]{%
          \includegraphics[trim={0 0 0 2cm},scale=0.55]{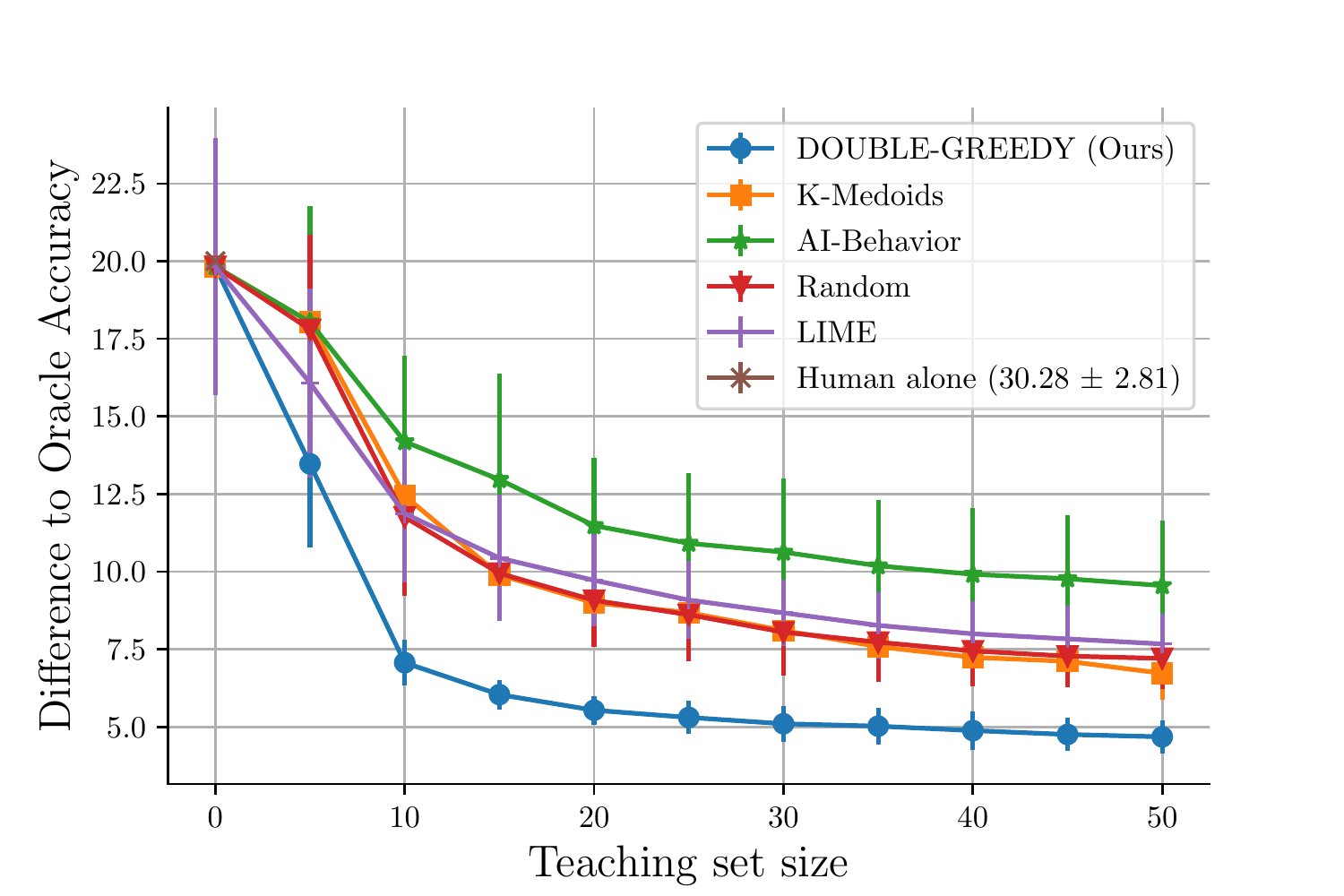}%
        \label{fig:hotpot_g_a}%
        }%

           \vspace{1cm}
          \subfloat[Setting A and \texttt{CONSISTENT-RADIUS} ]{%
        \includegraphics[trim={0 0 0 2cm},scale=0.55]{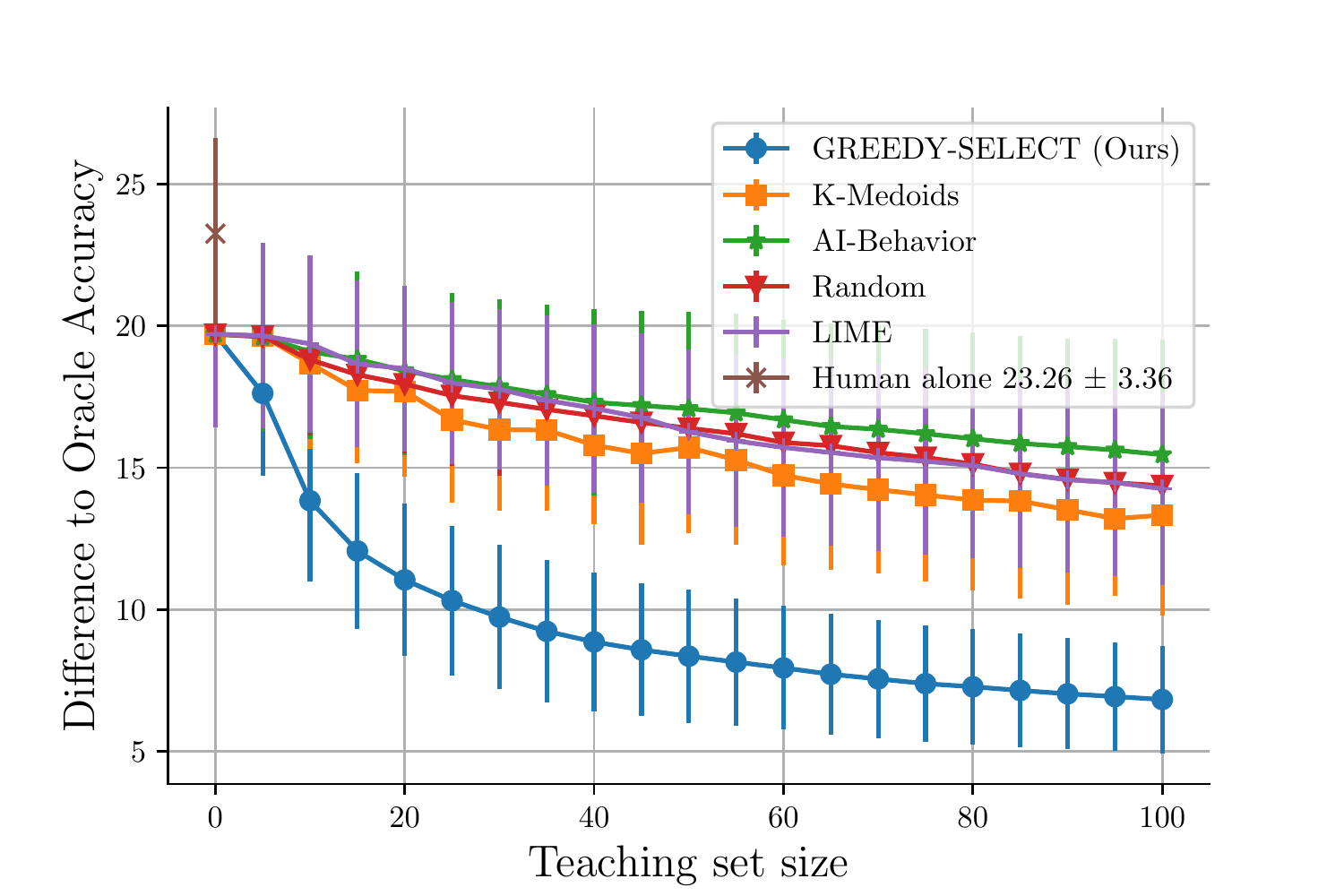}%
        \label{fig:hotpot_c_a}%
        }%
    \subfloat[Setting B and \texttt{GREEDY-RADIUS} ]{%
          \includegraphics[trim={0 0 0 2cm},scale=0.55]{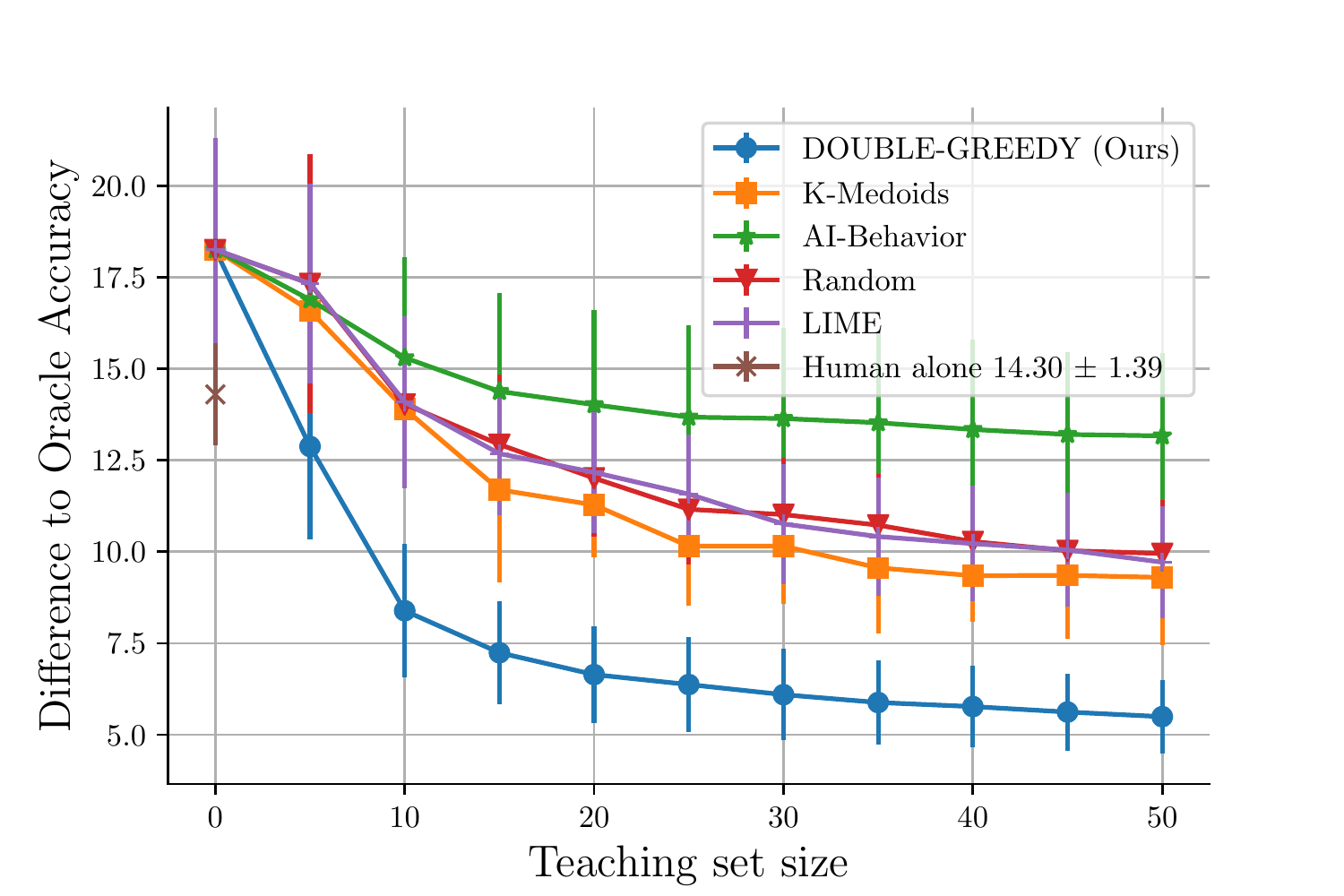}%
        \label{fig:hotpot_g_b}%
        }%
        

    \caption{Teaching set size versus the negative difference between the human's learner test accuracy under the different methods compared to ORACLE. We plot the average result across 10 trials and standard deviation as error bars.}
      \label{fig:hotpot_teach_cmplx}
\end{figure}

Before we experiment with real human users, we evaluate the teaching complexity, i.e. the relation between teaching set size and human accuracy, of our teaching algorithm on simulated human learners that follow our assumptions. We further evaluate the robustness of our approach when we do not have full knowledge of the human parameters.  

\paragraph{AI and Human model.} We use the simulated AI model with $k_p = 15$  and a  vector of errors  $err_p$ where for each $i$, $err_p[i]$ is drawn \textit{i.i.d.} from $\textrm{Beta}(\alpha_{ai},\beta_{ai})$.
 The human predictor is analogous to the AI model  with a different vector of probabilities $err_p'$ sampled from $\textrm{Beta}(\alpha_{h},\beta_{h})$.
The human prior  thresholds the probability error of the human to a constant $\epsilon$. Finally, the human similarity measure is the RBF kernel on the passage embeddings i.e. $K(x,x')= e^{-|x-x|^2}$. In this setup both the human and AI contexts are identical and the AI does not send any messages to the human.

\paragraph{Baselines.} We implement a domain cover subset selection baseline in K-Medoids, the  LIME selection strategy by \cite{ribeiro2016should} with 10 features per example following \cite{lai2020chicago} (LIME), random selection baseline (RANDOM) and a    baseline that greedily selects the point that helps a 1-nearest neighbors learner best predict the AI errors (AI-BEHAVIOR). 
Finally, we also compare to the optimal rejection rule computed with knowledge of the human and AI error rates  by picking the lower one (ORACLE). The ORACLE rejector is an upper bound on achievable performance by any possible rejector regardless of the human student model. 

\paragraph{Experimental setup.} We will compare to the baselines as we vary the size of the teaching set $D_T$.  To illustrate the effectiveness of the teaching methods, we focus on two settings: A) the Human is less accurate than the AI but their prior rejector rarely defers where we set the following  and B) the Human is more accurate than the AI but their prior rejector over defers to the AI.
These two settings is where teaching is most beneficial as the prior is erroneous.
Specifically in setting A)  we set the following: $(\alpha_{ai} = 2, \beta_{ai}=1)$ (the pdf is a straight line from the origin to $(1,2)$),   $(\alpha_{h} = 1, \beta_{h}=1)$ (uniform distribution) and $\epsilon = 0.1$  and B) we set $(\alpha_{ai} = 1, \beta_{ai}=1), (\alpha_{h} = 2, \beta_{h}=1)$ and $\epsilon = 0.9$.
We evaluate for each setting 10 different random settings of the human and AI error probability vectors and average the results. 

\begin{figure}

\centering
\begin{tabular}{lr}
\toprule
Condition &  Oracle Gap @n=30  \\
\midrule
Full Information & 6.38 $\pm$  1.56 \\
Missing $g_0$ & 6.90 $\pm$ 1.80  \\
Noisy Radius & 9.74 $\pm$ 3.0 \\
Missing $h$ & 13.47 $\pm$ 5.07 \\
No Information+Noise&  15.12 $\pm$ 4.00\\
\midrule
Prior only & 16.72 $\pm$ 1.22 \\
Human Alone & 19.8 $\pm$ 2.80 \\
\bottomrule
\end{tabular}
        \captionof{table}{Test Accuracy gap between \texttt{DOUBLE-GREEDY} and ORACLE at teaching set of size 30 under various conditions. This is performed under setting B.}
    \label{fig:hotpot_noise}
\end{figure}

\paragraph{Results.} Figure \ref{fig:hotpot_teach_cmplx} shows the gap between Oracle and  human accuracy on the dev set compared to the size of the teaching set for each of the methods. We can see that our approach is able to outperform the baselines  under  setting $B$ with \texttt{CONSISTENT-RADIUS}.  
We observe a wide gap between our method and the baselines, this is because the teaching examples here must focus on only a select number of the clusters and cover them sufficiently. 
 With the greedy radius selection, we require fewer  examples to reach high accuracy and the gap between our method and the baselines narrows.


\paragraph{Robustness to Misspecification of Human model.} We evaluate accuracy when the human is not learning the correct radius; this simulates noise in the learning process. The radius $\gamma_i$ that the human learns is a noisy version of $\hat{\gamma_i}$ where we add a uniformly distributed noise $\delta \sim \mathcal{U}(-(1-\hat{\gamma_i})/2,(1-\hat{\gamma_i})/2)$ to it.
We then evaluate when we have no knowledge of the prior rejector $g_0$ or/and  no knowledge of the human predictor $h$.
When we don't know either of these parameters, we replace them by a random binary vector $\text{Bernoulli}(1/2)^n$ on the teaching set.
Results are shown in Table \ref{fig:hotpot_noise}. We can see that even if we don't have knowledge about the prior,  accuracy is not impacted. However, if we don't have knowledge about the predictor $h$, then  performance drops significantly.  To evaluate how much information about $h$ we need to properly teach the human, we learn a teaching set assuming the human's error probability is $err_p' + \bm{\delta}$ where $\bm{\delta}$ has each component drawn from $\{-\delta,\delta\}$ uniformly where $\delta >0$. On setting $B$ with \texttt{DOUBLE-GREEDY}, we can tolerate up to $0.25$ error in knowledge about cluster error probability with no noticeable drop in performance; full results are in Appendix \ref{apx:synth_experiments}.
Note that when we don't have any knowledge about the human and the learning process is noisy, teaching is  impacted.

\subsection{Crowdsourced Experiments Details}\label{sec:crowd_details}
\begin{figure*}[t]
    \centering
    \resizebox{\textwidth}{!}{
        \subfloat[Testing interface ]{%
          \includegraphics[width=7.7cm,height=8.5cm,trim=5.8cm 1.5cm 9cm 0cm, clip]{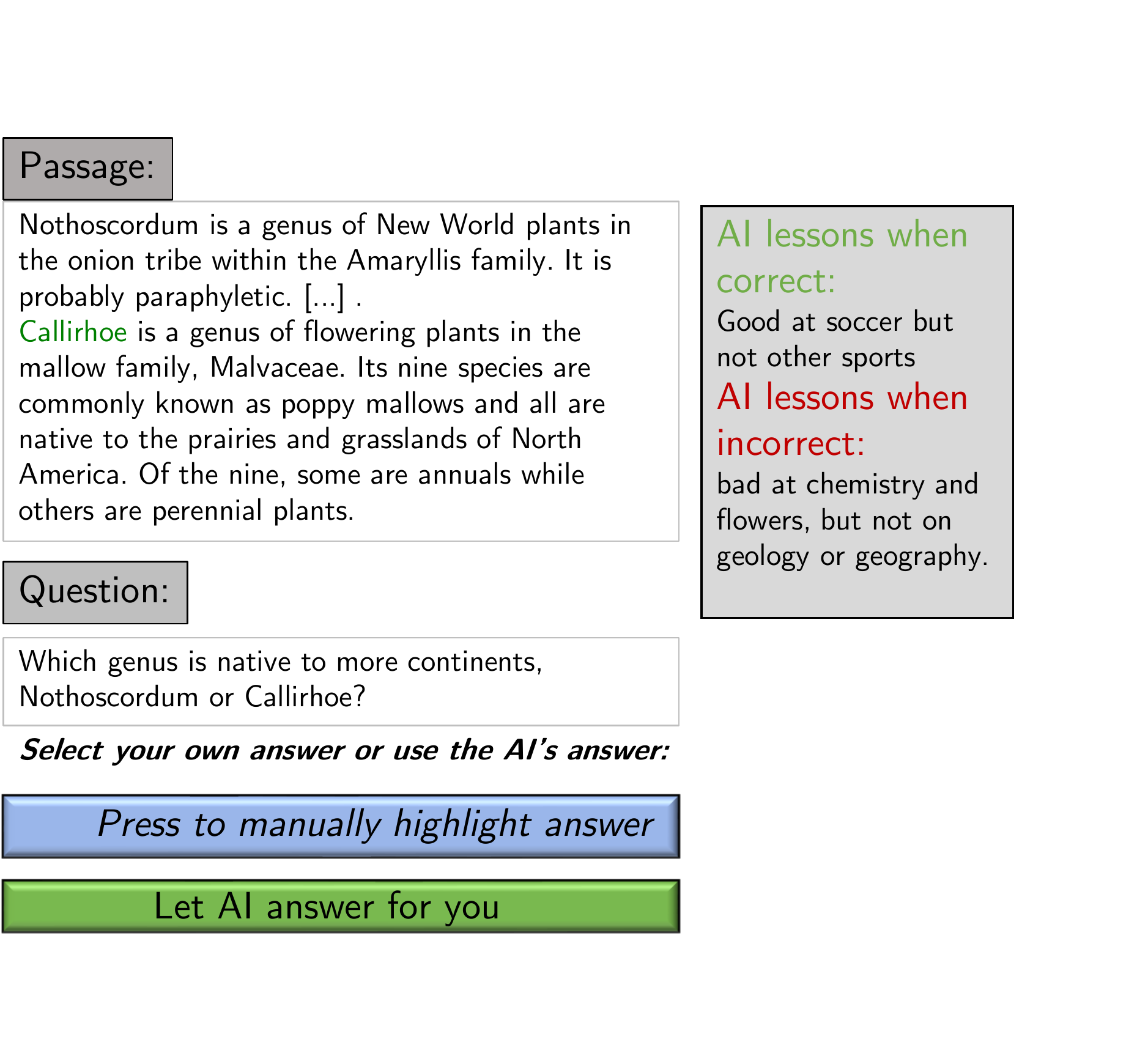}%
        \label{fig:test_interface}%
        }%
    \subfloat[Teaching interface ]{%
        \includegraphics[ width=8.7cm,height=8.5cm,trim=6cm 0.8cm 4cm 0cm, clip]{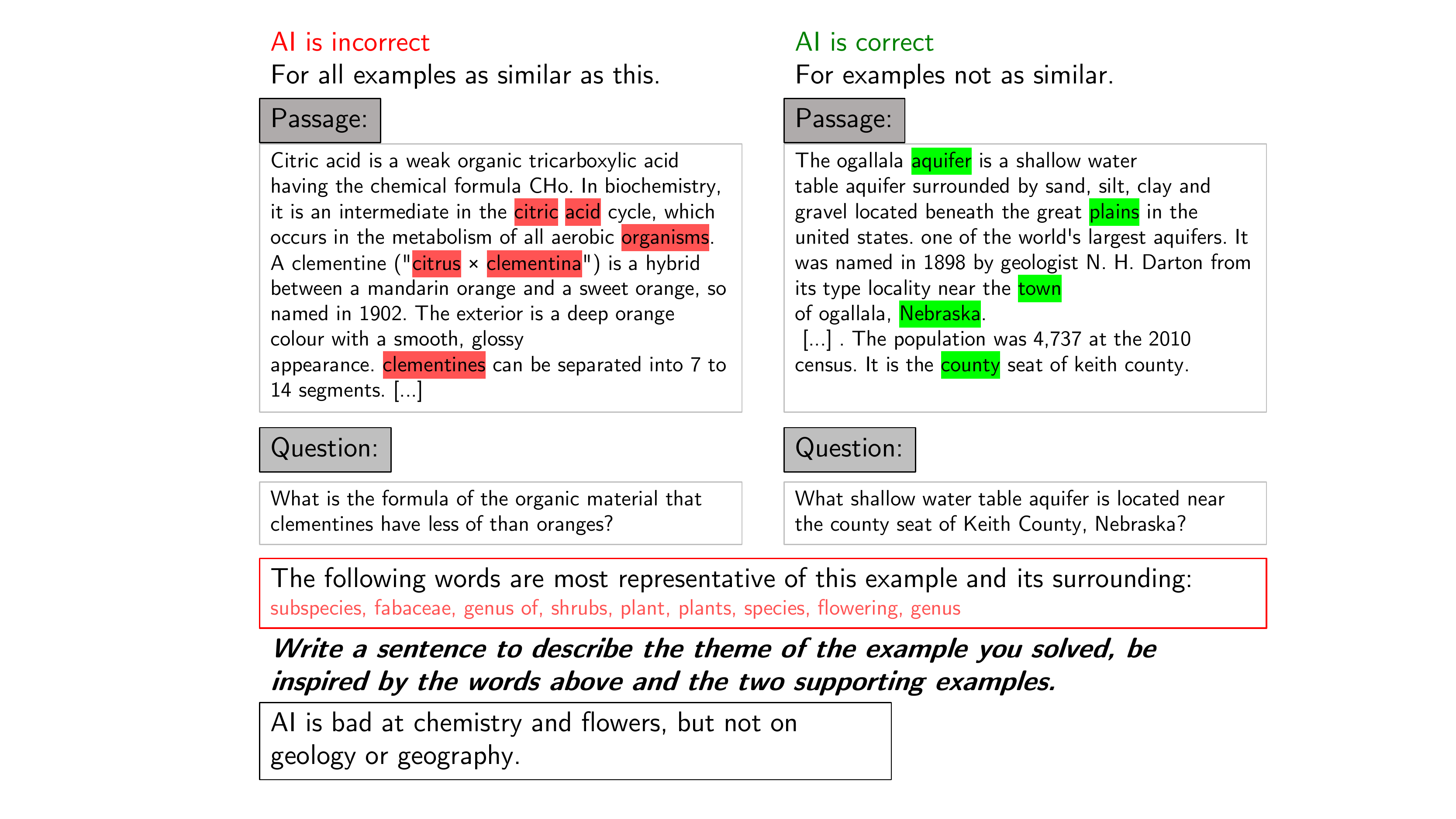}%
        \label{fig:teach_interface}%
        }%
}
    \caption{On the left in subfigure (a) is the testing interface shown for an example. This is the same interface that is also shown at the beginning of each teaching example. After the human predicts and we  are in the teaching phase, we show them the correct answer and transition to the interface in subfigure (b) that shows the two supporting examples for the example in (a), the top weighted words in the region and asks the user to write down their rule for the example.}
      \label{fig:interface_test}
\end{figure*}
\begin{table*}
\small
\
\centering
    \resizebox{\textwidth}{!}{
\begin{tabular}{lrrr|rrrr}
\toprule
Metric & Ours-Teaching (all) &  No-Teaching  & LIME (all) & Ours (acc)  & Ours (inacc) & LIME (acc)  & LIME (inacc)  \\
\midrule
Overall F1 & 58.2 $\pm$ 3.4 & 57.6 $\pm$ 3.4 & 52.9 $\pm$ 3.4  &  62.8 $\pm$ 4.7 & 53.5 $\pm$ 4.9   &  56.5 $\pm$ 6.4 & 52.0 $\pm$ 4.2\\
Defer F1 &50.7 $\pm$ 4.7  & 57.8 $\pm$ 4.9 & 48.1 $\pm$ 5.3 & 53.4 $\pm$ 6.7 & 50.0 $\pm$ 6.8   &  44.6 $\pm$ 9.0 & 49.9 $\pm$ 6.5\\
Non-Defer F1 &67.6 $\pm$ 4.7  & 57.6 $\pm$ 4.7 & 56.9 $\pm$ 4.6 & 73.92$\pm$ 6.2 & 60.6 $\pm$ 7.1   &  70.0 $\pm$ 8.6 & 53.7 $\pm$ 5.4\\
Time/ex (min) & 0.60 $\pm$ 0.03 & 0.62 $\pm$ 0.03 & 0.68 $\pm$ 0.04  & 0.54 $\pm$ 0.04 & 0.68 $\pm$ 0.05   &  0.65 $\pm$ 0.08 & 0.69 $\pm$ 0.05\\
AI-Reliance (\%) &55.2 $\pm$ 3.6  & 48.9 $\pm$ 3.6 & 45.4 $\pm$ 3.6 & 53.3 $\pm$ 4.9 & 58.9 $\pm$ 5.0   &  52.8 $\pm$ 3.6 & 43.6 $\pm$ 4.3\\
\bottomrule
\end{tabular}
}
\caption{Comparison of the metrics between our teaching condition (split into all participants, those who gave accurate lessons (acc) and those who didn't (inacc), see description below), the \texttt{No-teaching+AI-prediction} condition and LIME teaching. Shown are averages across all participants with 95\% confidence interval error bars. The F1 of the AI alone in this setting is 46.7\%; we did not separately measure the F1 of the human in isolation. }
\label{fig:turk_results}
\end{table*}

\paragraph{Testing user interface.} Our user interface during testing is shown in Figure \ref{fig:test_interface} which shows a paragraph and its associated question. The human can either submit their own answer or let the AI answer for them using a special button.
However, the interface does not display the AI's answer or any explanation, which forces the user to rely solely on their mental model and the teaching examples to make a prediction. This was done so that we can control for the effect of teaching solely, as showing the AI prediction at test time leaks information about the AI beyond what was shown in the teaching set. Moreover, not showing the AI prediction forces the human to explicitly think about the AI performance. The right panel next to the passage shows the lessons that the user wrote down during teaching.

\paragraph{Teaching  user interface.} Following our teaching algorithm, during teaching, the worker is first faced with the same user interface as in test time. The difference is that {\em after} they answer, they receive feedback on the correctness of their answer and can see the AI's answer.
We then show the human the two constrasting examples with LIME word highlights. As a high level description of the local region, we show the top 10 most weighted words obtained by LIME in the ball surrounding the original teaching example \cite{ribeiro2016should} (see Figure \ref{fig:teach_interface}).  After they observe the two supporting examples, they are  asked to write a sentence that describes the lesson of the example. These sentences are available during test-time for the workers to review as help for answering new questions. 

\paragraph{Experimental Design and Baselines.} The experimental teaching setup proceeds in three stages.
The first stage (Stage 0) is a tutorial that introduces the task with two examples and where we gather the worker's demographic information, knowledge of machine learning and how often they visit Wikipedia. Stage 1 is the teaching stage where the worker solves 9 teaching examples and stage 2 is the testing phase where the worker solves 15 questions with no feedback. After the two stages is an exit survey where users are asked about their decision process for using the AI.  The two stage experimental design mimics what we believe would be a realistic deployment in practice; we don't expect feedback to be possible during deployment, but rather only in a specialized teaching phase. We randomly assign each participant to one of three conditions.

In the first condition the participants go through the entire pipeline described above (\texttt{Ours Teaching}).
 The second is condition is called (\texttt{LIME-Teaching}) where
 LIME is first used to obtain 18 examples. During teaching, users are asked to solve the first 9 questions and are then shown: LIME highlights of the example, performance feedback and asked to write a lesson of what they learned. Then users view the 9 remaining examples with LIME highlights without needing to solve them or write lessons. The difference with our method is that workers don't see the supporting examples or the word level description of the regions.
The third is 
a baseline condition (\texttt{No-teaching+AI-prediction}) that makes the following modifications to the experimental design: the participants skip the teaching stage (Stage 1) and immediately proceed to the testing phase (Stage 2). However, during the testing phase, the participants \emph{can see the AI prediction} before they press the use AI button which gives them an edge compared to the teaching condition.

\paragraph{Participants} We recruited 50 US based participants from Amazon Mechanical Turk per each condition (150 total) and initial pilot studies were also conducted with graduate students in computer science at a US university. 
Participants in the non-teaching baseline were paid \$3 for 10 minutes of work and those in the teaching condition received \$6 for 20 minutes of work. Any demographic information we gathered in our study is kept confidential and workers were asked to consent to their use of their responses in research studies.

\paragraph{AI and Test Set details.}
The simulated AI had $k_p = 11$ and was randomly chosen to have probability of error 0 or 1 on each cluster. This means there are clusters where the AI is perfect on and other clusters where the AI is always wrong.  We  split the HotpotQA dev set into two parts 80:20 for the teaching and testing set respectively.
To obtain the 9 teaching examples we run \texttt{GREEDY-SELECT} with the consistent radius strategy with no knowledge of $g_0$ or $h$. The examples  in the testing phase was obtained first by filtering the data using K-medoids with $K=200$ as a way to get diverse questions. Then each participant received 7 random  questions from the filtered set on which the AI was correct and 8 on which the AI is incorrect. 

Further details can be found in Appendix \ref{apx:crowd_experiments}.

\subsection{User Study Observations and Results}\label{sec:user_study_observations}

\begin{table}
\small
\
\centering
    \resizebox{\textwidth}{!}{
\begin{tabular}{lrr|rr}
\toprule
Metric & Ours-Teaching (ID) &  No-Teaching (ID)  & Ours (OOD)  & No-Teaching (OOD)   \\
\midrule
Overall F1 & 56.8 $\pm$ 3.6 & 56.0 $\pm$ 3.6 &  70.9 $\pm$ 10.5 & 72.86 $\pm$ 10.7  \\
Defer F1 &51.42 $\pm$ 4.9  & 57.8 $\pm$ 5.2  & 42.95 $\pm$ 17.2 & 56.7 $\pm$ 18.8   \\
Non-Defer F1 &63.7 $\pm$ 5.2  & 54.4 $\pm$ 5.1  & 96.05$\pm$ 5.82 & 85.0 $\pm$ 11.5   \\
AI-Reliance (\%) &56.1 $\pm$ 3.8  & 49.4 $\pm$ 3.8 & 47.3 $\pm$ 11.6 & 42.9 $\pm$ 11.8   \\
\bottomrule
\end{tabular}
}
\caption{Comparison of the metrics on clusters that were seen during teaching with our method (ID for in distribution) compared to performance on clusters that  were not seen during teaching (OOD for out of distribution). We also show the performance of the no-teaching baselines on the two cluster sets as a reference point. The errors on the OOD estimates are much higher as there are much fewer samples in the not-seen clusters. }
\label{fig:turk_results_ood}
\end{table}

 \paragraph{Teaching enables participants to better know when to predict on their own, but not when to defer to the AI.} The first three columns of Table \ref{fig:turk_results} display the metrics measured across both conditions on all participants. We can first note that participants with teaching are able to predict overall just as well as participants in the baseline no-teaching condition who have additional information about the AI prediction at test time. Moreover, participants who received teaching can better recognize when they are able to predict better than the AI. There is a difference significant at $p$-value 0.05 ($t=2.9$, from a two sample t-test) of the F1 score when the human doesn't defer between our method and the no-teaching baseline and significant at $p$-value 0.001 ($t=3.2$) compared to LIME. However, the participants in the teaching condition deferred to the AI when it was incorrect more often than those in the no-teaching baseline condition. A positive difference significant at $p$-value 0.05 ($t=-2.0$) in F1 when the humans defers for  \texttt{No-teaching+AI-prediction} workers. An explanations for this is that the participants might press the use AI button on examples where their own prediction agrees with that of the AI instead of manually selecting the answer which takes  more effort. 

\paragraph{Accurate teaching lessons might predict improved task performance and our method teaches more participants than LIME.} Given our knowledge about the clusters and the AI, the correct form of the teaching lesson of each example is "AI is good/bad at TOPIC" where TOPIC designates the theme of each cluster amongst a set of 11 topics which include soccer, politics, music and more. Manually inspecting the lessons of the 50 participants without seeing their test performance, we found that 25 out of 50 participants in our teaching condition were able to properly extract the right lesson from each teaching example. The remaining 25 participants were split into two camps: those who gave explanations on question/answer type or too broad or narrow of explanations e.g. "AI is good at people" rather than a specific subgroup of musicians for example (14 out of 50), and those who gave irrelevant explanations  (11 out of 50, this group  performed non trivially and so could not be disqualified). Table \ref{fig:user_lessons} in Appendix \ref{apx:crowd_experiments} gives examples of the actual lessons that users wrote. Results for participants who had accurate vs not accurate lessons are shown in the last four columns of Table \ref{fig:turk_results}.
The  participants who had accurate lessons had a 9 point average overall F1  difference significant  at $p$-value $0.01$ compared to those with inaccurate lessons. 
 With LIME-Teaching  we found that only 14 out of 50 participants were able to properly extract the right lessons. The difference between LIME and our method in enabling teaching is significant at $p$-value $0.02$ with $t=2.3$, however, we observe that accurate teaching has a similar effect in both conditions. Note, that even when participants have accurate lessons, they often don't always follow their own recommendations as evidenced by the low Defer F1 score.

\paragraph{Differences in performance on in-distribution and out-of-distribution examples.} During teaching with our method we let the users solve 9 examples, each corresponding to a unique cluster. The data domain is in fact split into 11 clusters where the AI has a different error probability in $\{0,1\}$ on each of them. Thus, there are 2 clusters where users have not seen examples from, which we call the out-of-distribution examples (OOD), and 9 from which they have, the in-distribution examples (ID). In table \ref{fig:turk_results_ood} we show the different metrics split into ID and OOD distribution for teaching participants in our method and for the no-teaching participants as a reference point. LIME-Teaching participants observe all the clusters during teaching so there is no distinction between ID and OOD. We can first observe a very high F1 for OOD examples where the human predicts (Non-Defer F1) for our method. This is also the case for the non-teaching participants, thus the increase in F1  lies with the nature of the examples in the OOD clusters rather than the distinction of them being ID versus OOD. On the other hand, we observe that Defer F1 is higher by 8.36 points on average for ID examples compared to OOD with our teaching method while we do not observe a difference in Defer F1 for the baseline non-teaching group.
 However, the results are not significant as the 95\% confidence intervals overlap.


\section{Additional Synthetic Experiments}

\textbf{Dataset.} To complement our NLP-based experiments, we run a study 
on the CIFAR-10 image
classification dataset \cite{krizhevsky2009learning} consisting of $32 \times32$ color
images drawn from 10  classes.  For CIFAR we use a WideResNet \cite{zagoruyko2016wide} with no data augmentation that achieves 90.46\% test accuracy and  the model is trained to minimize the cross entropy loss with respect to the target. We split the dataset into three distinct parts: training set for AI model (90\% CIFAR train, 45k), teaching set to obtain teaching images (10\% of CIFAR train set, 5k) and test set for the human learner (CIFAR test set, 10k). 
 
\textbf{Setup.} We let $X=Z$ and use the respective models' last layer encodings as the input space to the teaching algorithm. The message the AI sends is the pair $A=(\hat{y}, \hat{c})$ consisting of the AI prediction and a confidence score (softmax output of model). We assume the human is following the human rejector Assumption 2 and is perfectly learning the radius and actions. We consider the human expert models considered in \cite{mozannar2020consistent}: let  $k \in [10]$, then if the image is in the first k classes the expert is perfect, otherwise the expert predicts randomly. The human prior rejector defers if  the AI's confidence $\hat{c}$ is less than $\epsilon=0.5$.

\textbf{Results. } We show the results in Table \ref{fig:cifar_setting_1} for various teaching set sizes for the expert $k=6$ and a learning curve in Figure \ref{fig:curve_cifar_greedy}; full results are in Appendix \ref{apx:synth_experiments}. We compare our approach to solving the problem as learning to defer with the AI deferring to the human: we compare to the surrogate loss baseline in \cite{mozannar2020consistent}, the confidence baseline in \cite{raghu2019algorithmic} and a ModelConfidence baseline which optimizes over the prior parameter $\epsilon$. We find that with only 4 teaching examples, \texttt{DOUBLE-GREEDY} increases accuracy from 90.98 to 96.3 $\pm$ 0.1 on the test set.

\begin{figure}[H]
\small
\centering
\begin{tabular}{lr}
\toprule
Method &  CIFAR (acc)   \\
\midrule
Prior only & 90.98 $\pm$ 0.0   \\
 \texttt{DOUBLE-GREEDY} @T=4 &96.3 $\pm$ 0.1  \\
 \texttt{DOUBLE-GREEDY} @T=8 & 96.4 $\pm$0.1 \\
 \texttt{DOUBLE-GREEDY} @T=14 &96.5 $\pm$0.1  \\
  K-Medoids @T=4 &94.58 $\pm$0.3  \\
 K-Medoids @T=8 &95.5 $\pm$ 0.2  \\
  K-Medoids @T=14 &96.5 $\pm$ 0.2  \\

 Random @T=8 & 95.3 $\pm$ 0.5  \\
\midrule
Oracle & 97.91 \\
Surrogate Loss \cite{mozannar2020consistent}  & 97.1  \\
Confidence \cite{raghu2019algorithmic}  &95.5 \\
ModelConfidence  &93.94  \\
\bottomrule
\end{tabular}
\captionof{table}{Synthetic experiment on CIFAR-10, showing the test Accuracy for our method \texttt{DOUBLE-GREEDY} at different teaching set sizes and learning to defer baselines.  }
\label{fig:cifar_setting_1}
\end{figure}

\begin{figure}[H]
    \centering
    \includegraphics[scale=0.6]{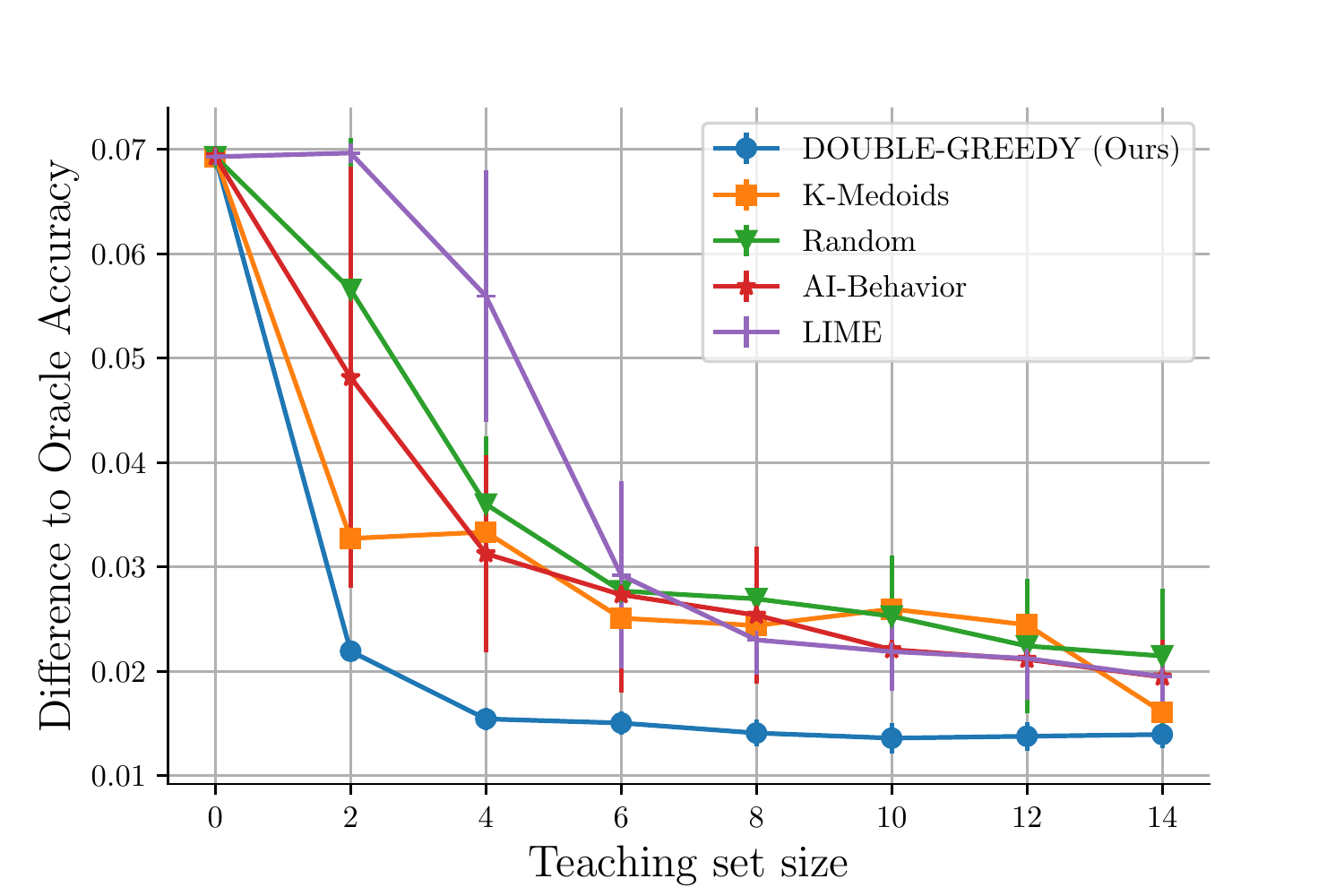}
    \caption{Synthetic experiment on CIFAR-10, showing difference between the performance of the methods and ORACLE (defined as taking the optimal decision at test time) for expert $k=6$. }
    \label{fig:curve_cifar_greedy}
\end{figure}

\section{Discussion}\label{sec:discussion}

Our work provides a general recipe for onboarding human decision makers to AI systems. We propose an exemplar based teaching strategy where humans are asked to predict on real examples and then with the help of similar examples and top features for the neighborhood, the human derives an explanation for the AI performance.

One limitation of our human
experiments is that we used a simulated AI that has an easier
to understand error boundary. This enabled us to have a
more in-depth study of the crowdworker responses than otherwise
would have been possible.
Having a simulated AI  which we perfectly understand where its error regions are, enables us to define what the "lessons" should be and thus evaluate if users are learning correctly. Future user studies will evaluate with non-simulated AI models. We hypothesize that the example selection algorithm presented in this work will be sufficient, however, we might require better methods to illustrate the neighborhood for each example. 
Another limitation is that our test-time interface did not include model explanations, which was done to eliminate additional confounding factors when comparing approaches. Future work will evaluate whether the effect of teaching remains as significant when evaluating with test-time model explanations.
Other limitations include the fact that we are using a proxy task of passage based question answering  and  proxy tasks have been documented to be  misleading for  evaluating AI systems \cite{buccinca2020proxy}. Another limitation is the use of MTurk which may not ensure high quality workers and the final limitation is that our study only focuses on the onboarding phase of AI deployment. 

Teaching is used in our work to influence human's perception of an AI model; this can be potentially used to manipulate workers into relying on AI agents in high stakes settings if the AI predictions during teaching were fabricated. While our work was conducted in a low stakes scenario and was designed to portray an accurate reflection of the AI performance, it is possible by manipulating the AI predictions during teaching to have the worker learn any desired rejector. %
We believe if the data used during teaching is not manipulated, then our approach can serve to give an unbiased overview of the AI.

\section*{Acknowledgments}

HM and DS were supported by NSF AitF award CCF-1723344.

%% file: appendix.tex
\clearpage
\appendix
\input{appendix/related_work} 
\clearpage
\input{appendix/proofs}
\clearpage
\input{appendix/ai_error_regions}

\clearpage

\section{Synthetic Experiments Details and Results}\label{apx:synth_experiments}
All experiments were run on a Linux system with a NVIDIA Tesla K80 GPU, 25 GB of RAM on Python 3.7. We use the scikit-learn package to run the clustering algorithms \cite{scikit-learn}, LIME package for the selection baseline \cite{ribeiro2016should} \footnote{\url{https://github.com/marcotcr/lime}}, ELI5 package to obtain the text LIME highlights \footnote{\url{https://eli5.readthedocs.io/en/latest/index.html}} and the Sentence Transformers package for the embedding models \cite{reimers2019sentence} \footnote{\url{https://github.com/UKPLab/sentence-transformers}}

\subsection{Misspecification results}

\begin{figure}[H]

        \centering
          \includegraphics[trim={0 0 0 0.2cm},scale=0.55]{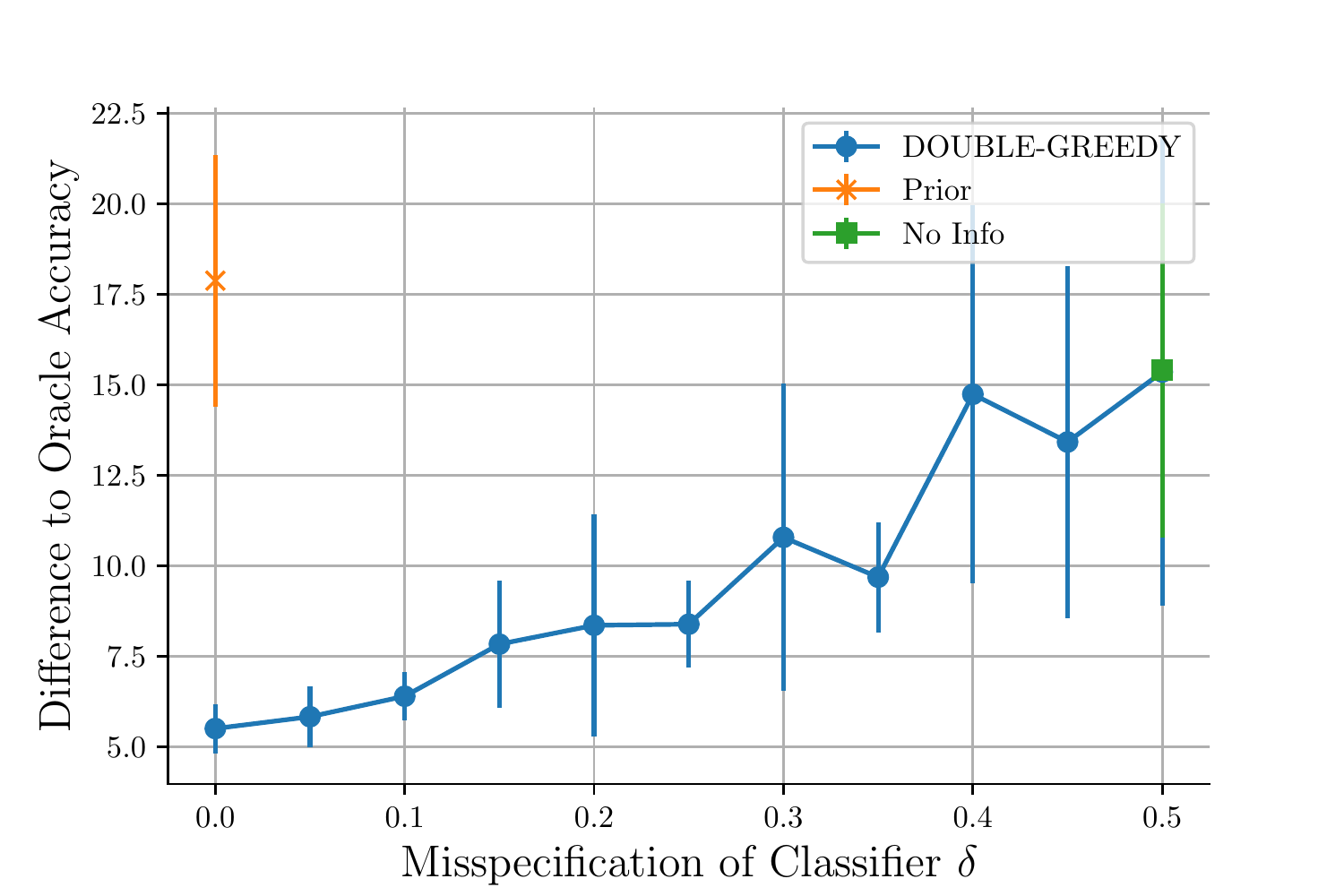}
           \caption{Difference in Oracle accuracy at teaching size @T=30 for the \texttt{DOUBLE-GREEDY} method assuming an error in $h$ by $\bm{\delta}$ in setting B. } \label{fig:noise_inh}
    \end{figure}

 To evaluate how much information about $h$ we need to properly teach the human, we learn a teaching set assuming the human's error probability is $err_p' + \bm{\delta}$ where $\bm{\delta}$ has each component drawn from $\{-\delta,\delta\}$ uniformly where $\delta >0$. 
Figure \ref{fig:noise_inh} shows the difference to ORACLE accuracy as we increase the misspecification of the human predictor. In this experiment, we assume knowledge of the prior rejector $g_0$ and that the human is perfectly learning the radius given by the teaching algorithm. What this experiment impacts is the computation of the optimal deferral decision $r_i$ computed by our algorithm to obtain $S^*$. At the limit when $\delta=0.5$, we assume that the human expert error rate is uniformly $0.5$ across the domain, which is the same as having the human predictions $h \sim Bin(1/2)$ on the teaching set.

In Table 1 in the paper, we evaluate what happens when the human is not learning the radius perfectly, this simulates noise in the learning process. The radius $\gamma_i$ that the human learns is a noisy version of $\hat{\gamma_i}$, specifically we add a uniformly distributed noise $\delta \sim \mathcal{U}(-(1-\hat{\gamma_i})/2,(1-\hat{\gamma_i})/2)$.

\section{Additional Synthetic Experiments }
\subsection{CIFAR-10}

\begin{figure}[H]
    \centering
    \includegraphics[scale=0.6]{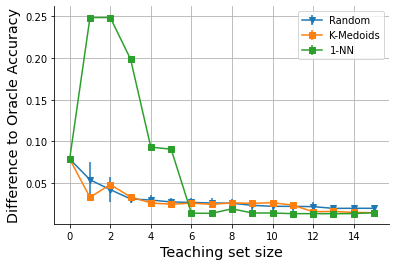}
    \caption{Comparing a 1-nearest neighbor rejector model to the radius nearest neighbor model introduced in Assumption \ref{ass:human_rejector_form} for expert $k=6$. The "1-NN" line is obtained by first obtaining $T$ points using K-medoids and then running a 1-NN rejector on these points with the label assigned to each point being the optimal deferral decision $r_i$. We can see that 1-NN struggles with less than $6$ examples, but then reaches a steady state that has the same error as the radius nearest neighbor model. The effectiveness of the radius nearest neighbor model when the teaching set is very small is due to the local nature of each update with the addition of a teaching example.       }
    \label{fig:knn_vs_radius}
\end{figure}

\begin{figure}[H]
    \centering
    \includegraphics[scale=0.6]{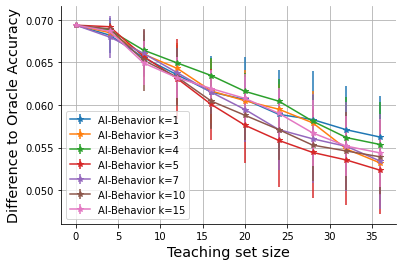}
    \caption{Performance of the AI-Behavior baseline as we vary the parameter $K$: the AI-Behavior baseline uses a $K$-nearest neighbor rejector and at each teaching step selects the point that best reduces the error of the rejector at detecting the AI's errors. We show results for the human expert $k=6$ with the consistent radius strategy $\alpha=1$. We can see that the parameter $K$ has little effect and thus we use  a natural choice of $K=6$.  }
    \label{fig:ai_behavior_k}
\end{figure}

\begin{figure}[H]
\centering
\begin{subfigure}{\textwidth}
\centering

\centering
  \includegraphics[scale=0.6]{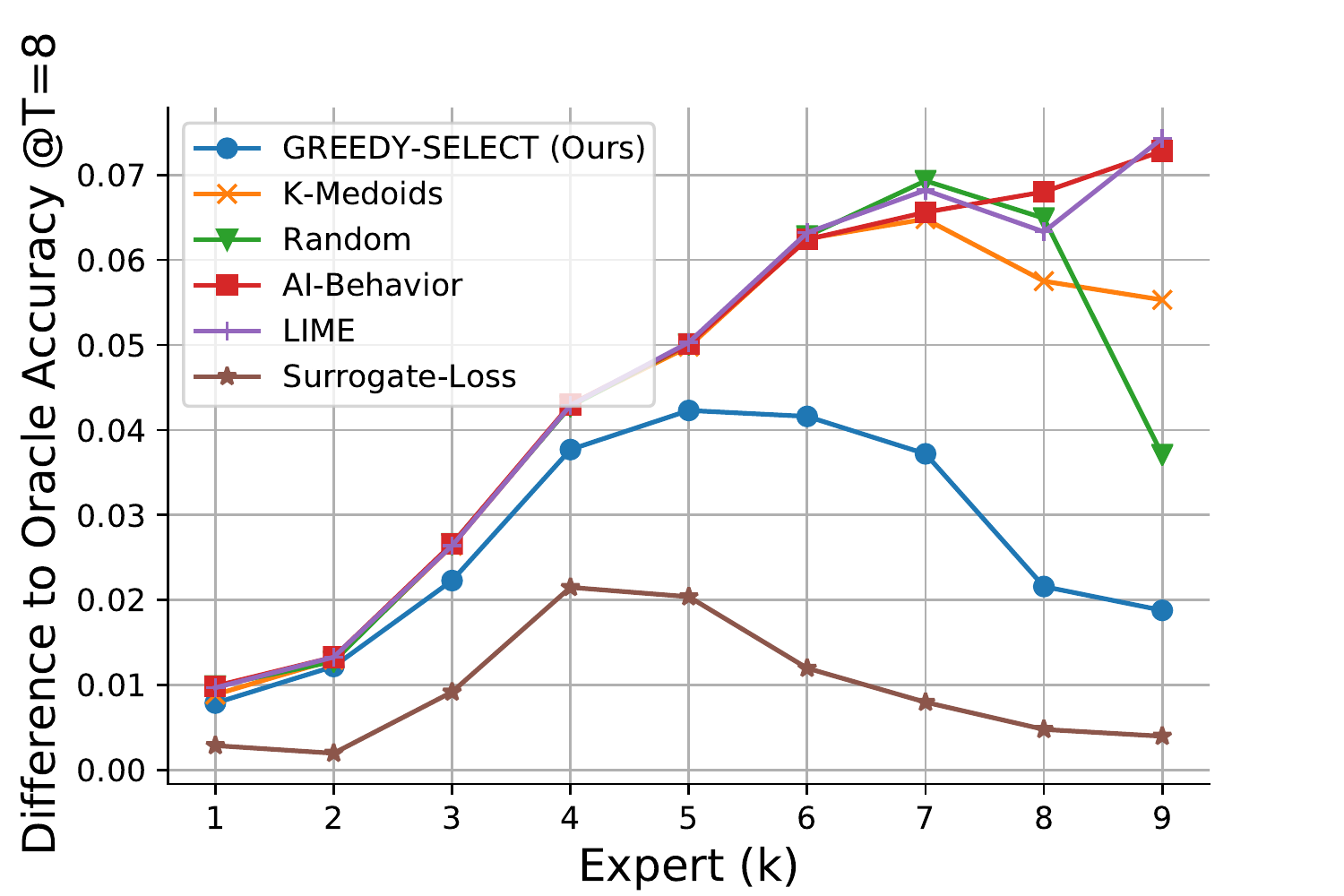}
\caption{Teaching size of 8 points}
\end{subfigure}

\begin{subfigure}{\textwidth}
\centering
  \includegraphics[scale =0.6]{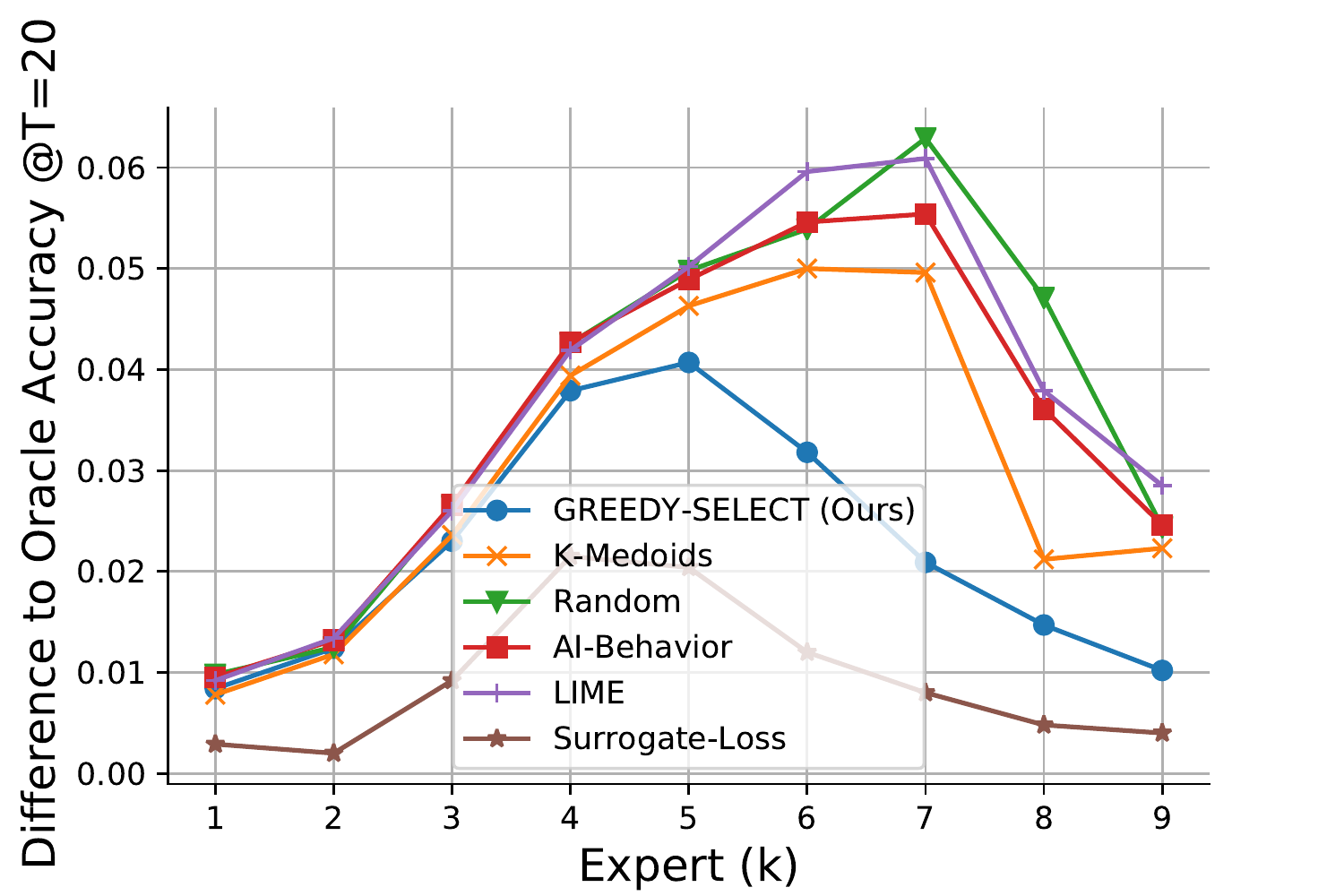}
\caption{Teaching size of 20 points}
\end{subfigure}

\begin{subfigure}{\textwidth}
\centering
  \includegraphics[scale =0.6]{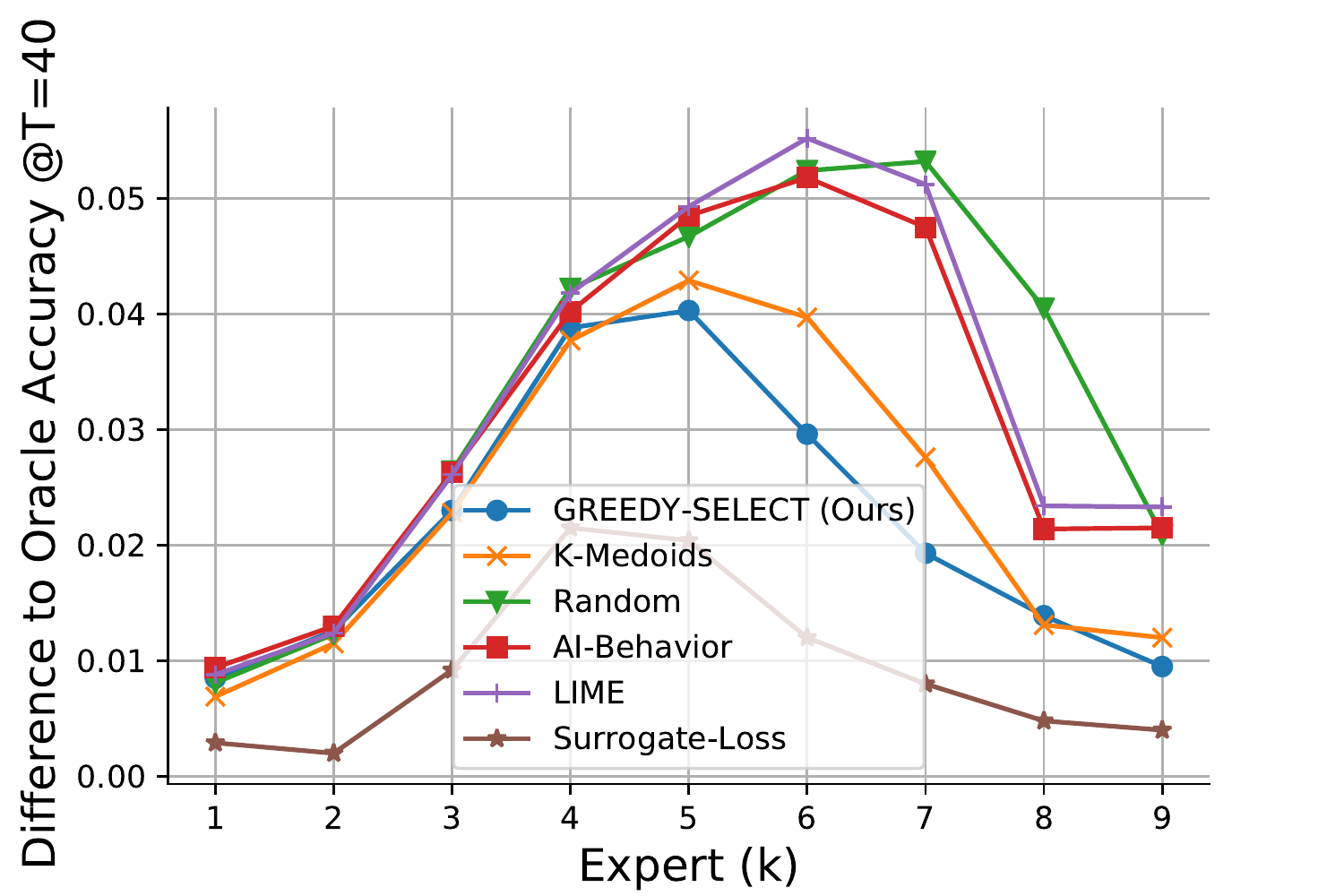}
\caption{Teaching size of 40 points}
\end{subfigure}

\caption{Extended legend: Varying the human parameter $k$ (number of classes human can classify)  and plotting the difference to oracle accuracy for all the baselines when using the consistent radius strategy including the surrogate-loss learning to defer method of \cite{mozannar2020consistent} at 3 different teaching set sizes.}
\label{fig:cifar_varyingk}
\end{figure}

\begin{figure}[H]
\centering
\begin{subfigure}{\textwidth}
\centering

\centering
  \includegraphics[scale=0.6]{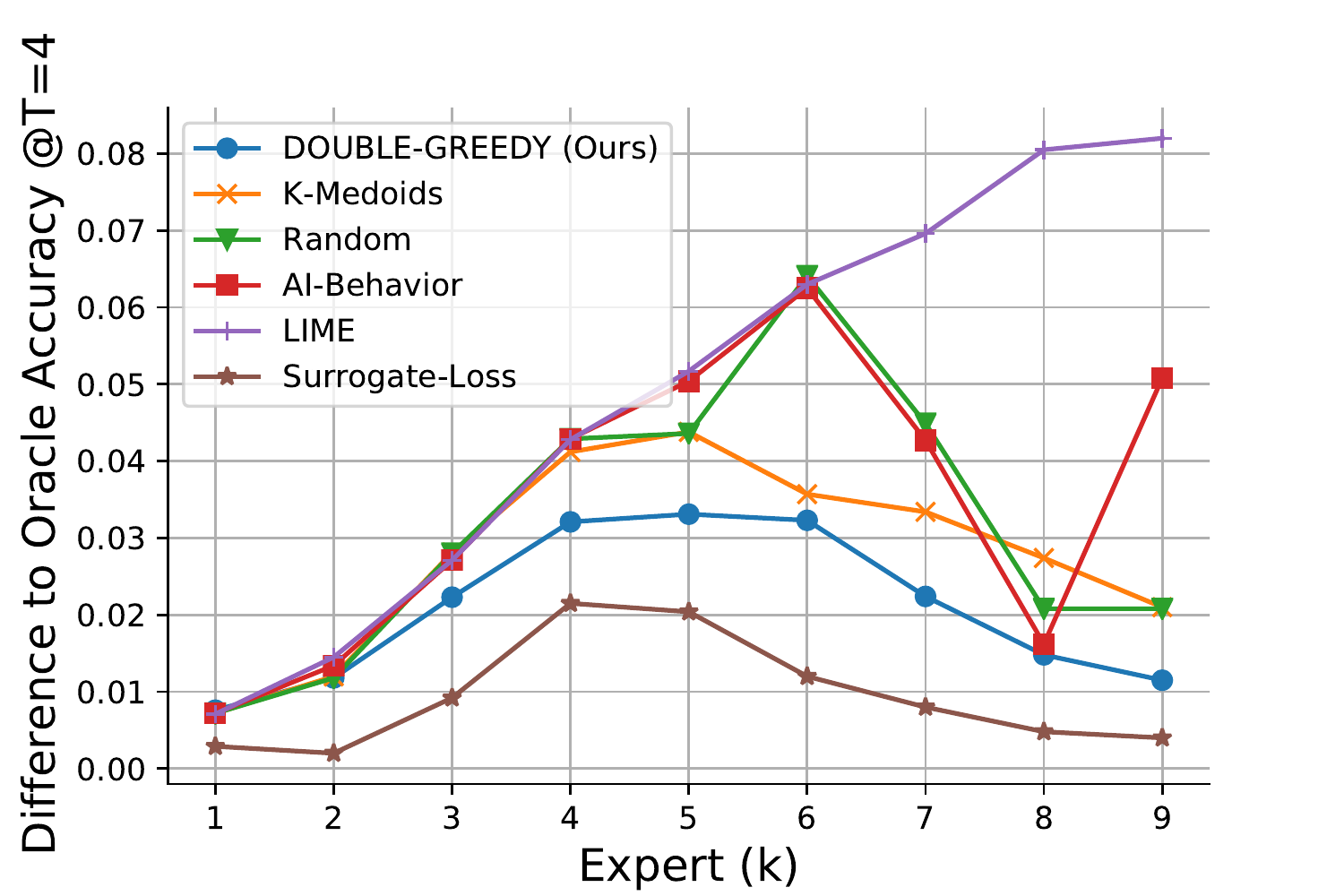}
\caption{Teaching size of 4 points}
\end{subfigure}

\begin{subfigure}{\textwidth}
\centering
  \includegraphics[scale =0.6]{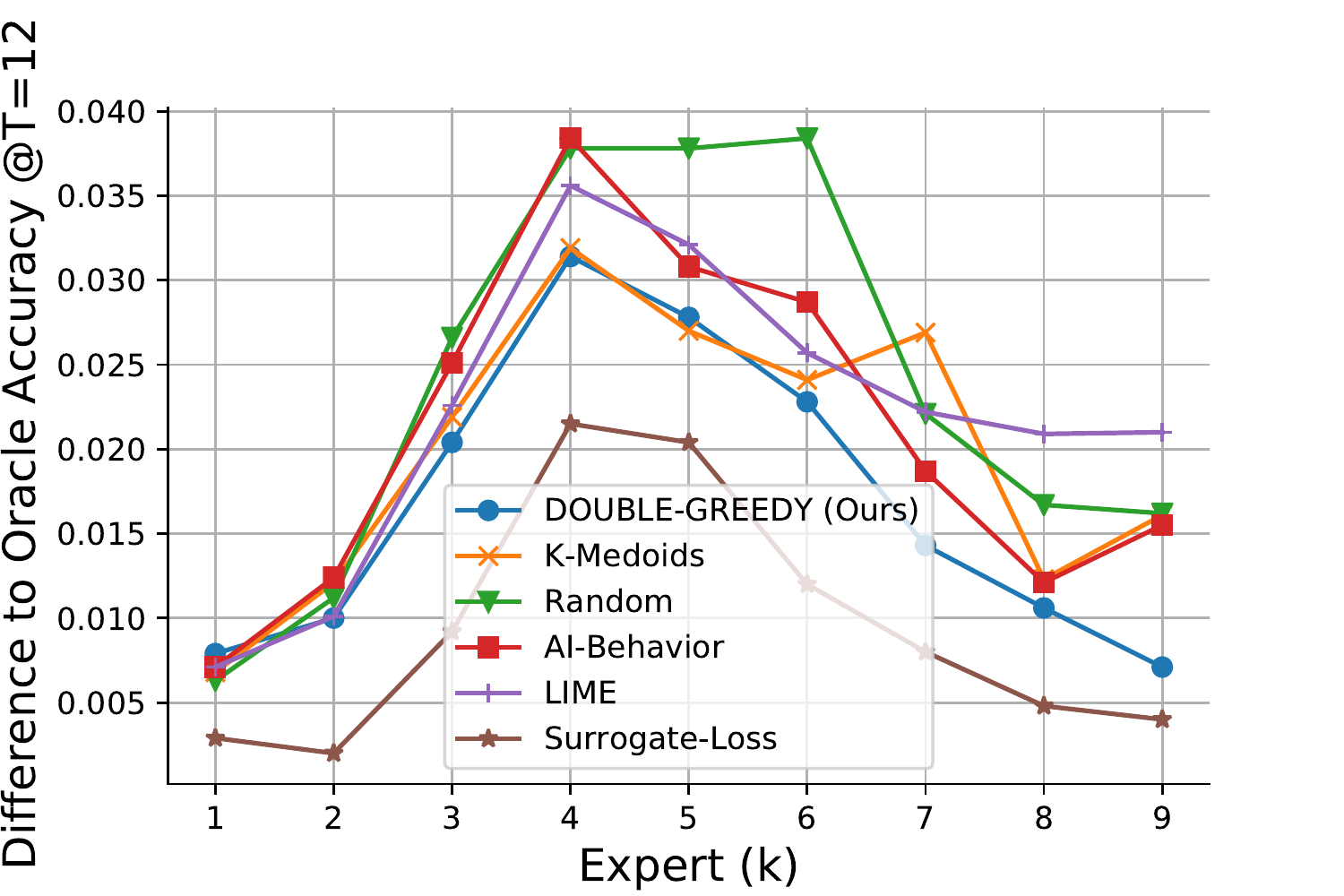}
\caption{Teaching size of 12 points}
\end{subfigure}

\begin{subfigure}{\textwidth}
\centering
  \includegraphics[scale =0.6]{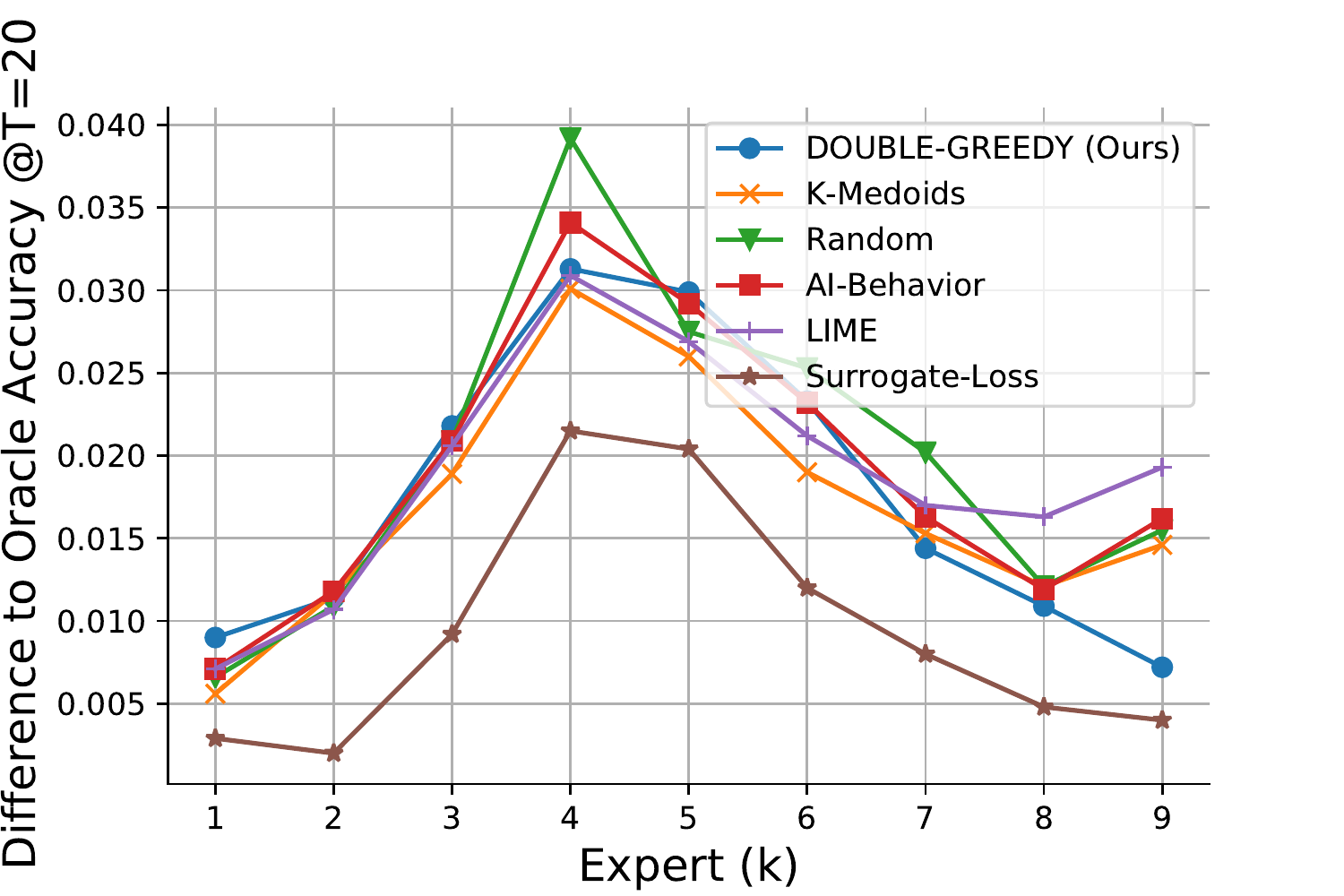}
\caption{Teaching size of 20 points}
\end{subfigure}

\caption{Extended legend: Varying the human parameter $k$ (number of classes human can classify)  and plotting the difference to oracle accuracy for all the baselines when using \texttt{DOUBLE-GREEDY} including the surrogate-loss learning to defer method of \cite{mozannar2020consistent} at 3 different teaching set sizes.}
\label{fig:cifar_varyingk_greedy}
\end{figure}

\subsection{Guassian Data Illustration}
Figure 1 illustrates the rejector for a linear classification setting, here we formalize this as a mixtures of Gaussian setup and show the performance of our selection algorithm both quantitatively and qualitatively.

\paragraph{Setup.} As an illustrative setting where we can visually inspect the teaching set, we perform experiments on  two dimensional Gaussian mixture data.  The covariate space is $\mathcal{X}= \mathbb{R}^2$ and target $\mathcal{Y} = \{0,1\}$, we assume that there exists two  sub-populations in the data denoted $A=1$ and $A=0$.
Furthermore, $X|(Y=y,A=a)$ is normally distributed according to $ \mathcal{N}(\mu_{y,a}, I)$. The group proportion is $\bP(A=1) =0.5$ and the means are sampled from a uniform distribution. 
The AI follows the Bayes solution for group $A=1$ which here corresponds to a hyperplane and the human classifier follows the Bayes solution for group $A=0$, which is another hyperplane. We assume the human's prior rejector is to reject based on a tresholding of the predictor confidence i.e. $g_0(x) = \bI \{ ||h(x)|| \leq \epsilon \} $ .  We assume that the similarity kernel is the RBF kernel $K(x,x')= e^{-||x-x'||^2}$.

\begin{figure}[H]
    \centering
    \includegraphics[scale=0.6]{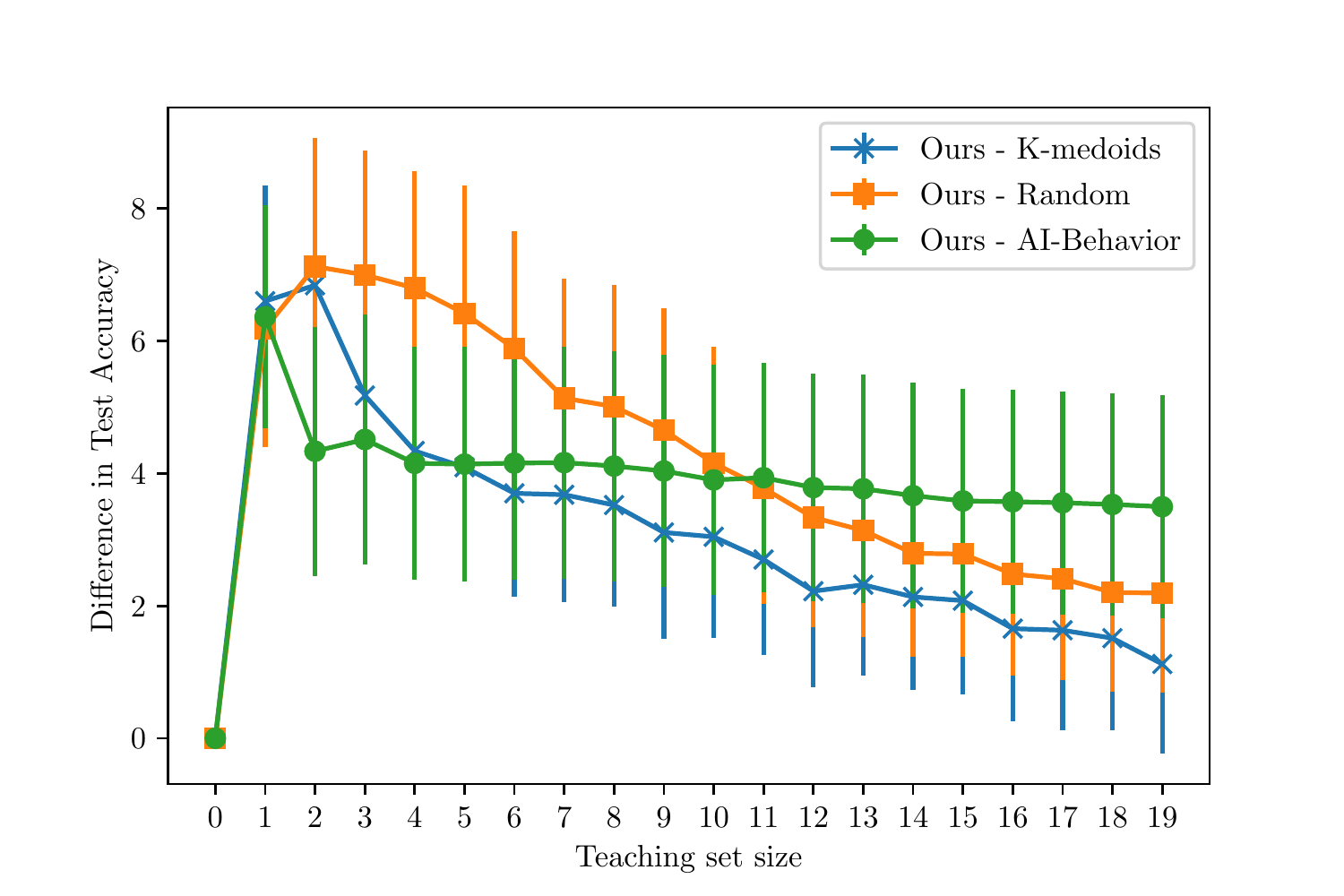}
    \caption{Teaching complexity plot for synthetic Gaussian data setup. The x-axis shows the difference in test human accuracy between our method and the baselines. Plotted are the averages over the 100 trials along with 95\% confidence interval error bars for the average.}
    \label{fig:toy_diff}
\end{figure}

\begin{figure}[H]
\centering
\begin{subfigure}{\textwidth}
\centering

\centering
  \includegraphics[scale=0.35]{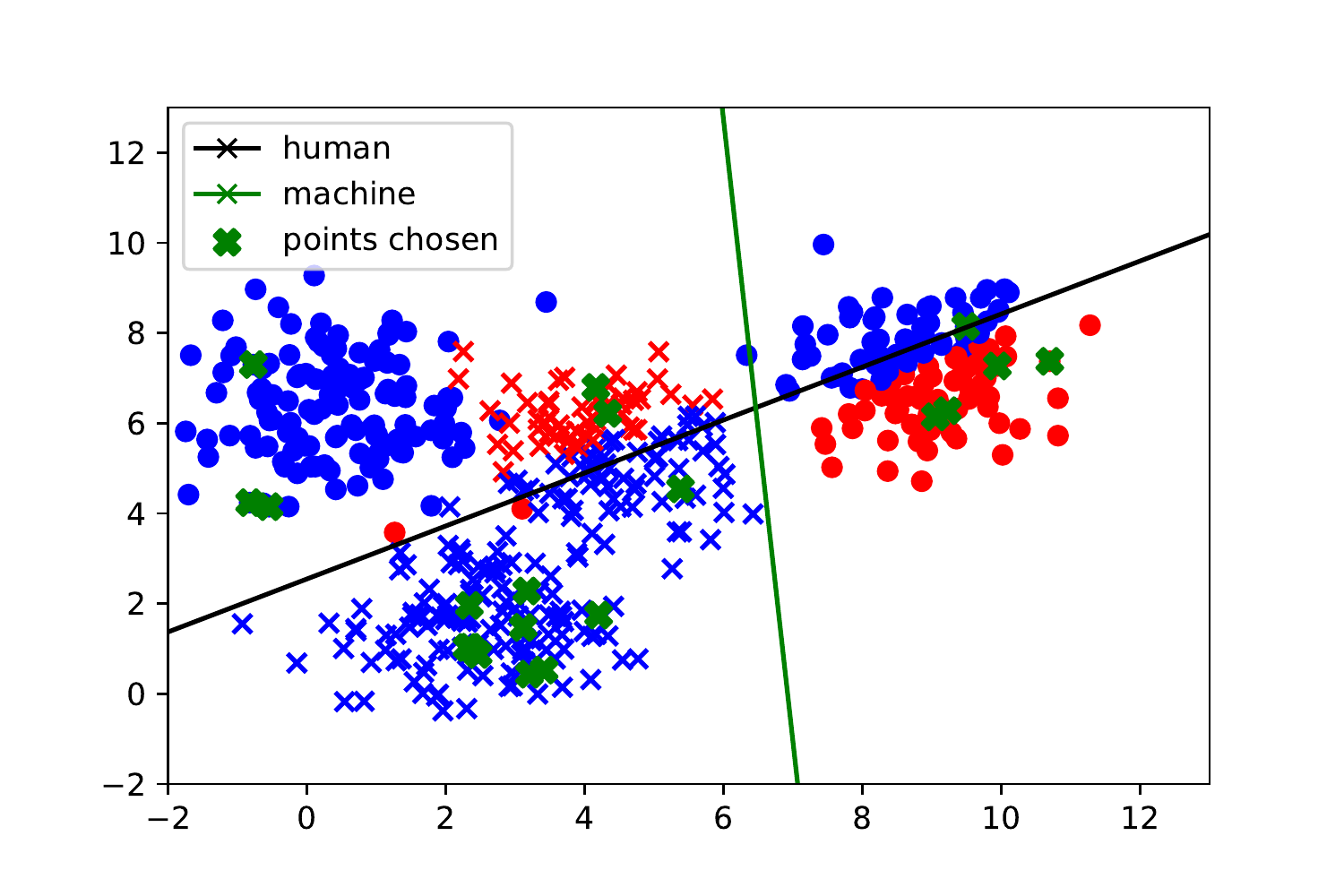}
\caption{Prior rejector with points chosen at step 20.}
\end{subfigure}

\begin{subfigure}{\textwidth}
\centering
  \includegraphics[scale =0.35]{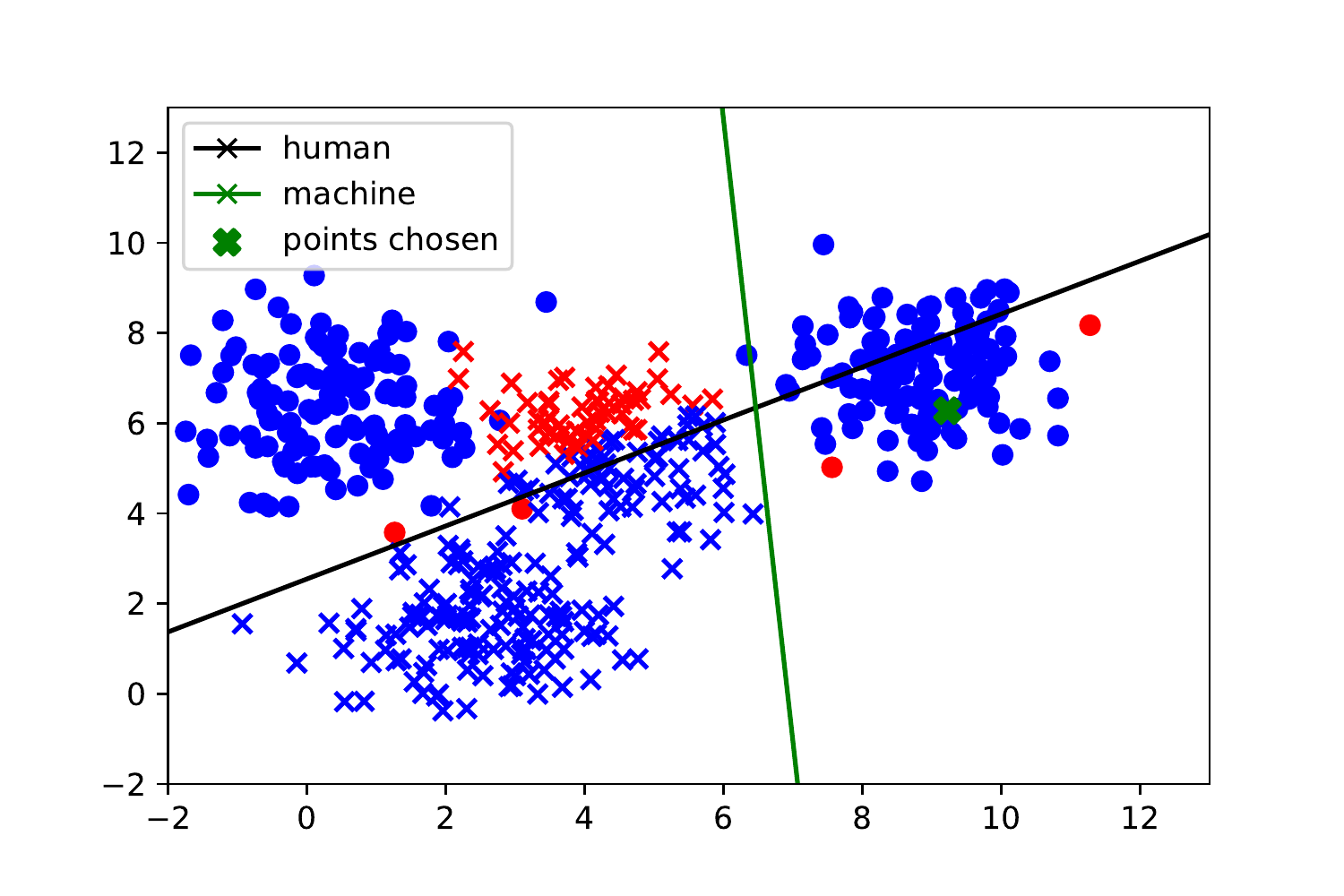}
\caption{Step 1 .}
\end{subfigure}

\begin{subfigure}{\textwidth}
\centering
  \includegraphics[scale =0.35]{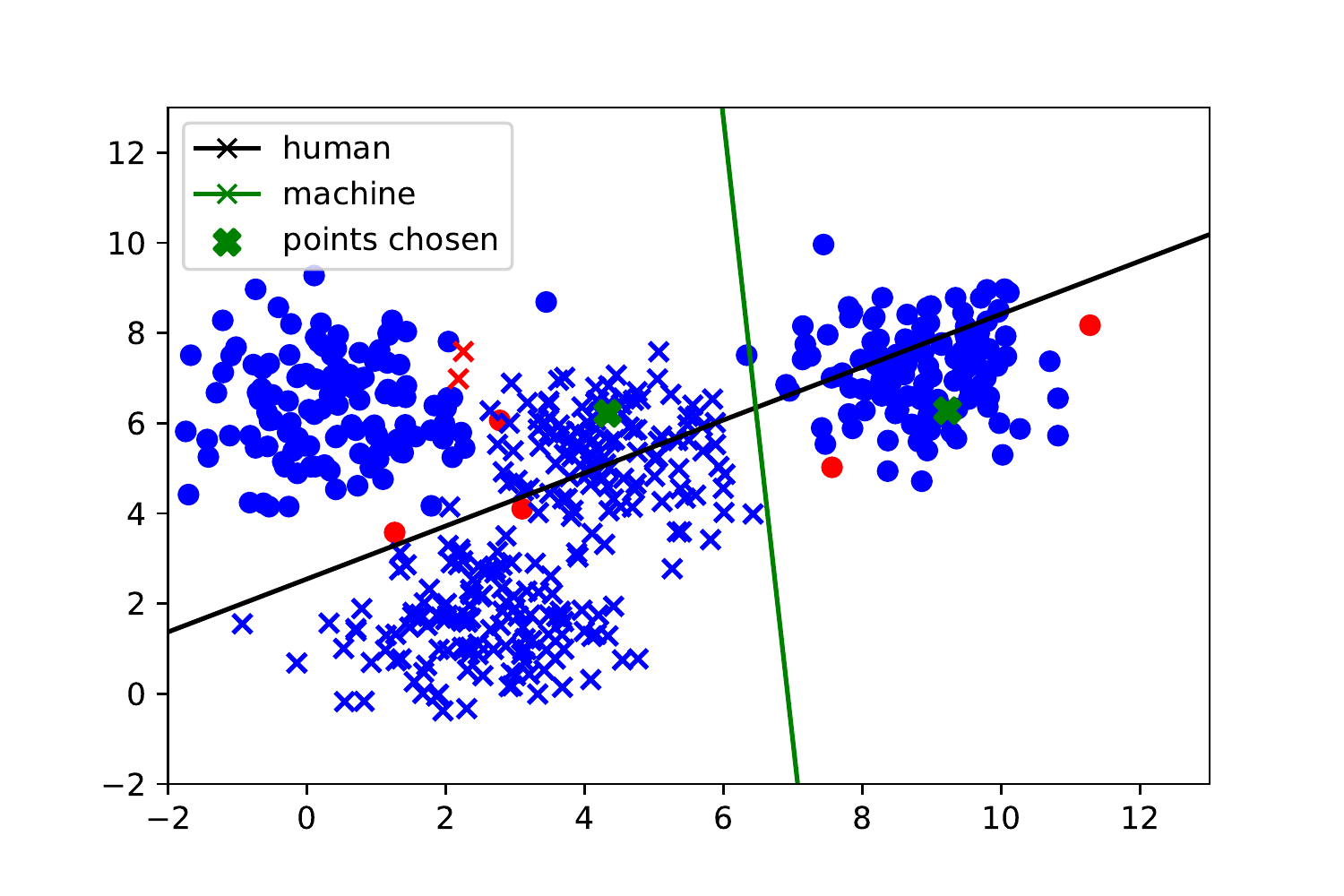}
\caption{Step 2 .}
\end{subfigure}

\begin{subfigure}{\textwidth}
\centering
  \includegraphics[scale =0.35]{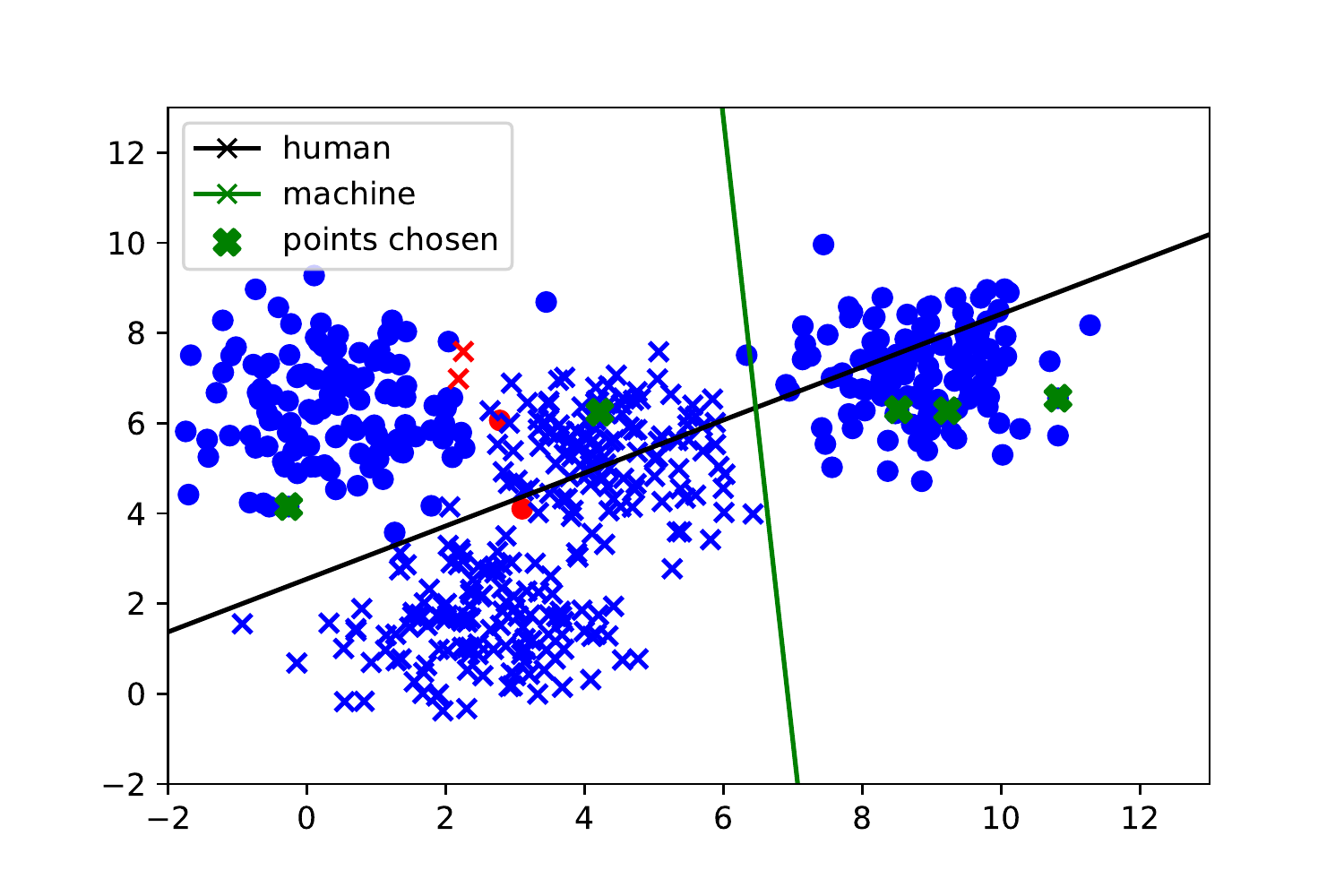}
\caption{Step 5 .}
\end{subfigure}

\begin{subfigure}{\textwidth}
\centering
  \includegraphics[scale =0.35]{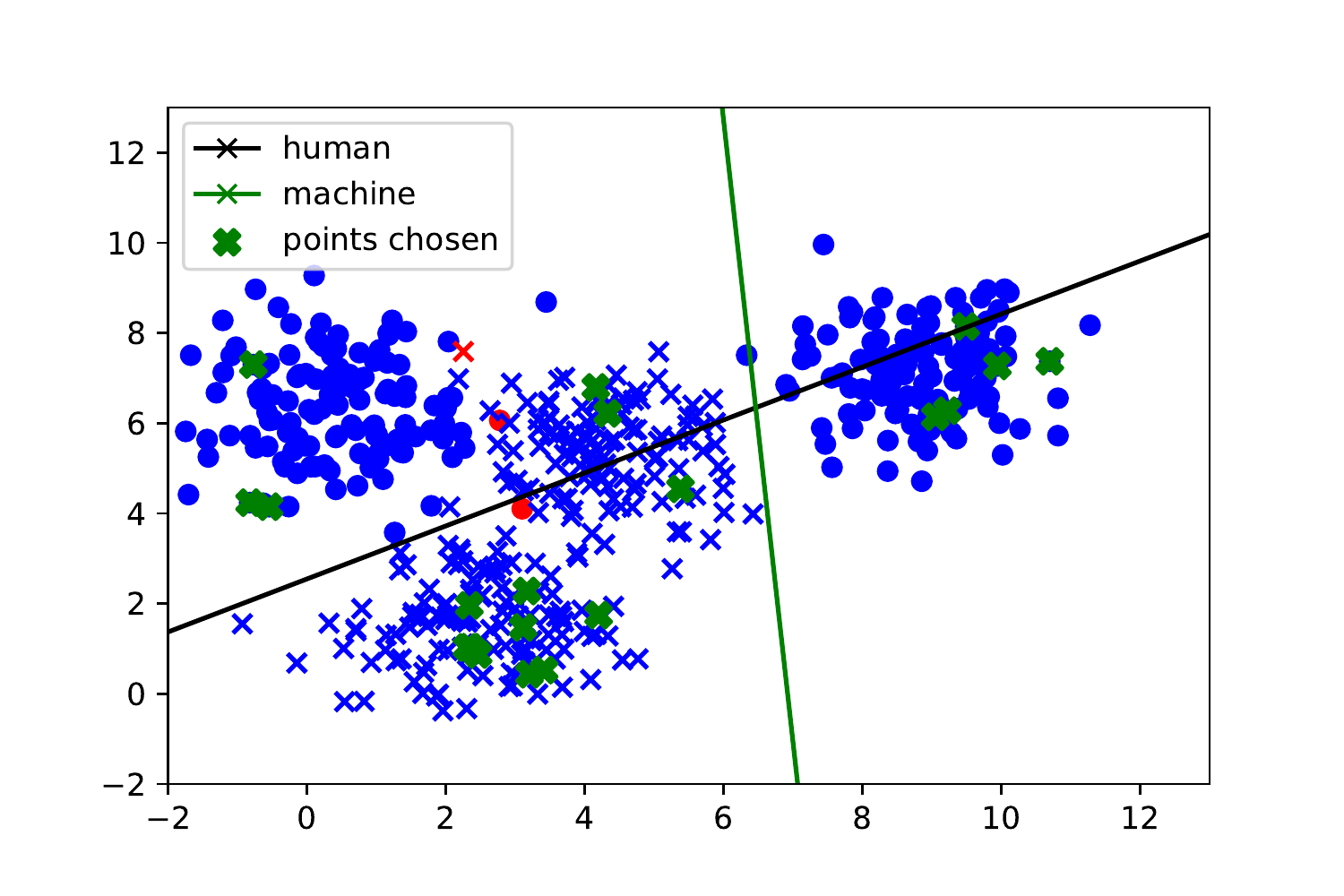}
\caption{Step 20 .}
\end{subfigure}

\caption{Extended legend: blue dots indicate a correct decision while red dots indicate mistakes. Points with an "x" are labels 1 while points with an "o" are labels 0 (in the Y space). The lines labeled human and machine are the respective classifiers.}
\label{fig:synthetic_pointschosen}
\end{figure}

\paragraph{Results.} For 100 trials, we generate data with random means  and measure the difference in system accuracy between our approach and the baselines as we vary the size of the teaching set. Results are shown in Figure \ref{fig:toy_diff}. Figure \ref{fig:synthetic_pointschosen} shows the points chosen on a given configuration. 

\clearpage
\section{Crowdsourced Experiments Details and Results}\label{apx:crowd_experiments}

\subsection{Experiment Details}

\paragraph{Participants.} We recruited 50 US based participants from Amazon Mechanical Turk per each condition (100 total), workers were required to have a HIT approval rate higher than 95\% and over 100 HITs approved. Initial pilot studies were also conducted with graduate students in computer science at a US university. 
Participants in the baseline were paid \$3 for 10 minutes of work and those in the teaching condition received \$6 for 20 minutes of work. Any demographic information we gathered in our study is kept confidential and workers were asked to consent to their use of their responses in research studies. We submitted an IRB application and the IRB declared it exempt as is. We followed standard protocol and additionally provided the IRB exemption and details to our user study participants. We filter participants who don’t answer the tutorial questions correctly and we also filter for all baselines that workers at least answer one question correctly on their own beyond the first question.

\paragraph{AI and Test Set details.} The simulated AI used in the study was obtained by first performing K-means with $K=25$ on the dev set of HotpotQA, and then manually filtering the data to obtain 11 clusters that are more distinct. 
 The test set used in the testing phase was obtained first by filtering the data using K-medoids with $K=200$ as a way to get diverse questions. We then created 20 test sets by sampling  7 random  questions from the filtered set on which the AI was correct and 8 on which the AI is incorrect. The order of the examples in the test set was shuffled for each participant.

\paragraph{Cluster Topics.} The AI used in the study had 11 different clusters on which it's errors were defined. Table \ref{fig:cluster_topics} shows the main theme and most common Wikipedia categories for each cluster.

\begin{table}[h]
\caption{Cluster main theme (manually obtained) and top 3 Wikipedia categories of examples in clusters for the AI used in the MTurk study. }

\begin{tabularx}{\textwidth}{llX}
\toprule
Cluster ID & Main Theme & Wikipedia Categories  \\
\midrule
1  & Plants & Poaceae genera, Flora of Mexico, Dioecious plants  \\
2  & Singers, Musicians & 21st-century American singers, Grammy Award winners, American male guitarists \\
3  & Movies, Actors & American films, British films, American male film actors  \\
4  & Sites, Hotels & Casino hotels,
Casinos in the Las Vegas Valley,
Resorts in the Las Vegas Valley \\
5  & Writers, Magazines & 20'th-century American novelists,
American male non-fiction writers,
American women novelists \\
6  & Composers, Plays & 19th-century classical composers,
Operas,
Male classical pianists \\
7  & Games & Windows games,
PlayStation 4 games,
Xbox One games \\
8  & Universities & Universities and colleges,
Colonial colleges,
Private universities in New York \\
9  & Soccer & Premier League players,
English Football League players,
Association football midfielders \\
10  & Sports (non soccer) & American men's basketball players,
NFL player,
NBA All-Stars \\
11  & Politics & 21st-century American politicians,
Presidential Medal of Freedom recipients,
Republican Party members  \\
\bottomrule
\end{tabularx}

\label{fig:cluster_topics}
\end{table}

\paragraph{User Lessons.} In Table \ref{fig:user_lessons} we show examples of the lessons that the crowdworkers wrote during the teaching phase for the proposed teaching method. We show examples of the lessons on the first 3 examples in the teaching phase and separate the participant lessons into 4 categories: participants who wrote accurate lessons, participants who wrote irrelevant lessons (not relevant to the question  or required no effort to write), participants who wrote complex lessons that don't pertain to the example topic and finally participants who wrote narrow lessons that are on topic but only apply to the example and not the neighborhood of the example. In Table \ref{fig:turk_results} we separated user metrics into two groups accurate lessons and inaccurate lessons, this corresponds to grouping accurate lessons versus the rest in the lesson categorization of Table \ref{fig:user_lessons}. Furthermore, in the body of section \ref{sec:user_study_observations} we distinguish between accurate lessons, narrow and complex lessons (combined into one group) and finally  irrelevant lessons.

\begin{table}[h]
\caption{ Example of lessons that users in the Ours-Teaching condition wrote during the teaching phase. We show examples of the lessons on the first 3 examples in the teaching phase and separate the participant lessons into 4 categories: participants who wrote accurate lessons, participants who wrote irrelevant lessons (not relevant to the question  or required no effort to write), participants who wrote complex lessons that don't pertain to the example topic and finally participants who wrote narrow lessons that are on topic but only apply to the example and not the neighborhood of the example. }

\begin{tabularx}{\textwidth}{llX}
\toprule
Lesson Type & Example ID & Actual Lesson \\
\midrule
Accurate Lessons & 1 & The AI is not good at answering questions about plants. \\
Accurate Lessons & 2 & The AI is better at Politics and geography than at sports. \\
Accurate Lessons & 3 & The AI is bad at answering questions about movies \\
\midrule
Irrelevant Lessons & 1 & I understood AI is good at answering \\
Irrelevant Lessons & 2 & AI focus on the institution \\
Irrelevant Lessons & 3 & AI omitted important terms \\
\midrule
Complex Lessons & 1 & It seems to be better at answering questions where the absolute same phrases are used in the question as the passage and where both answers are in the question, maybe? \\
Complex Lessons  & 2 & The ai is good at answering questions that has to do with cities and numbers though not good with words that has to do with repeated words. \\
Complex Lessons  & 3 & The AI can't decipher clues, example, the other movie was based on a book that came out after the other movie but the AI couldn't figure out that that must mean the movie based on that book must then also have come out after the other movie. \\
\midrule
Narrow Lessons & 1 &  The AI isn't good at multi-faceted questions about continental species.\\
Narrow Lessons & 2 & The topic was politics and the AI is good at answering questions about specific areas when the question can be answered by looking for specific information about one section but not when it involves integrating multiple pieces of information from the paragraph.\\
Narrow Lessons & 3 &  The AI isn't good at comparing media release dates..\\
\bottomrule
\end{tabularx}

\label{fig:user_lessons}
\end{table}

\section{Extended Discussion}

One limitation of our human
experiments is that we used a simulated AI that has an easier
to understand error boundary. This enabled us to have a
more in-depth study of the crowdworker responses than otherwise
would have been possible.

Having a simulated AI to which we perfectly understand where it's error regions are (but note this is highly non trivial for someone who doesn't know how it was trained), enables to define what the "lessons" should be and thus evaluate if users are learning correctly. This ability to evaluate if users are actually learning through their written lessons enables to test two things:
\begin{enumerate}
    \item  Do people learn the correct lessons using our teaching method?
    \item Do those who learn the correct lessons apply them perfectly?
\end{enumerate}

And our answers in our paper to these questions are: 1) yes but only half the people are able to, 2) not quite, since even those with perfect lessons don't show perfect accuracy (in Defer F1).
What is interesting about this last observation tell us that even if people know the rules, and have them written and shown on the screen, they might still apply it incorrectly. With a non simulated AI, it would have been difficult for us to figure out the answers to the questions as the underlying lessons are not pre-determined.
 For an initial experimental study on teaching, we need to understand better how do humans make decisions and how we can try to use their lessons to possibly provide feedback and better guide them.

Another limitation is that
our test-time interface did not include model explanations or predictions. This was done for multiple reasons:

\begin{itemize}
    \item The AI predictions and explanations reveal information about it's underlying performance at test time. If two different crowdworkers received different test sets, then their knowledge about the AI may be different. Therefore if the participants belonged to two different experimentation conditions, then the test set becomes a confounding factor we need to control for.
    \item When model explanations are not available or are not effective, the effect of teaching becomes more important as it is the only way the human's mental model is formed. Thus the choice of the teaching method  becomes more important.
    \item  If the AI prediction is available at test time and workers press on the "Use AI answer" button, there is an unobservability issue that arises: are workers pressing the button because they trust the AI, or are workers pressing the button because they came up with the same answer on their own? Removing their ability to see the AI prediction alleviates the problem.
\end{itemize}

\clearpage
\section{User Interface Screenshots}

\begin{figure}[h]
    \centering
        \resizebox{\textwidth}{!}{

    \includegraphics{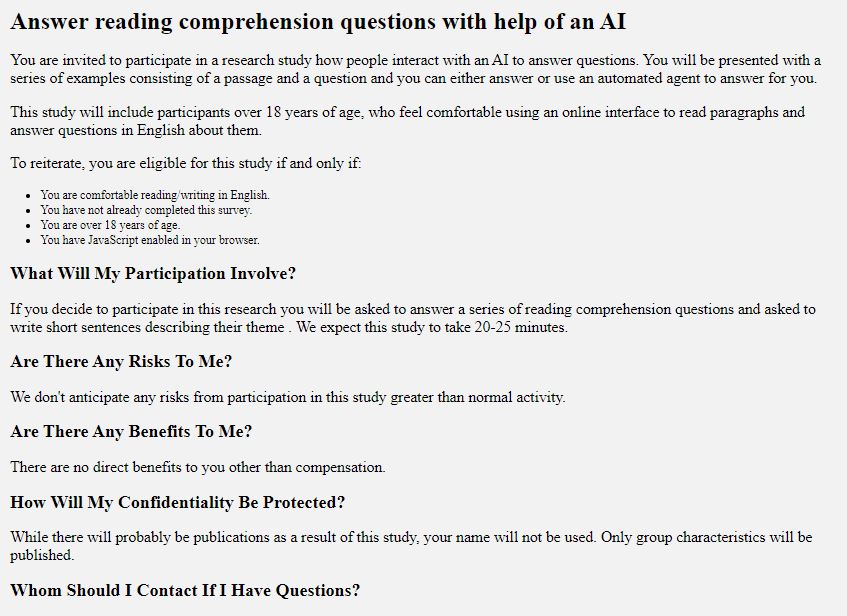}
}
    \caption{Consent form to be confirmed before entering experiment}
\end{figure}

\begin{figure}[h]
    \centering
        \resizebox{\textwidth}{!}{

    \includegraphics{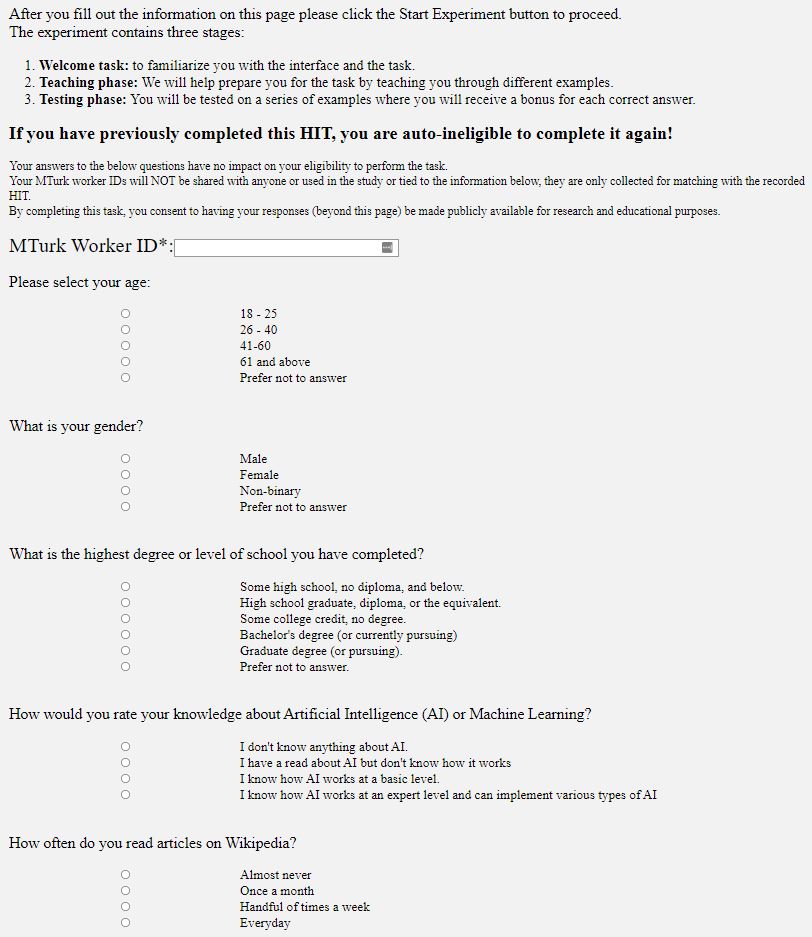}
}
    \caption{Information collected about workers prior to experiment. MTurk worker ID was only saved for cross-checking and then deleted.}
\end{figure}
\begin{figure}[h]
    \centering
        \resizebox{\textwidth}{!}{

    \includegraphics{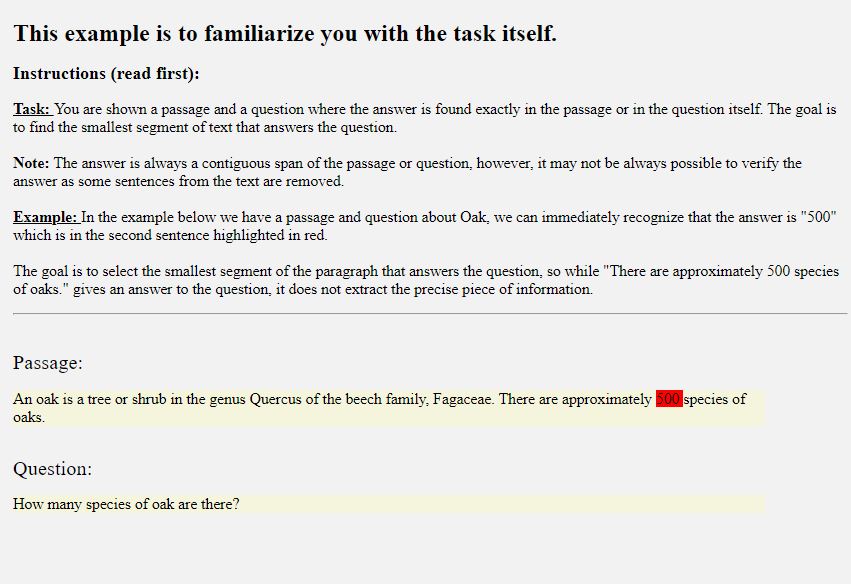}
}
    \caption{First step of the tutorial introducing the task}
\end{figure}

\begin{figure}[h]
    \centering
        \resizebox{\textwidth}{!}{

    \includegraphics{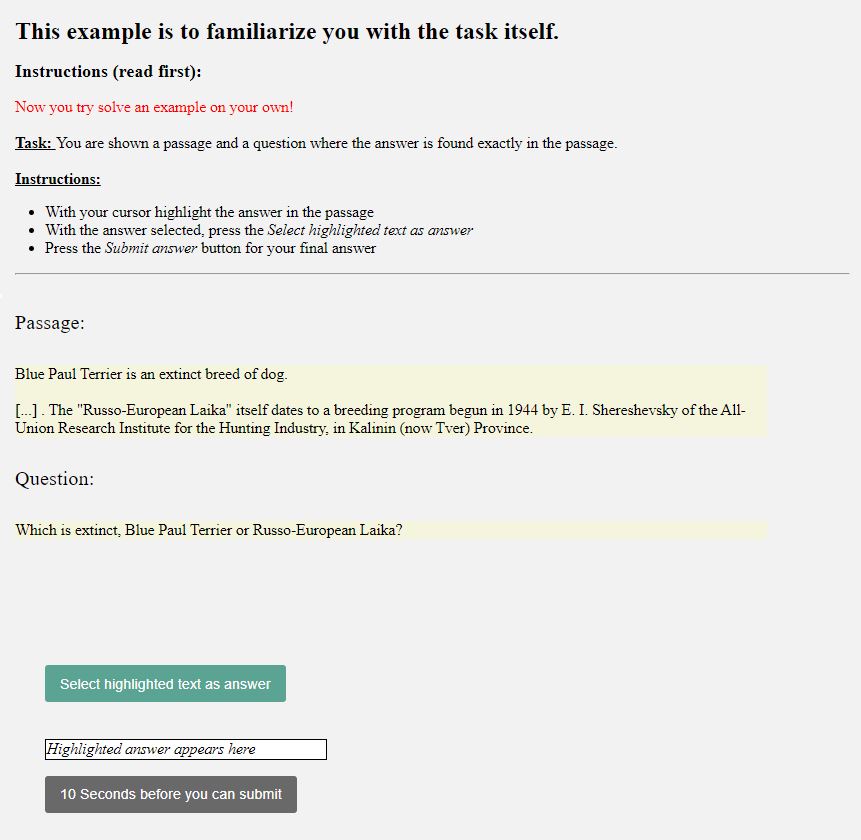}
}
    \caption{Second step of the tutorial solving without AI help}
\end{figure}

\begin{figure}[h]
    \centering
        \resizebox{\textwidth}{!}{

    \includegraphics{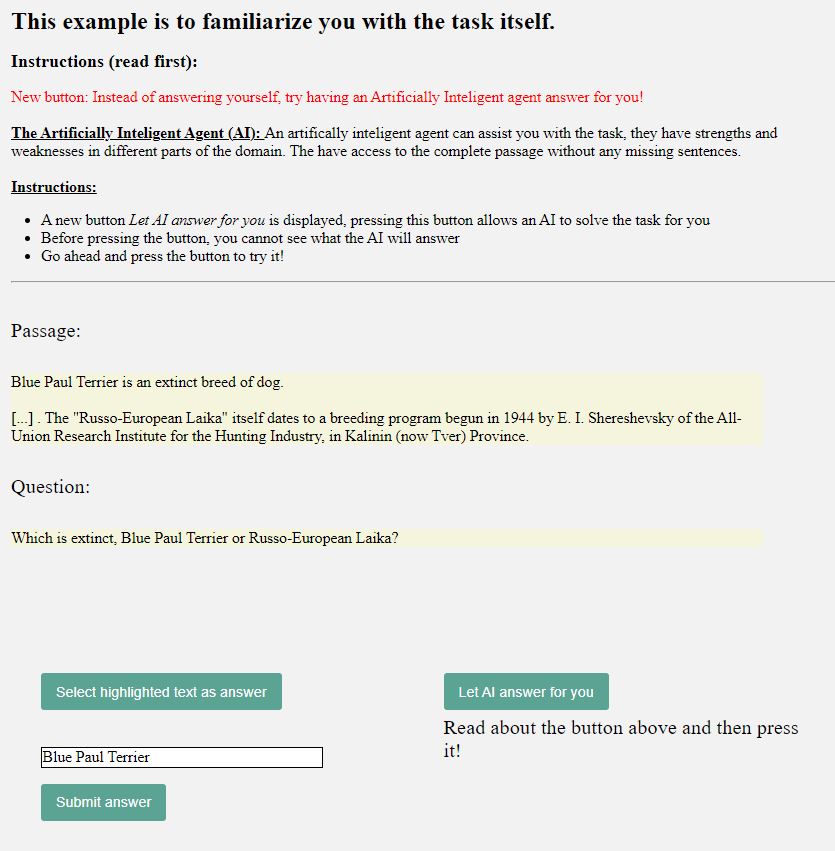}
}
    \caption{Third step of the tutorial solving with AI help}
\end{figure}

\begin{figure}[h]
    \centering
        \resizebox{\textwidth}{!}{

    \includegraphics{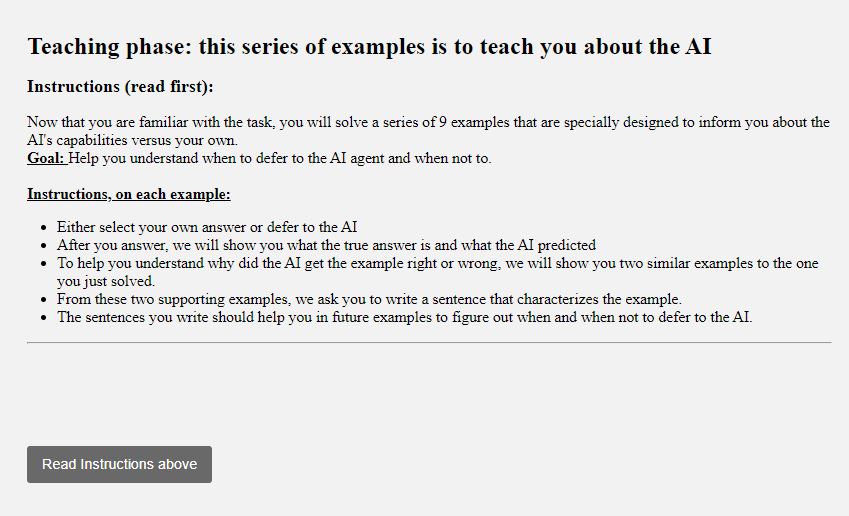}
}
    \caption{Teaching instructions}
\end{figure}

\begin{figure}[h]
    \centering
        \resizebox{\textwidth}{!}{

    \includegraphics{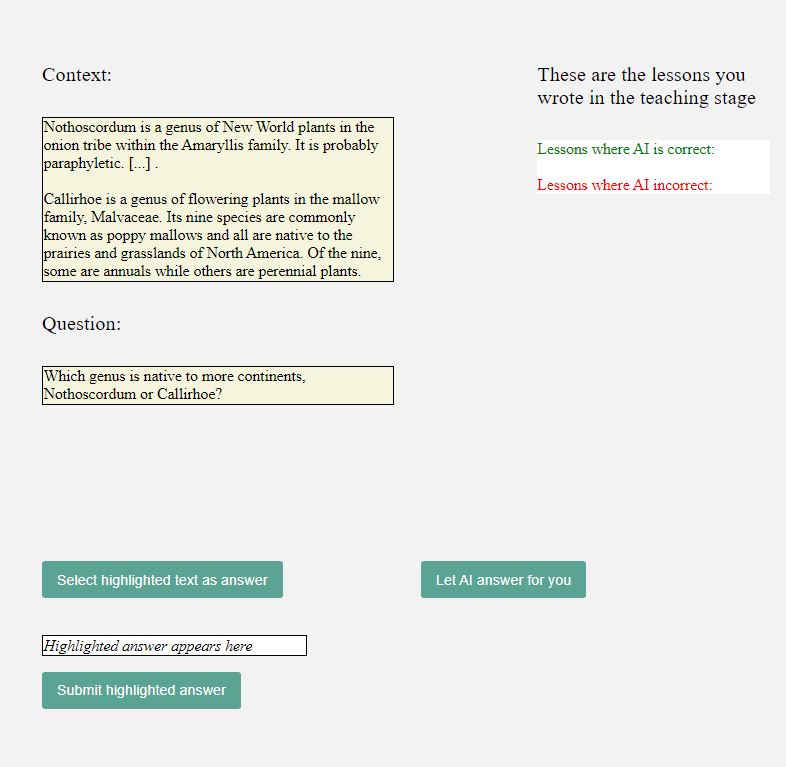}
}
    \caption{Teaching initial example to be solved by the human.}
\end{figure}

\begin{figure}[h]
    \centering
        \resizebox{\textwidth}{!}{

    \includegraphics{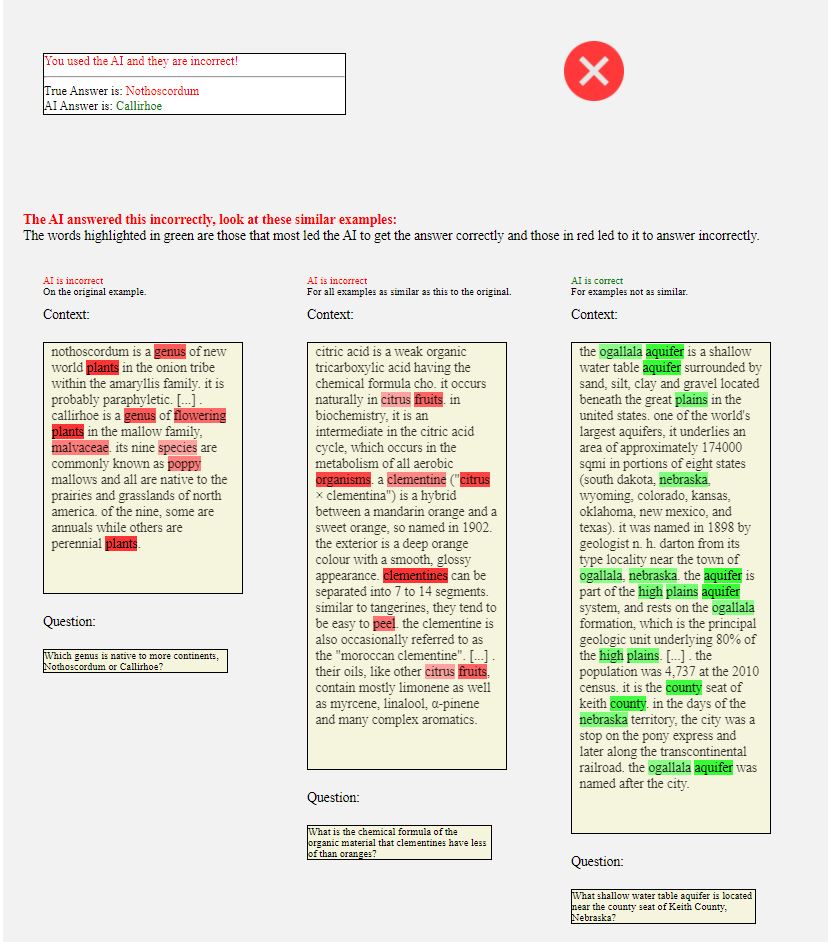}
}
    \caption{Feedback shown after human solves the example along with supporting examples.}
\end{figure}

\begin{figure}[h]
    \centering
        \resizebox{\textwidth}{!}{

    \includegraphics{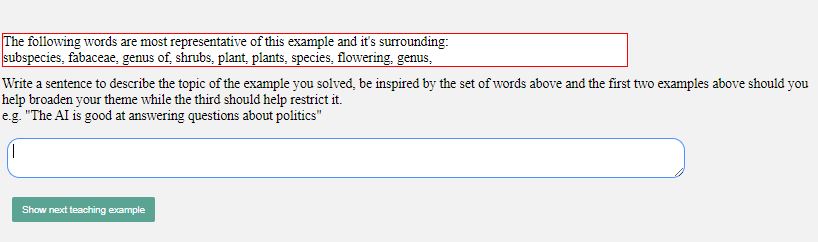}
}
    \caption{Top words for the teaching example along with instructions for lesson writing}
\end{figure}

\begin{figure}[h]
    \centering
        \resizebox{\textwidth}{!}{

    \includegraphics{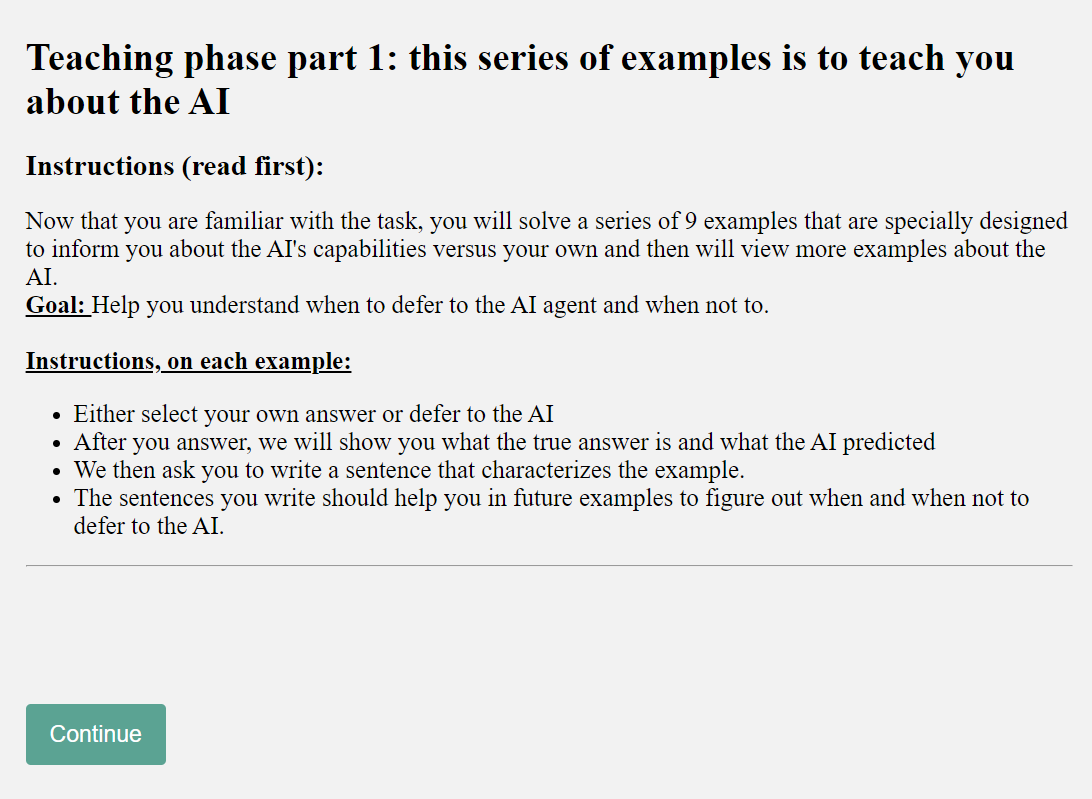}
}
    \caption{The LIME-Teaching user teaching introduction}
\end{figure}

\begin{figure}[h]
    \centering
        \resizebox{\textwidth}{!}{

    \includegraphics{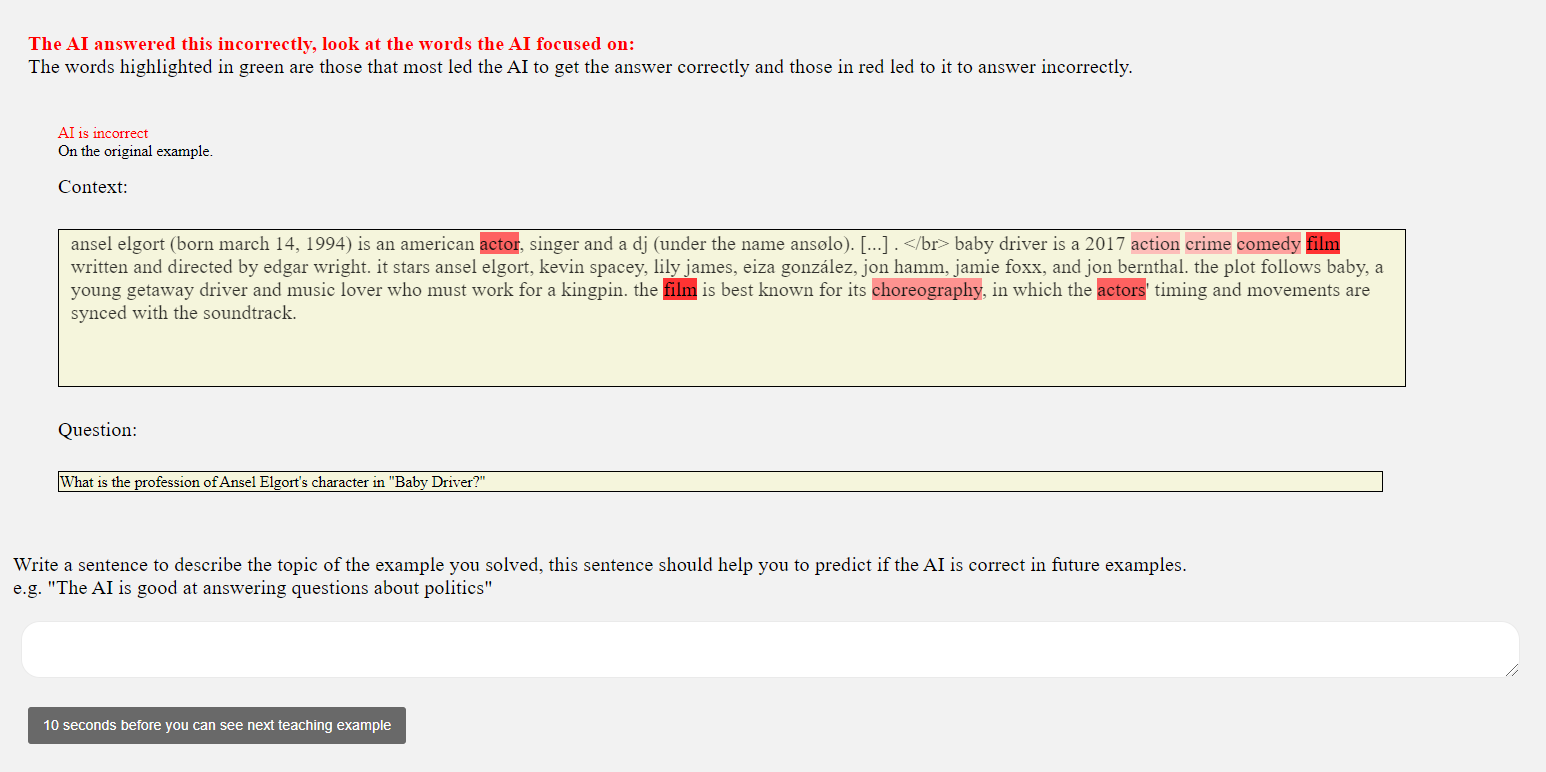}
}
    \caption{The LIME-Teaching feedback after answering teaching question.}
\end{figure}

\begin{figure}[h]
    \centering
        \resizebox{\textwidth}{!}{

    \includegraphics{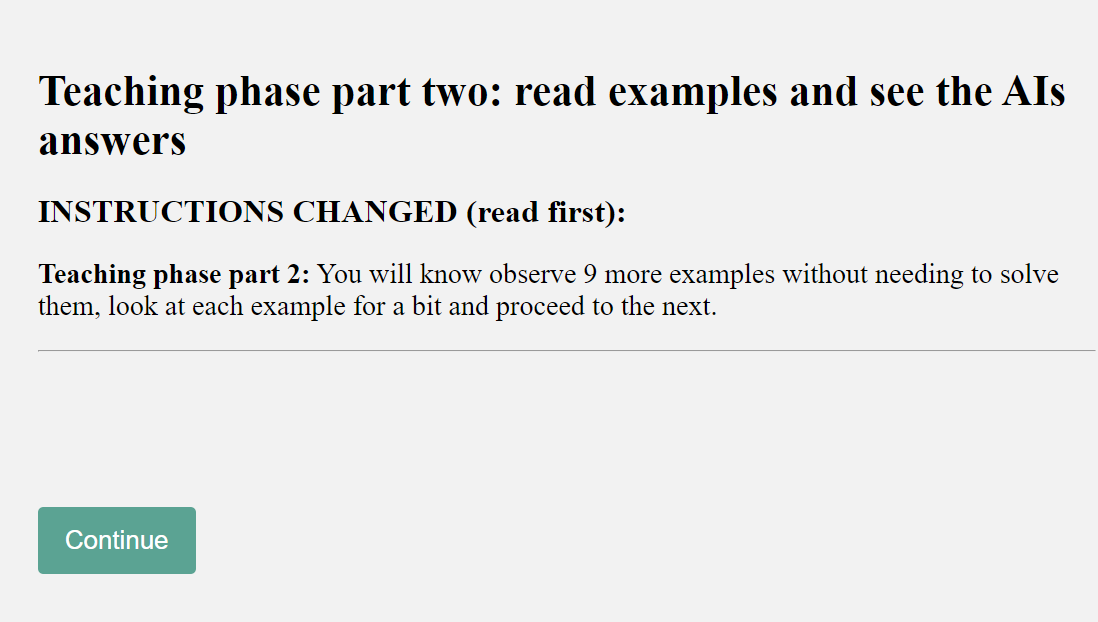}
}
    \caption{The LIME-Teaching  teaching introduction to second part of the teaching phase}
\end{figure}

\begin{figure}[h]
    \centering
        \resizebox{\textwidth}{!}{

    \includegraphics{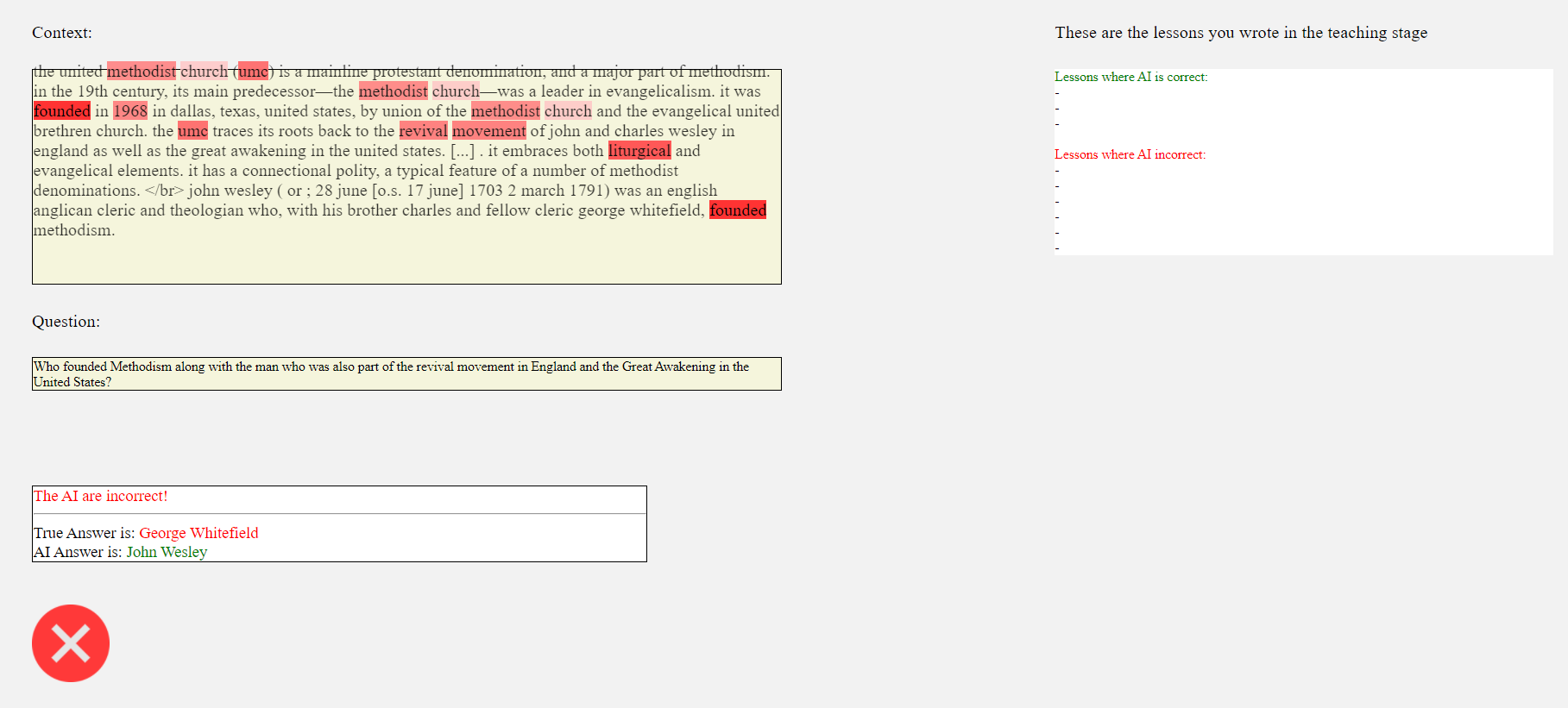}
}
    \caption{The LIME-Teaching user interface of the second part of the teaching phase where users observe examples and the AI answers.}
\end{figure}

\begin{figure}[h]
    \centering
        \resizebox{\textwidth}{!}{

    \includegraphics{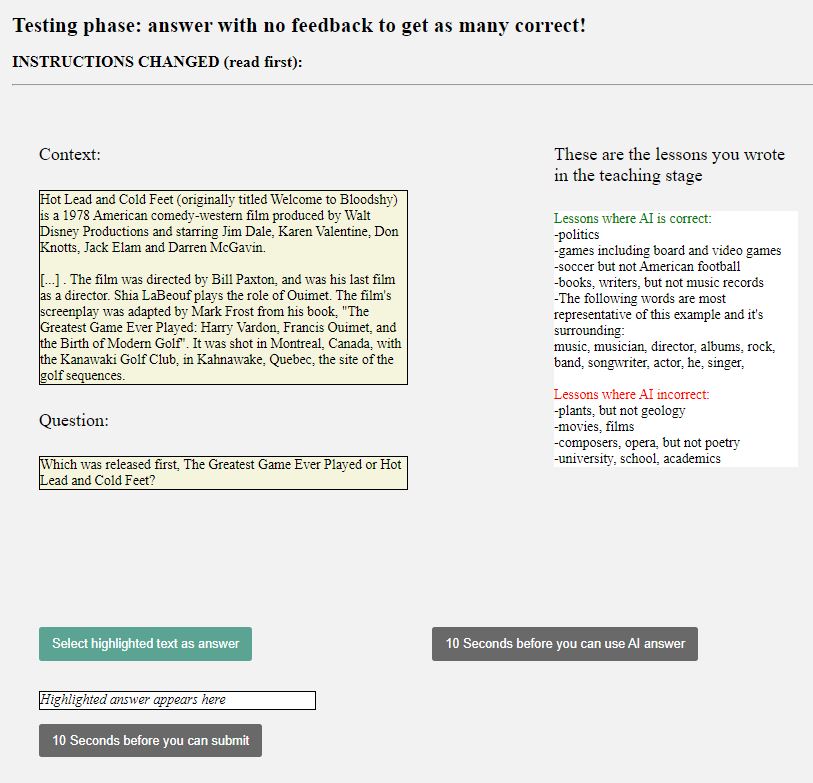}
}
    \caption{Interface during testing.}
\end{figure}

\begin{figure}[h]
    \centering
        \resizebox{\textwidth}{!}{

    \includegraphics{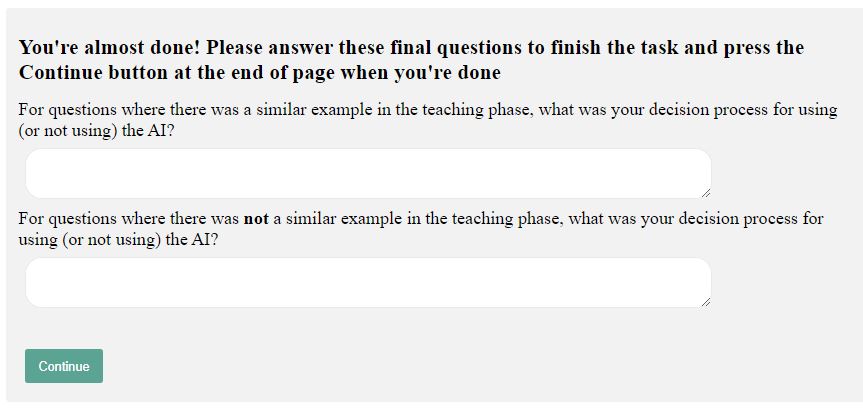}
}
    \caption{Questions collected after workers complete experiment for the Teaching condition.}
\end{figure}

\clearpage

%% file: appendix/related_work.tex
\section{Extended Related Work} \label{apx:related_work}

\paragraph{Human-AI interaction.}
A significant amount of research has tried to understand the role of explanations on Human-AI team performance. 
\cite{lai2019human} investigates the role of increasing levels of AI explanation on performance and find that beyond showing predicted labels accuracy does not increase. \cite{lage2019evaluation} identify different regularizers that optimize for factors that help humans better simulate and verify AI predictions on recommendation tasks. \cite{smith2020no} investigates how the ability of humans to provide feedback to the model reduced user frustration on a text classification task. \cite{hase2020evaluating} evaluated different explanation methods on the adult income dataset and on a movie reviews dataset found that only LIME helped for simulating the model and that subjective user ratings of explanation quality were not predictive of effectiveness. 
More research on the adult income dataset found that showing AI confidence improved trust but failed to improve AI-assisted accuracy \cite{zhang2020effect}.  \cite{kocielnik2019will} studies how different types of errors an AI may have will lead to different perceptions of the AI by the user, and how setting expectations of the AI capabilities (e.g. its accuracy) improves the user experience. \cite{bansal2020does} show on a beer/book reviews sentiment classification task and on LSAT multiple choice questions that AI explanations beyond confidence scores don't improve performance but rather increase blind trust in the AI system. \cite{levy2021decisionaid} on the task of annotating clinical texts show that clinicians generally build a mental model of when to rely on automation, however, when the AI presents a complete suggestion versus an incomplete one, this causes experts to show less agency and makes them more likely accept wrong answers. In similar lines, \cite{suresh2020misplaced} studied how do humans incorporate AI recommendations as a function of their correctness and their prior knowledge of machine learning, and showed that people follow incorrect AI recommendations for tasks
they predominantly complete correctly and that incorrect-abnormal recommendations were followed significantly less than incorrect normal recommendations. 
\cite{chu2020visual} on the task of age prediction from images showed that the addition of explanations in  the form of saliency maps did not improve accuracy nor did the quality of the saliency maps have much impact. 
\cite{suresh2021intuitively} propose to visualize a given input's nearest neighbors to help better reason about the model's uncertainty and show an editor that allows users to edit aspects of the input and see how model predictions change, they found that this interface allowed some clinicians to 
build better intuition about the AI capabilities and limitations. Finally, a line of work has focused on human-AI interaction in healthcare applications: on chest X-rays \cite{xie2020chexplain,gaube2021ai},  diabetic retinopathy \cite{beede2020human}, skin cancer \cite{tschandl2020human} and breast cancer \cite{bejnordi2017diagnostic}. \cite{yin2019understanding} study the effect of initial debriefing of stated AI accuracy compared to observed AI accuracy in deployment and find a
significant  effect of stated accuracy on trust, but that diminishes quickly after observing the model in practice; this reinforces our approach of building trust through examples that simulate deployment.

\paragraph{Explainability.} Methods for explaining the decisions of ML models range from feature attribution (e.g. LIME \cite{ribeiro2016should}), saliency methods for computer vision tasks (Grad-CAM \cite{selvaraju2017grad}), Example-based explanations  \cite{kim2016examples} and others. One of the basic forms of model explanations is calibrated confidence scores \cite{guo2017calibration,tohme2021improving,vanslette2020general}
These methods for explainability start from a set of desiderata (natural assumptions of what an explainability method should provide) and then formulate a given method that can be implemented without further data requirements. The common pitfall of these methods is that they are agnostic to the downstream expert, the desiderata is formulated from a perspective of a rational expert and are sometimes justified from user studies. 

\paragraph{Machine Teaching.} Machine teaching (MT) refers to the problem of choosing a minimally sized dataset that enables a student learner to learn a specific target function \cite{zhu2018overview}. Given a hypothesis class, its teaching dimension is the smallest sized set that enables an ERM learner to pick out the optimal classifier \cite{goldman1995complexity,kumar2021teaching}.
To mimic human learners, \cite{singla2014near} proposes a Bayesian learner based on a prior over a discrete hypothesis class, the learner maintains a distribution over each hypothesis that updates with each teaching example. They evaluate their approach on an image classification task where crowdworkers learn to distinguish different animals. 
 This setting was extended to include explanations in the form of attention \cite{chen2018near,su2017interpretable} and errors in learning priors and teacher knowledge \cite{devidze2020understanding}. \cite{dasgupta2019teaching} aims to teach a consistent black-box learner, while this formulation is attractive in regards to a human learner, the algorithm they provide requires an excessive amount of queries to the human that go beyond the teaching examples presented. \cite{hunziker2018teaching}  teaches a forgetful human learner multiple concepts where each concept maps to a single example, but the human may forget the concept later on. Our work separates itself by the use of a novel radius nearest neighbor model to approximate the human learning process.
 
\paragraph{Human Learning.} 
\cite{bornstein2017reminders} make the claim that humans makes decision by sampling similar experiences from memory instead of computing reward estimates for each possible action. Their experimental study involves users performing a two-armed bandit task with each example having a unique identifier. 
\cite{giguere2013limits} make two claims about how humans make decisions: the first is that people often retrieve a limited set of items from memory when making decisions and the second is that training humans on idealized instances is more advantageous than training them on noisy or hard instances. They base their claims on two experiments: one where humans classify horizontal lines of different lengths and the other where they judge outcomes of baseball games. 
\cite{richler2014visual} review the literature on visual category learning, how we distinguish between different visual objects. They make a distinction between two different models of human decision making. The first is example-based that models assume that
a category is  represented in terms of the
particular exemplars that have been experienced
during learning.  The other is rule based, people try to explicitly learn categories by forming simple rules. The conjecture is that for hard tasks, the example-based model is more accurate while for simpler ones, the rule-based approach is the driver.

\paragraph{Nearest Neighbor Compression.} Our human student model is a more general case of a weighted nearest neighbor learner, this makes the teaching problem equivalent to that of compressing the number of samples nearest neighbors requires. Seminal work on compressing nearest neighbors introduced the condensed nearest neighbor rule \cite{gates1972reduced} and follow-up work introduced more robust versions but that still require the existence of a consistent subset  \cite{angiulli2005fast}. More recent work has focused on the generation of compressed subsets \cite{kusner2014stochastic,zhong2017fast,gupta2017protonn}.

\subsection{Relation to Learning to Defer.}

Our framework of Human-AI assisted decision making, dubbed teaching to defer (TTD), and its associated framing can be considered as the analog of the learning to defer framework described in \cite{mozannar2020consistent} (LTD). In our setting, the human observes the  AI prediction and then makes a prediction. In LTD, the AI model first decides using a rejector whether to predict on its own or defer to the human. There is no interaction in LTD between the human and the AI as the goal is to reduce the burden on the human expert. We borrow the notion of a rejector to formalize the thought process of the human deciding to or not to use the AI prediction. Table \ref{tab:ttd_ltd_differences} highlights some of the main differences between the two frameworks.

\paragraph{System Objective.} The objective in our framework is stated in equation \eqref{eq:second_opinion_loss}, which can be compared to the  the system objective from LTD \cite{mozannar2020consistent}:

\begin{equation}
    L(h,r)= \label{eq:ltd_reject_loss}  \bE_{(x,y)\sim \mathbf{P},m \sim M|(x,y)} \ [ \ l(x,y,h(x)) \bI_{r(X) = 0} +  l_{\textrm{exp}}(x,y,m) \bI_{r(x)=1} \  ]  
\end{equation}
Beyond the fact that in TTD, the human controls the rejector and in LTD the AI controls the rejector, a technical difference is the input to the rejector function $r$: in LTD it's the AI domain $X$, while in TTD it's the human domain $Z$ and the AI prediction $\pi(X)$.

\paragraph{Human-AI Interaction.} In LTD, when the AI predicts or defers, it does so without observing the human's prediction, and when the human predicts, they do so without seeing the AI prediction. On the other hand, in our framework, the human observes the AI's prediction and explanation before making their final prediction. This allows the human and AI to combine their  predictions in a way that the LTD framework does not allow. 

\begin{table}[h]
\caption{Comparison on different dimensions between the teaching to defer framework in this paper (TTD) and the learning to defer framework (LTD) from \cite{mozannar2020consistent,madras2018predict,raghu2019direct}. }

\begin{tabularx}{\textwidth}{l|XX}
\toprule
Dimension & LTD \cite{mozannar2020consistent}  & TTD (this paper)  \\
\midrule
Information at training & Samples from AI domain $X$, label $Y$, human prediction $M$ & Samples from AI domain $X$, Human domain $Z$, label $Y$, and error distribution of AI and Human\\

Information at testing &  AI domain $X$ &  AI domain $X$, Human domain $Z$, AI prediction $\pi$\\
AI training & joint training with rejector & trained without knowledge of human rejector   \\
Knowledge about human & samples of prediction & error distribution \\
Form of rejector & no constraint & radius nearest neighbor defined by Assumption \ref{ass:human_rejector_form} \\
Interaction between Human and AI & No by design, AI doesn't see the Human prediction and Human doesn't see the AI prediction & Yes \\
Final decision maker & AI or Human & Human \\
Does Human observe each example & No, since AI might not defer & Yes \\
Ease of Deployment & Needs re-training for every human expert & Same deployment for any human expert \\
\bottomrule
\end{tabularx}

\label{tab:ttd_ltd_differences}
\end{table}

%% file: appendix/proofs.tex
\section{Theoretical Results and Proofs}\label{apx:proofs}

\subsection{Further Derivations}
We expand on section "Teaching a Student Learner" \eqref{sec:approach} and decompose the loss of the human learner. 

Since $\pi_Y(x)$ and $h(Z,A)$ are known and fixed, we can assign to each deferral decision at each point $i$ a cost $c_i(r) \in \mathbb{R}^+$ and abstract away the inner classification decisions:
\begin{equation}
L(D) := \sum_{i \in S} l_c(r(z_i,a_i;D);c_i)
\end{equation}
An example of $l_c$ is $l_c^{b}:=r(x_i;D) c_i(1) + (1-r(x_i;D)) c_i(0)$ which can be made equivalent to the 0-1 classification loss.
It may be the case that neither of $c_i(0)$ or $c_i(1)$ are zero since there may be multiple correct decisions or that both be may be non-zero and equal. Now we further decompose the loss $L$ into errors made by the prior and errors due to the learned rejector:

\begin{align}
L(D) &= \sum_{i \in S \ | B(z_i) \neq \emptyset}  l_c \left( \frac{\sum_{j \in B(z_i)} \mathbb{I}\{r_j = 1\} K(z_i,z_j)  }{\sum_{j \in B(z_i)}  K(z_i,z_j) } ;c_i \right) \quad \textrm{(errors  by learned rejector )}  \\
&+\sum_{i \in S \ | \ B(z_i) = \emptyset}  l_c \left( g_0(z_i,a_i) ;c_i \right) \quad \textrm{(errors  by prior)}
\end{align}

In the paper, we proved a guarantee in Theorem 1 on the performance of the \texttt{GREEDY-SELECT} algorithm when the hyperparameter $\alpha$ is set to $1$ when optimizing the loss $L(.)$ (4). The loss $L$ involves the human learner $M(.)$, however, one component of the human learner was left unspecified which is how they set the radius $\gamma$ following every teaching example $z$. In what follows, we assume the human is perfectly learning the radius that the teaching process displays to them. Equivalently, when the human is shown the tuple $\{z,\gamma,r\}$ where $z$ is the teaching example, $\gamma$ is a radius and $r$ is the deferral action, they now follow the deferral action $r$ in the neighborhood of size $\gamma$ around $z$. 

When we set $\alpha=1$, this defines a unique radius $\gamma_i$ to each point $z_i \in S^*$ (the teaching set), this radius defines the largest neighborhood around $z_i$ such that the optimal deferral action in that neighborhood is $r_i$. Thus our teaching set becomes $S^*=\{x_i,z_i,\gamma_i,r_i\}$ and we can now simplify our optimization problem by only searching for the teaching point $z$ at each step (instead of jointly searching for the radius as well) as the radius is uniquely specified no matter what the current teaching set $D_t$ is.

\subsection{Proofs}

The following proposition is part of the proof of Theorem 1.

\textbf{Proposition 2.} \textit{ \noindent
Let $F(X)= L(\emptyset) - L(X)$, $F(.)$ is submodular, monotone and positive. 
 }
 
 \begin{proof}
\textbf{Monotonicity.}
 We prove that $L(.)$ is monotone decreasing which implies that $F(.)$ is monotone increasing. For notation simplicity we omit the AI message $A$ from the prior rejector and make it only a function of $Z$, the proof remains valid even if we add $A$.

Initially $D_0= \emptyset$ and $L(\emptyset)$ is the error rate of the human's prior rejector $g_0$ on the set $S$. 

\textbf{Induction argument:
}
In the first step $D_1 = \{z_{i1}\}$ where $z_{i1}$ is the training example that leads to the biggest error decrease of $L(.)$ (we  don't use this fact so that this holds for any training example added, also note that since there is a unique correspondence from $z_{i1}$ to $r_{i1}$ and $\gamma_{i1}$ we simplify the notation and only write $z_{i1}$), now note that:
\begin{align}
L(D_1) - L(D_0) = \sum_{i \in S \  s.t. \ z_{i1} \in B(z_{i})}  l_c \left( r_{i1}  ;c_i \right)  - l_c(g_0(z_i);c_i) \label{eq:diferenceloss_step1}
\end{align}
Note that other terms in the difference of equation \eqref{eq:diferenceloss_step1} cancel out, what is left are points in $S$ that the human starts to use their learned rejector on, i.e. those that are sufficiently close to $z_{i1}$ call these set of points $\mathcal{I}$. For each $i \in \mathcal{I}$, if it was the case that $g_0(z_i) \in \arg \min_d l_c(d;c_i)$, then we know that $r_{i1}$ and $g_0(z_i) $ have the same cost since $r_{i1}$ is the optimal decision by definition. Now suppose that $g_0(z_i) \notin \arg \min_d l_c(d;c_i)$, then it must be the case that $r_{i1} = 1 -g_0(z_i)$ and this achieves a lower loss than $g_0(z_1)$. Therefore we have that:
\[
L(D_1) - L(D_0) \leq 0
\]

Now suppose we are at step $t+1$ of the algorithm and we add example $z_{i(t+1)}$ to obtain $D_{t+1} = \{ z_{i1},\cdots,z_{i(t+1)}\}$. Let us compute the difference: 
\begin{align}
L(D_{t+1}) - L(D_t) = \sum_{i \in S \  s.t. \   B(x_{i}) = \{ z_{i(t+1)}\} }  l_c \left( r_{i(t+1)} ;c_i \right)  - l_c(g_0(z_i);c_i) \label{eq:diferenceloss_step2}
\end{align}
Note that if there  was point $i \in S$ where there exists $j \in D_t$ such that $z_j \in B(x_i)$, then the addition of $z_{i(t+1)}$ cannot change the final cost assigned to example $i$ as if $z_{i(t+1)} \in B(z_{i}$, then we must have  $r_{i(t+1)}=r_{j}$  by  assumption \ref{as:gamma_and_kernnel}. Thus the only element remaining in the difference is points that now have a neighbor in $D_{t+1}$ but not in $D_t$, meaning those that only have $z_{i(t+1)}$ in their ball. The argument is now exactly as in the base case so that:
\[
L(D_{t+1}) - L(D_t)  \leq 0
\]
which gives us the set of inequalities:
\[
L(D_m) \leq \cdots \leq  L(D_0)
\]
and note that $L(.)$ achieves it's minimum value at $L(S) = L(D_{|S|}) \leq L(D_m)$.

\textbf{Positivity.} Note that $F(.)$ is positive as we assume $l_c$ is positive and we obtain the result from monotonicity.

\textbf{Submodularity.
}
To make the proof easier, define the teaching ball $\Tilde{B}(D)$ to be the set of points in the training set $S$ that have any teaching point $Z \in D$   in their ball $B(.)$. This implies if $B(z_i) = \{ z_j \}$ then $z_i \in (\{z_j\})$; remember that $B(z_i)$ is the set of teaching points that are sufficiently close to $z_i$.
Let $A \subset B \subset S$, let $l \in S \setminus B$, let us compute:

\begin{flalign}
    &F(A \cup \{l\}) - F(A) - F( B \cup \{ l\}) + F(B) 
    =   L(A)-L(A \cup \{l\})  + L( B \cup \{ l\}) - L(B) \\
    &= \sum_{i \in S \  s.t. \ z_i \in \tilde{B}(z_{l}) \setminus \tilde{B}(A) }  l_c(g_0(z_i);c_i) - l_c\left( r_{l}  ;c_i \right)   \nonumber\\&+  \sum_{i \in S \  s.t. \ z_i \in \tilde{B}(z_{l}) \setminus \tilde{B}(B) }  l_c\left( r_{l}  ;c_i \right) - l_c(g_0(z_i);c_i)  \\
    &= \sum_{i \in S \  s.t. \ z_i \in \left(\tilde{B}(z_{l}) \cap \tilde{B}(B)\right) \setminus  \tilde{B}(A) }  l_c(g_0(z_i);c_i) - l_c\left( r_{l}  ;c_i \right) 
    \geq 0
\end{flalign}
The last term is positive as the optimal decisions $r_i$ always improve on the prior.

\end{proof}

\textbf{Theorem 1.} \textit{ \noindent
Let $F(X)= L(\emptyset) - L(X)$, $F(.)$ is submodular, monotone and positive. Moreover, the \texttt{GREEDY-SELECT} algorithm described above achieves the following performance compared to the optimal teaching set $D^*$:
\vspace{-2mm}
\begin{equation}
    \underbrace{L(D_m)}_{\textrm{loss of chosen set}} \leq (1-\frac{1}{e}) \underbrace{L(D^*)}_{\textrm{loss of optimal set}} + \frac{1}{e} \underbrace{L(\emptyset)}_{\textrm{loss of prior rejector}}
\end{equation} }

\begin{proof} The first statement of the theorem is proved in Proposition 2.

For the second statement of the theorem, the proof is simply restating the proof of Theorem 1.5 in \cite{krause2014submodular} in the context of our problem which we do here for clarity.
Let $D_i=(z_1, \cdots, z_i)$ the set that our algorithm produced at round $i$ and $D^*=(z_1^*, \cdots, z_K^*) $ the optimal set.

For all $i \leq m$:
\begin{align}
    F(D^*) &\leq F(D^* \cup D_i) \quad \textrm{(monotonicity)} \\
    &= F(D_i) + \sum_{j=1}^m F(D_i \cup D^*_{j-1} \cup z_{j}^*) - F(D_i \cup D^*_{j-1}) \quad \textrm{(telescoping)}\\
    &\leq F(D_i) + \sum_{z \in D^*} F(D_i  \cup z) - F(D_i ) \quad \textrm{(submodular $F$)}\\
    &\leq F(D_i) + m( F(D_i  \cup z^{i+1} ) - F(D_i )) \quad \textrm{(optimality of $z^{i+1}$)}\\
\end{align}

re-arranging this final inequality with $\delta_{i+1} =F(D^*) - F(D_i) $ we get:
\[
\delta_{i+1} \leq \delta_i (1-\frac{1}{m}) 
\]
iterating this last inequality till $m$, using the fact that $1-x \leq e^{-x}$ and restating things in terms of $L(.)$ gets the final result in the theorem.
\end{proof}

\subsection{Hardness result}
Theorem 1 gives a guarantee on the subset chosen by the greedy algorithm with an $1-\frac{1}{e}$ approximation factor, one can ask if we can do better. We prove that a generalization of our problem under Assumption \ref{as:gamma_and_kernnel} is in fact NP-hard. 

\textbf{Proposition 1.} \textit{ \noindent
Problem \eqref{eq:optimal_subset_prob_delta} is NP-hard.
 }

\begin{proof} For simplicity we assume that the AI and human domains are identical and don't consider the AI message in the human rejector or predictor. The proof can be straightforwardly extended to the case when the domains differ and including the AI message.

Suppose we are given a collection of finite sets $A_1,\cdots,A_n$ jointly
covering a set $W$. We reduce the problem of finding a smallest subcollection covering $W$ to the teaching problem \eqref{eq:optimal_subset_prob_delta}.

Let $S_V = W$, for each $A_j$, we associate it with a new teaching example $x_j \in S_T$ (unique from all elements of $S_V$ and other elements of $S_T$) such that its neighbors are exactly the elements of $A_j$ i.e. $K(x_j,x) = \infty$ iff $x \in A_j$ and $K(x_j,x)=0$ iff $x \notin A_j$ (we construct the function $K$ specifically to satisfy these requirements). Now we set the label $y_i =1$ for each example $i \in S_V \cup S_T$ and let  $h(x) = 1$ (human predictor) and $\pi_Y(x) = 0$ (AI predictor) for all $x$ and we set $g_0(x) = 1$ (human prior rejector) so that the prior is wrong on all example: we should never defer while the prior always defers so the correct deferral decision is $d_i =0$ (derived deferral decision) for all examples. We set the loss $l_c$ to simply be the $0-1$ deferral loss (cost of $1$ incurred if final prediction disagrees with label, otherwise a cost of $0$), with this in mind note that $L_V(\emptyset) = |S_V|$ as with $D= \emptyset$ we use the prior rejector on all examples which always errs.

Once we pick a new example $x_j$ (correspondence to the set $A_j$) to our set $D$ that we are choosing, the only terms that are affected are  those that are close to $x_j$ which are exactly the elements of $A_j$, so that $L_V(\{x_j \}) = |S_V| - |A_j|$. Iteratively, when we add another example $x_k$ to $D$ the only terms affected are those in the neighborhood of $x_k$ which are $A_k$, but now it may be the case that $A_j \cap A_k \neq \emptyset$, however since the deferral label associated to all examples is the same, which is to not defer, the loss of the elements in the intersection are not affected (in essence there is no double counting of the elements) so that now: $L_V(\{x_j,s_k \}) = |S_V| - |A_j \cup A_k|$. It is now clear to see that solving problem \eqref{eq:optimal_subset_prob_delta} with $\delta =0$ finds a set cover of $W$ with elements $A_1,\cdots,A_n$ as $L_V(D)$ simply counts how many elements of $S_V$ (correspondence to $W$) we don't apply the prior rejector to (i.e. elements we cover).

\end{proof}

\subsection{Efficient Implementation of Greedy Selection}

When $\alpha=1$, we provide an efficient implementation of the greedy selection algorithm  \texttt{GREEDY-SELECT}.

At each round, we have a teaching set $D_t$ from which we can construct a rejector function $g_t(.,.)$, at $D_0$ we have $g_0$ is the prior rejector.
Now  at round $t$, we calculate for each example on the training set $S$ the following quantity 

\begin{equation}
    E_i^t = \sum_{j \in S \ | \ K(z_j,z_i) \geq \gamma_i } \bI_{g_t(z_j,a_j) \neq r_j}
\end{equation}
$E_i$ counts the number of points that are in the neighborhood of $z_i$ that the current human rejector $g_t$ misclassifies, in other words it measures for each point $i$ how many points in the training set it will cause their deferral label to flip. Note that we are guaranteed that once a point is close enough to the teaching point $z_i$, it's deferral decision becomes optimal by Assumption \ref{as:gamma_and_kernnel}. 
 At at each round $t$ we pick the point $i^* = \arg \max_i E_i^t$.
 
 This algorithm has run-time $O(n^2m)$ where $n = |S|$, while the naive implementation of the algorithm has run-time $O(n^2m^2)$, the extra $m$ factor comes from having to simulate the human rejector to calculate the resulting loss. 
 
 When we are optimizing over the choice of radius jointly with the choice of training point, we have no other choice but to fully simulate the human rejector. But note that the optimization over the radius can be reduced to only looking at radius choices that are equal to kernel similarities on the training set. 

%% file: appendix/ai_error_regions.tex
\section{SAE Model Error Analysis}\label{apx:ai_error_region}
\paragraph{Predictions.} The below analysis is performed from allowing the model SAE-large model \cite{tu2020select} whose code is available at \footnote{\url{https://github.com/JD-AI-Research-Silicon-Valley/SAE}} to predict on the HotpotQA DEV set \cite{yang2018hotpotqa} with no distractor paragraphs. The model is ranked 20'th on the public leaderboard, and is the highest ranking model with publicly available code.

\subsection{Factors of difference}
\paragraph{Presence of distractors.} There are two types of question answer types in HotpotQA: yes/no answers and answers that are substrings from the passage. We eliminate yes/no questions and only focus on questions that admit an answer inside the passage which makes the validation set of size 6947 out of an original 7405. 
\begin{table}[h]
\label{table:sae_distr}
\caption{Performance on the dev set without yes/no questions.}
\vskip 0.15in
\begin{center}
\begin{small}
\begin{sc}
\begin{tabular}{lcr}
\toprule
Factor & Exact Match (EM) &F1  \\
\midrule
8 distractors
&
66.92 &	79.62\\
No distractors
&
68.79 &	82.75\\
\bottomrule
\end{tabular}
\end{sc}
\end{small}
\end{center}
\vskip -0.1in
\end{table}
We note that the absence of distractor paragraphs does not boost performance by a significant amount. In fact the model SAE first consists of a relevant paragraph extractor that feeds into the RoBERTa reader and that extractor works quite well as evidenced. 

\paragraph{Bridge vs comparison questions.} The questions in HotpotQA can be categorized into two types: \textbf{bridge} e.g. "“when was the
singer and songwriter of Radiohead born?”, to answer this question one first has to figure out who is the singer of Radiohead and then look up his date of birth, the other type are \textbf{comparison} questions such as that “Who has played for more NBA teams, Michael
Jordan or Kobe Bryant?". This categorization is provided already in the dataset. 

\begin{table}[H]
\caption{Performance based on question types.}
\vskip 0.15in
\begin{center}
\begin{small}
\begin{sc}
\begin{tabular}{lcr}
\toprule
Factor & Exact Match (EM) &F1  \\
\midrule
Bridge
&
68.31 &	83.25\\
Comparison
&
71.52 &	79.86\\
\bottomrule
\end{tabular}
\end{sc}
\end{small}
\end{center}
\vskip -0.1in
\end{table}
We can see that there is a difference in how question types affect performance, however it is not consistent across the two metrics to make a definite conclusion.

\paragraph{Passage Lengths.} Given the length of the two golden paragraphs, is there a difference in the performance over different sizes? 
As we can see below we observe no significant difference, in the last bucket of long passages we see a notable increase in F1 but that is due to limited sample size in extremely long passages.
\begin{figure}[H]
    \centering
    \includegraphics[scale=0.6]{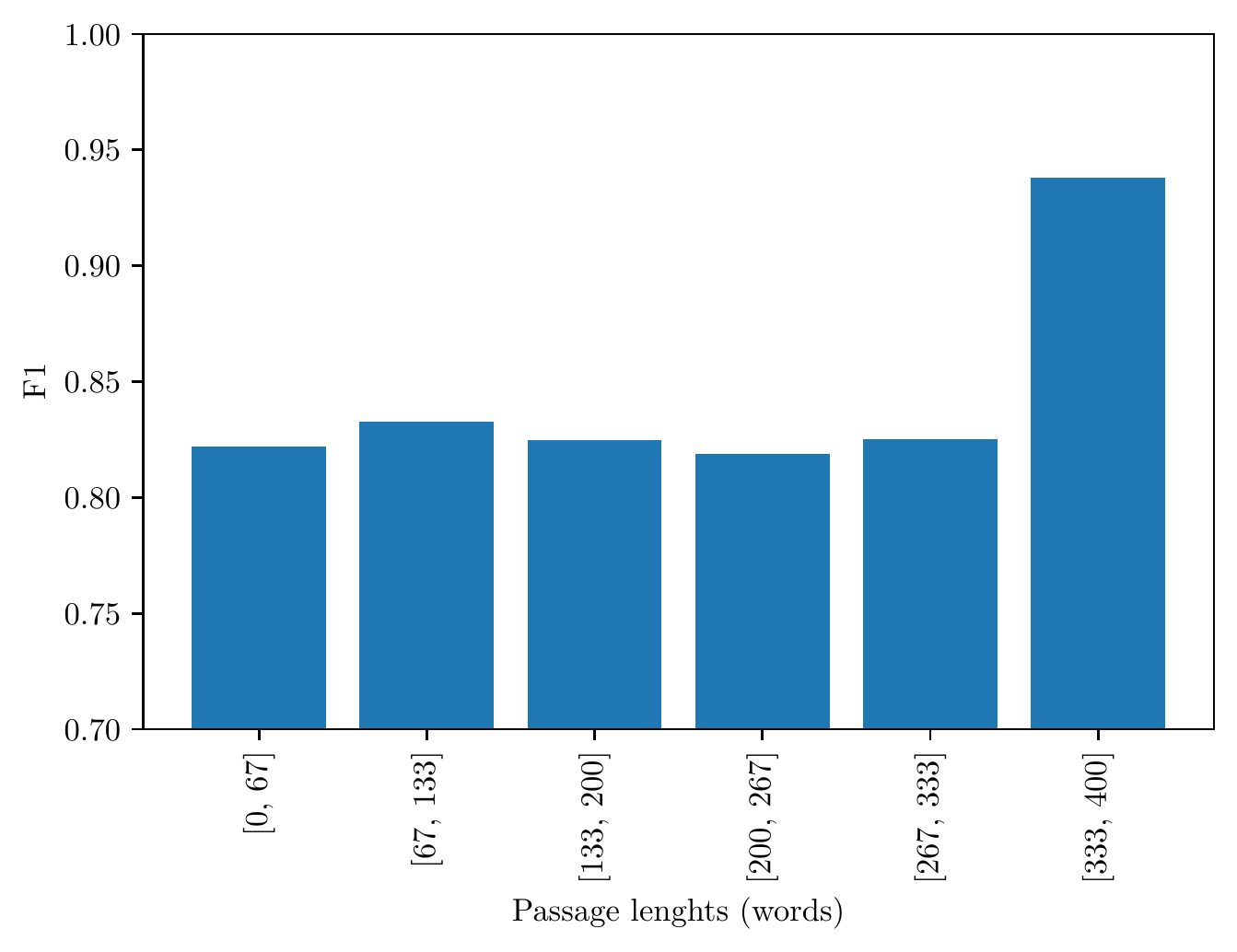}
    \caption{Performance across lengths of passages in terms of words. First bin contains very little samples to be significant. }
\end{figure}

\paragraph{Supporting fact lengths.} We plot the performance versus the number of supporting facts: the number of sentences one must read to answer the question, this is provided in the dataset explicitly. Note there are at least two sentences that one must read since all questions are multi-hop. We can see that there is no real difference across all lengths. 
\begin{figure}[H]
    \centering
    \includegraphics[scale=0.6]{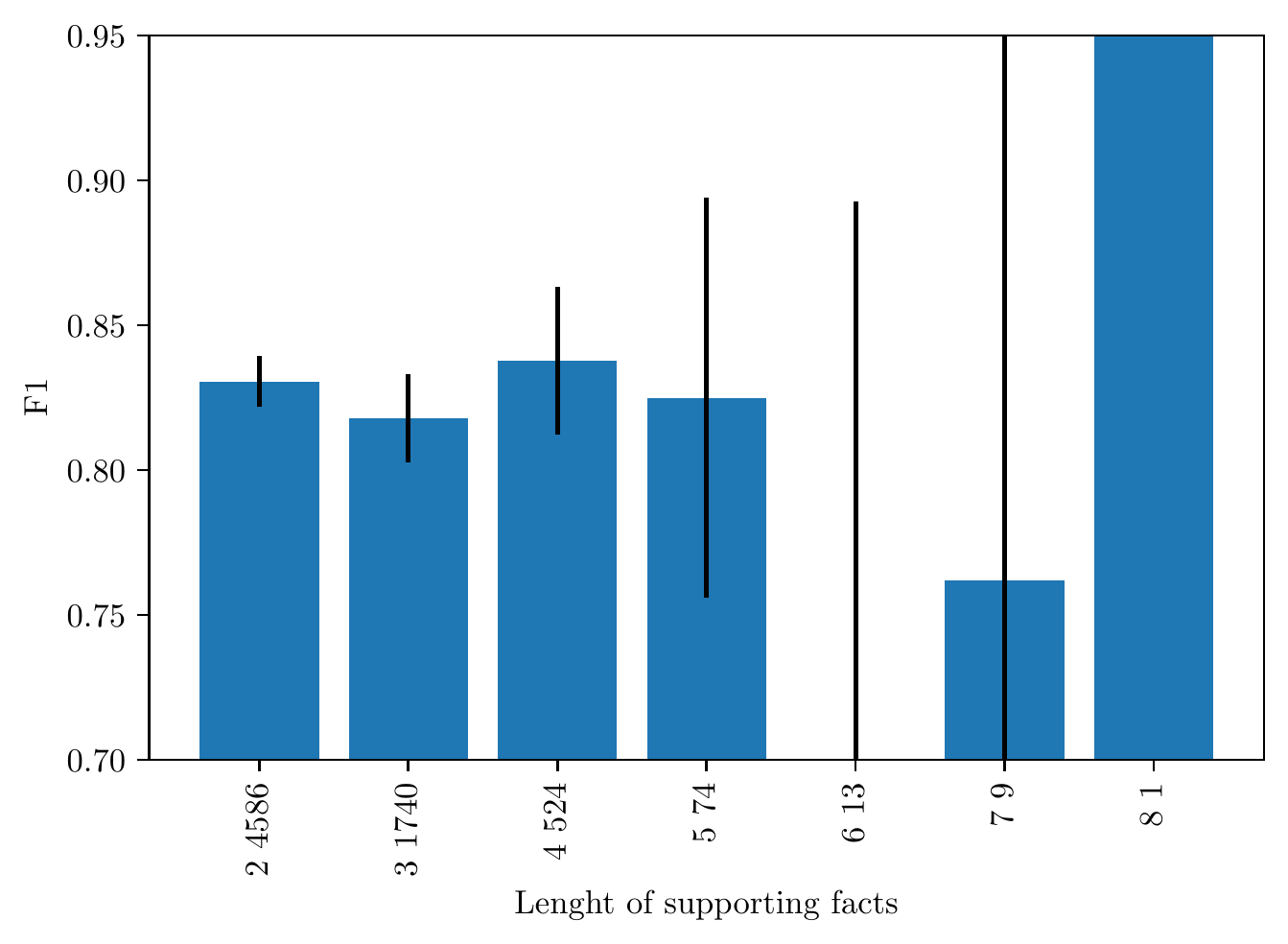}
    \caption{Performance  across number of supporting sentences. Black bars indicate 95\% confidence interval around the mean, the x axis is: (number of sentences, number of examples with that many sentences) }
\end{figure}

\paragraph{Passage and Question topics.} We try to see if there is a difference in performance when looking into the topics that the examples belong to. We first run an LDA with 15 topics on the passage concatenated with the question (we use the $gensim$ package \cite{rehurek_lrec}). We then categorize each example according to the topic with the largest coefficient in the LDA decomposition. 

\begin{figure}[H]
    \centering
    \includegraphics[scale=0.6]{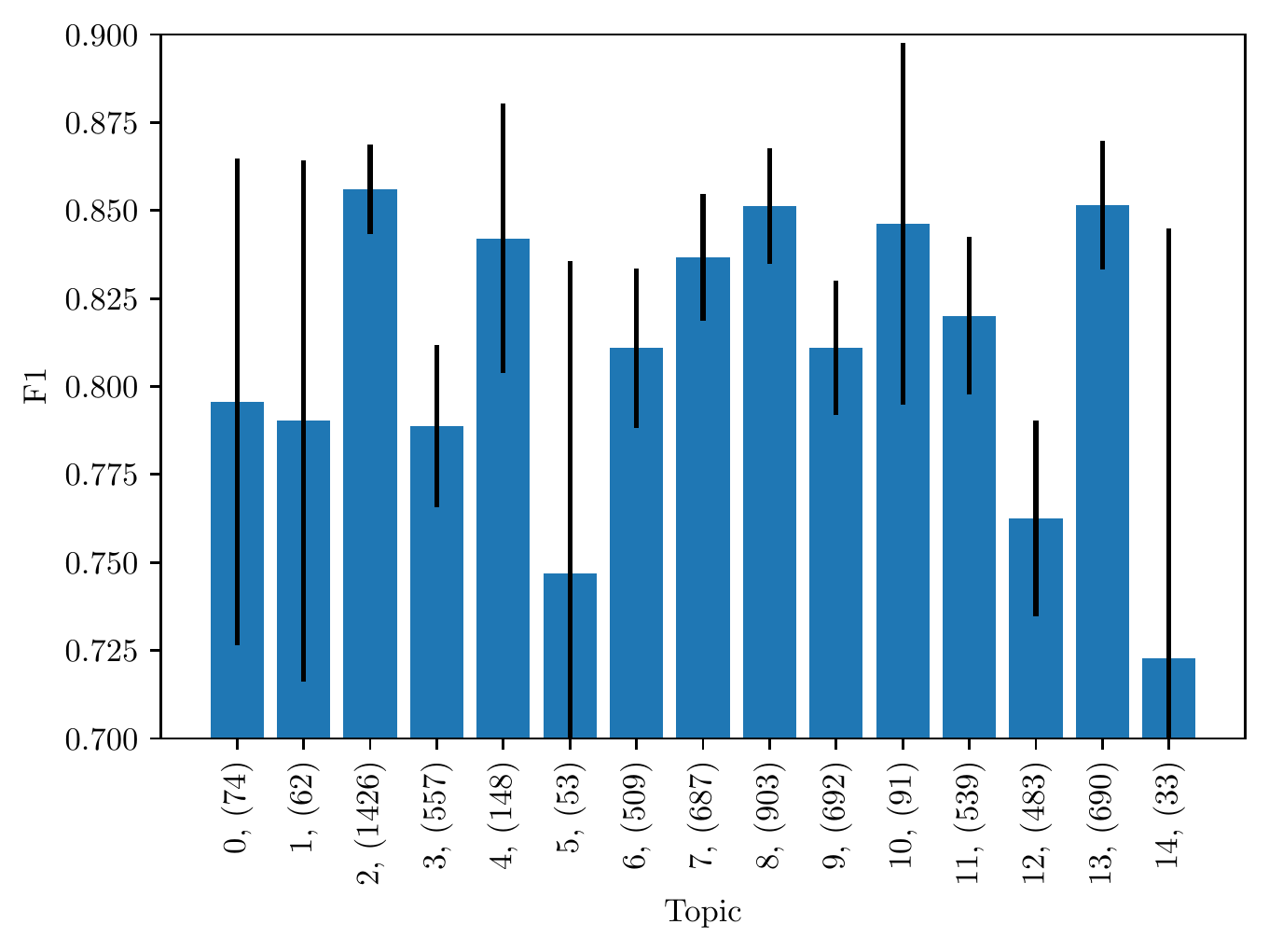}
    \caption{Performance across LDA topics }
    \label{fig:lda_topics}
\end{figure}

Plotted in Figure \ref{fig:lda_topics} are mean F1 across topic and 90\% confidence intervals and we observe no particular topic that has significant difference from others.

\paragraph{Question words.} We investigate difference in performance depending on the question word present in the question.
\begin{figure}[H]
    \centering
    \includegraphics[scale=0.6]{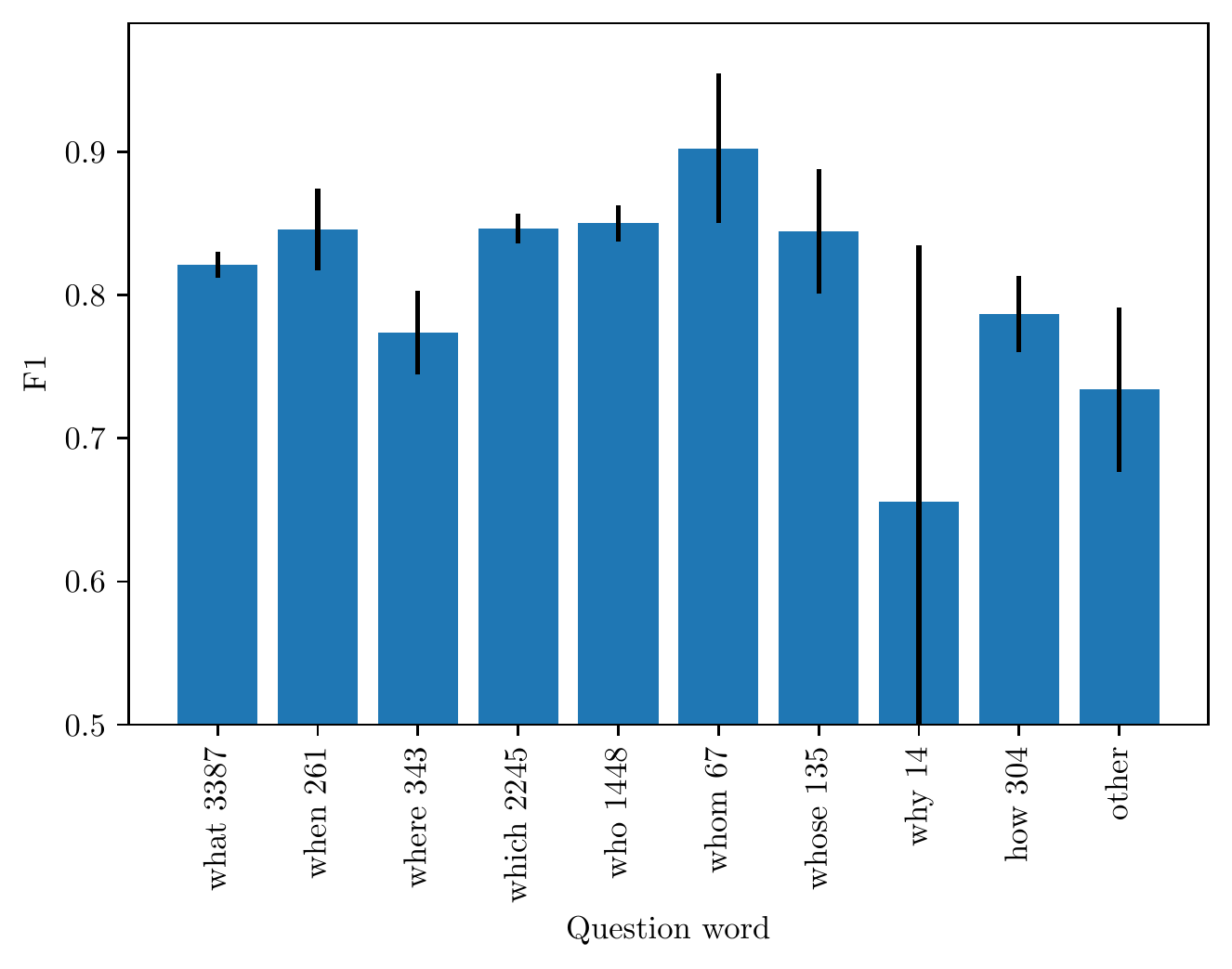}
    \caption{Performance (left EM, right F1) across question words }
\end{figure}

We can see that there is significant difference with "why" questions (however they are rare) and "how" questions to a lesser degree.

\subsection{Embedding clustering}
\paragraph{Model embeddings.} The SAE model last layer consists of a 512x1024 tensor: a 1024 representation of 512 tokens. This representation is then used to predict for each token the probability that it is the start or end of the answer with a linear layer. To get a vector representation of each example, we average out across tokens to obtain a single 1024 vector for each example. We take these vectors and cluster them using K-means (we do the analysis for multiple k's). We then plot the performance across each cluster below. 

\begin{figure}[H]
    \centering
    \includegraphics[scale=0.7]{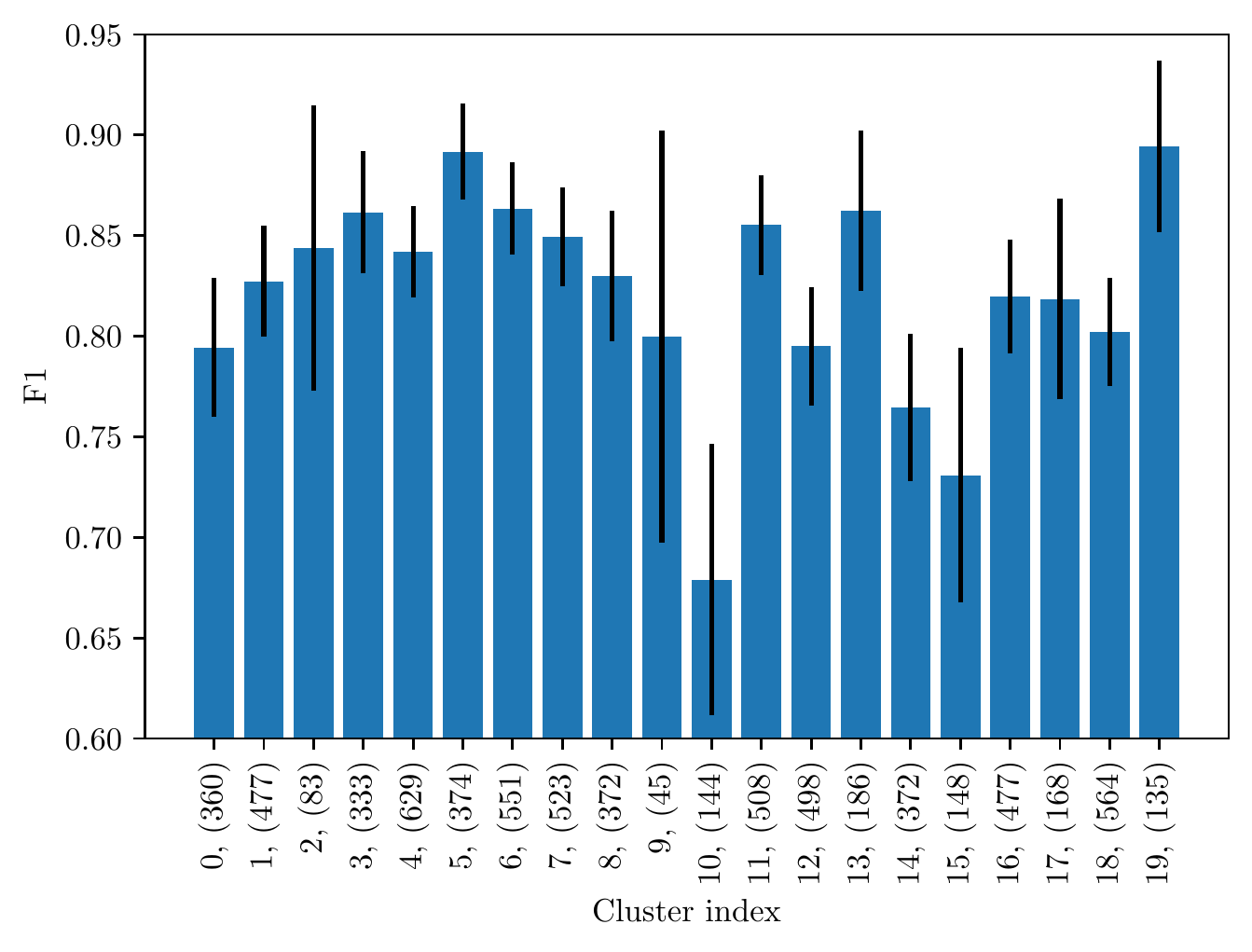}
    \caption{Performance (left EM, right F1) across model embeddings clusters. }
\end{figure}

We observe that cluster 10 has lower performance than average by a significant amount. Looking at examples from that cluster, no apparent theme emerges.

\paragraph{Passage embeddings.} We use the BERT sentence encoder \footnote{\url{https://github.com/UKPLab/sentence-transformers}} to get embeddings for the passage and cluster them using k-means. We repeat the exact process for the questions and answers.
\begin{figure}[H]
    \centering
    \includegraphics[scale=0.6]{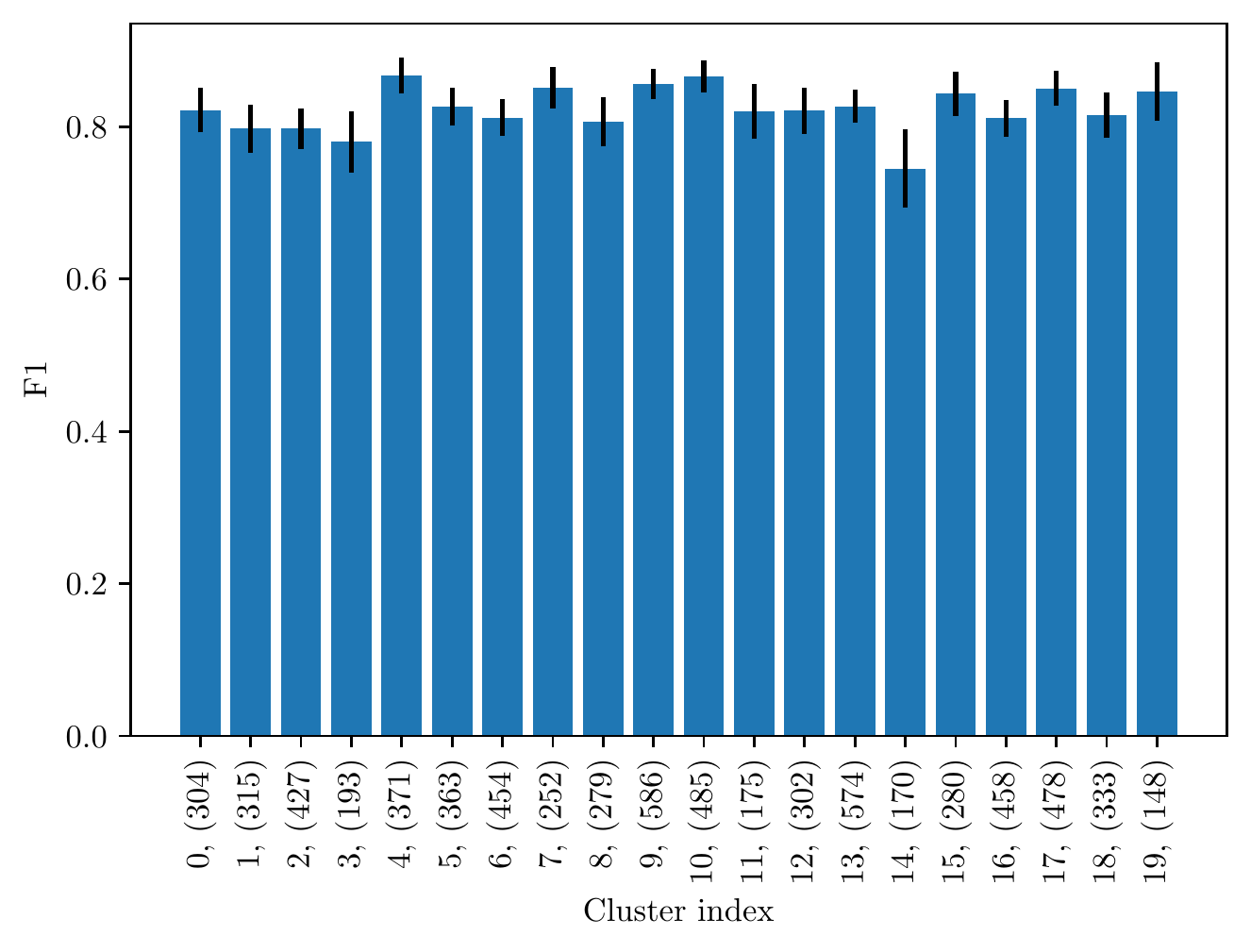}
    \caption{Performance  across passage embeddings clusters. No differences emerge significantly. }
\end{figure}

\paragraph{Question embeddings.}We can see that cluster 13 undeperforms, examining that cluster we can see a pattern of questions like "What city does Paul Clyne and David Soares have in common?", the theme is the "in common" at the end of the question.

\begin{figure}[H]
    \centering
    \includegraphics[scale=0.6]{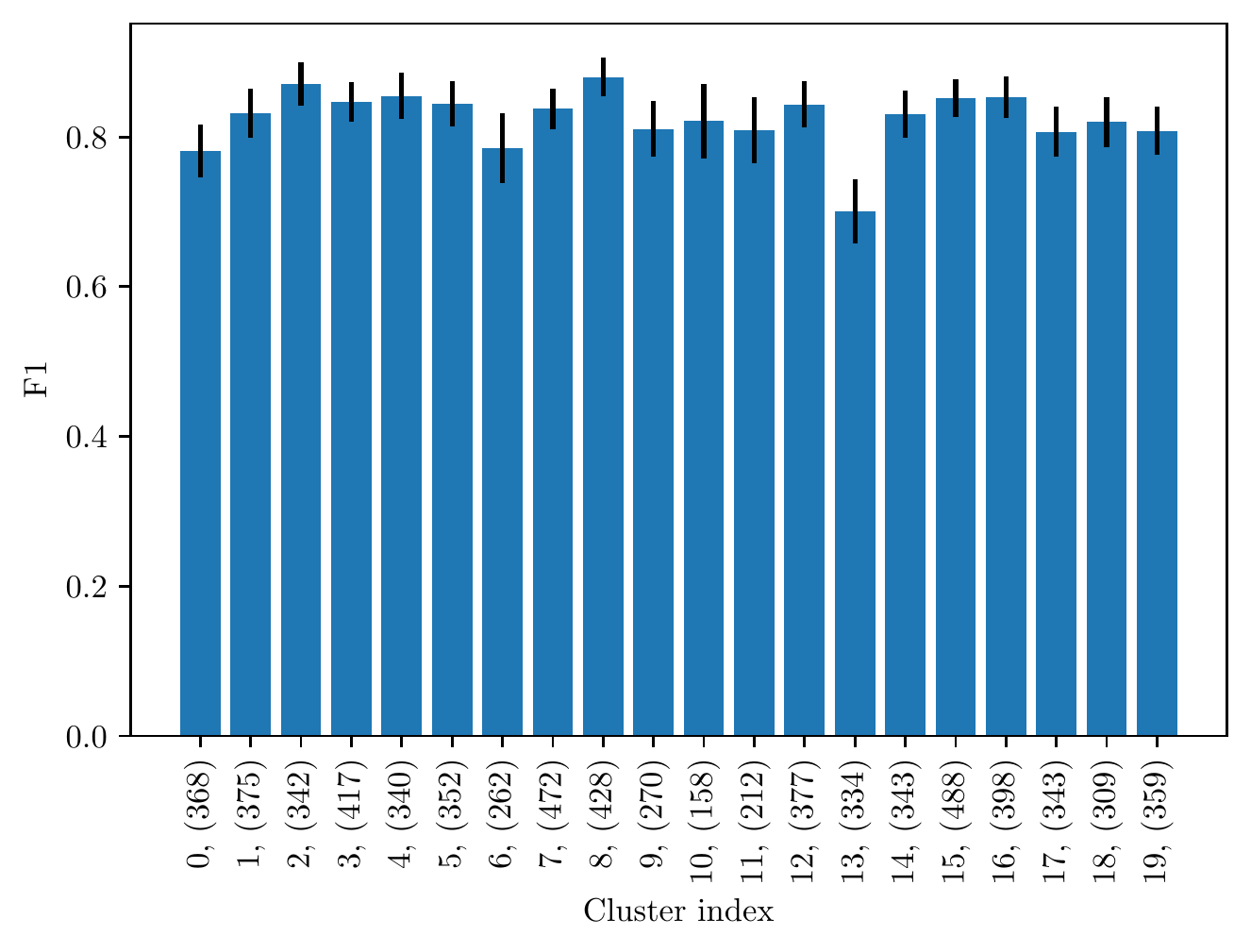}
    \caption{Performance  across question embeddings clusters. }
\end{figure}

\paragraph{Answer embeddings.} We observe no observable theme or significant differences.

\begin{figure}[H]
    \centering
    \includegraphics[scale=0.6]{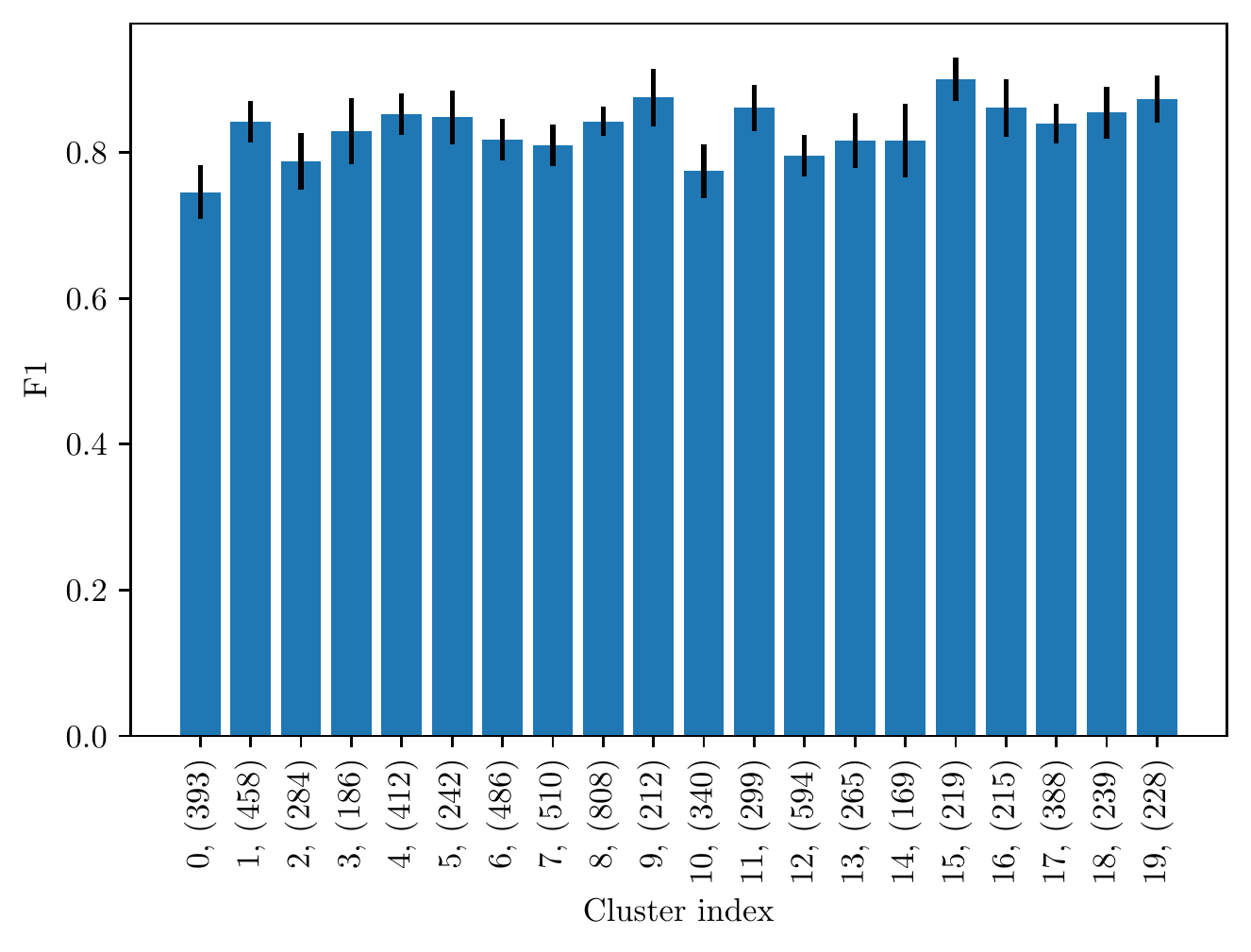}
    \caption{Performance  across answer embeddings clusters.}
\end{figure}